\documentclass[twoside,11pt]{article}

\usepackage[abbrvbib, preprint]{jmlr2e}
\usepackage{hyperref}       % hyperlinks
\usepackage{url}            % simple URL typesetting
\usepackage{booktabs}       % professional-quality tables
\usepackage{amsfonts}       % blackboard math symbols
\usepackage{nicefrac}       % compact symbols for 1/2, etc.
\usepackage{microtype}      % microtypography
\usepackage{xcolor}         % colors
\usepackage{amsmath}        % align
\usepackage{stackengine} % for gaps between underbrace
\usepackage{wrapfig} % 

\DeclareMathOperator*{\Attn}{Attn}
\newcommand{\ignore}[1]{}

\definecolor{nvibgold}{HTML}{FFBF00}
\definecolor{nvibgreen}{HTML}{009e73}
\definecolor{nvibpurple}{HTML}{9400d3}

\usepackage{lastpage}
\jmlrheading{~}{~}{~}{~}{~}{~}{Fabio Fehr and James Henderson}
\ShortHeadings{Nonparametric Variational Regularisation}{Fehr and Henderson}
\firstpageno{1}

\begin{document}

\title{Nonparametric Variational Regularisation \\ of Pretrained Transformers}

% \title{NVIB for Pretrained Transformers}
% \title{A Variational Bayesian Interpretation  \\ of Pretrained Transformers}
% \title{A Bayesian Reinterpretation  \\ of Pretrained Transformers}
% \title{Bayesian Reinterpretation and Regularisation of Pretrained Transformers}
% \title{An Information-Theoretic Perspective \\ of Pretrained Transformers}
% \title{Nonparametric Variational Post-Training Regularisation  \\ of Pretrained Transformers}
% \title{Nonparametric Variational Regularisation \\ of Transformers Post-Pretraining}
% \title{Nonparametric Variational Regularisation \\ of Pretrained Transformers}
% Nonparametric Variational Fine-tuning TBA
% \title{A Bayesian Boost \\ of Pretrained Transformers}
% \title{A Bayesian Reinterpretation of Pretrained Transformers with Post-Training Regularisation}
% \title{Post-hoc regularisation with Nonparametric Variational Information Bottleneck}
% \title{Post-hoc regularisation for pretrained Transformers using NVIB}

% The \author macro works with any number of authors. There are two commands
% used to separate the names and addresses of multiple authors: \And and \AND.
%
% Using \And between authors leaves it to LaTeX to determine where to break the
% lines. Using \AND forces a line break at that point. So, if LaTeX puts 3 of 4
% authors names on the first line, and the last on the second line, try using
% \AND instead of \And before the third author name.

\author{\name Fabio Fehr \email fabio.fehr@idiap.ch \\
       \addr Idiap Research Institute\\
       Ecole Polytechnique F\'ed\'erale de Lausanne\\
       Switzerland
       \AND
       \name James Henderson \email james.henderson@idiap.ch \\
       \addr Idiap Research Institute\\
       Switzerland}

\editor{TBA}

\maketitle
% \author{%Nonparametric
%   % David S.~Hippocampus
%   % \thanks{Use footnote for providing further information about author (webpage, alternative address)---\emph{not} for acknowledging funding agencies.} \\
%   Fabio Fehr\textsuperscript{1,2} \ \ \ \ \  James Henderson\textsuperscript{1} \\ 
%   \textsuperscript{1}Idiap Research Institute, Switzerland\\
%   \textsuperscript{2}Ecole Polytechnique F\'ed\'erale de Lausanne, Switzerland\\
%   \texttt{firstname.lastname@idiap.ch} \\
%   % examples of more authors
%   % Coauthor \\
%   % Affiliation \\
%   % Address \\
%   % \texttt{email} \\
%   % \AND
%   % Coauthor \\
%   % Affiliation \\
%   % Address \\
%   % \texttt{email} \\
%   % \And
%   % Coauthor \\
%   % Affiliation \\
%   % Address \\
%   % \texttt{email} \\
%   % \And
%   % Coauthor \\
%   % Affiliation \\
%   % Address \\
%   % \texttt{email} \\
% }

\maketitle

\begin{abstract}
  % The abstract paragraph should be indented \nicefrac{1}{2}~inch (3~picas) on
  % both the left- and right-hand margins. Use 10~point type, with a vertical
  % spacing (leading) of 11~points.  The word \textbf{Abstract} must be centered,
  % bold, and in point size 12. Two line spaces precede the abstract. The abstract
  % must be limited to one paragraph.

  % Introduce Large Pretrained and fine tuned models and they use attention. Powerful and important and here to stay
  The current paradigm of large-scale pre-training and fine-tuning Transformer large language models has lead to significant improvements across the board in natural language processing.
  % Problem - (1) Scaling: more parameters means overfitting and less generalisation (need for regularisation) and (2) Adaption: Typically we just fine-tune. What if we cannot afford to finetune and update models given a domain shift (or small amounts of data?). 
    However, 
    %continually scaling to larger models leads to challenges of overfitting and domain adaption. As the number of parameters increase these models can become over-specified to their task. 
    such large models are susceptible to overfitting to their training data, and 
    as a result the models perform poorly when the domain changes. Also, due to the model's scale, the cost of fine-tuning the model to the new domain is large.
  Nonparametric Variational Information Bottleneck (NVIB) has been proposed as a regulariser for training cross-attention in Transformers, potentially addressing the overfitting problem.
  % Model - NVIB is an information theoretic regulariser of attention functions.
  % Contributions - (1) Extending NVIB for multihead attention which results in applications to self, causal and cross-attention. 
  We extend the NVIB framework to replace all types of attention functions in Transformers, and 
  % (2) Identity Initialisation allows the NVIB model and denoising attention to be connected to pretrained models. 
  show that existing pretrained Transformers can be reinterpreted as Nonparametric Variational (NV) models using a proposed identity initialisation.  
  % (3) Changing the initialisation allows for post-hoc regularisation without fine-tuning.
  We then show that changing the initialisation introduces a novel, information-theoretic post-training regularisation in the attention mechanism, which improves out-of-domain generalisation without any 
  training. 
  % update of the original model weights or backward passes.  
  % (4) Changing the prior to an empirical (informed) prior allows for domain adaption. Changing the initialisation introduces the information-theoretic regularisation without the need for fine-tuning, addressing the challenge of overfitting.
  %Moreover, we show that changing the NVIB prior allows for a novel form of domain adaption without the need for any fine-tuning. 
  This success supports the hypothesis that pretrained Transformers are implicitly NV Bayesian models.
  
% Results + claims
% (1) We claim that NVIB can be connected to pretrained models and achieve an exact equivalence.
    %  We show this with certain initialisation we get the same results.

% (2) We claim that NVIB can be used to regularise attention based models post-hoc 
    % We show the plots of changing initialisations. We also show that there exists models that are potentially better even in domain.
% (3) We claim that the empirical prior can be used as a form of domain adaption
    % We show that when shifting the domain (not task) with a fine tuned model we can get improvements.
    
  % Applications of NVIB and denoising cross-attention to pretrained sequence-to-sequence models. Advantages: Improved downstream task generalisation, regularisation and sparsity of large language models and new applications such as interpolation. Makes the claim that Transformers are inherently using Nonparametric Variation Information Bottleneck.
\end{abstract}

% \fabio{Story Pretrained models are good - NVIB now works with this.  }

% \fabio{Why should people care? - Get minor improvements for free, understand more about transformers. Better generalisation. General method}

\section{Introduction}

% Background - Pretraining + NVIB
Self-supervised pretraining of Transformer models  \citep{devlin-etal-2019-bert, lewis-etal-2020-bart, raffel2020-t5,zhang2022opt} has been hugely successful, improving performance in virtually every NLP task.  This indicates that the inductive bias of attention-based representations is extremely effective for language, but it is not clear which characteristics of Transformers are essential to this inductive bias and which are implementation details.  
We shed light on this issue by developing a variational Bayesian reinterpretation of 
%the representations learned by 
pretrained Transformers, which more explicitly characterises how information about the real data distribution is represented.  
% and by showing that this reinterpretation accurately characterises the pretrained models' information about the real data distribution.  

In previous work, \citet{hendersonfehr2023} derive a variational Bayesian generalisation of the cross-attention layer of Transformers, called Nonparametric Variational Information Bottleneck (NVIB). NVIB provides an information-theoretic sparsity-inducing regulariser over attention-based representations when used to train an attention-based model.  In this paper, we investigate the possibility that NVIB also provides an accurate theoretical model of existing attention-based models which have been pretrained without NVIB regularisation. 
We extend NVIB to all the uses of attention in Transformers (multi-head, cross- and self- attention, in encoders and decoders), and propose a method for converting a pretrained Transformer into the weights of an equivalent Nonparametric Variational (NV) Bayesian model.  

% Reinterpretation - Why is this good?
% This reinterpretation of pretrained Transformers as an equivalent Bayesian model allows us to consider uncertainty and incorporating a prior distribution over Transformer embeddings.
This equivalent NV-Transformer embeds text into a probability distribution which adds uncertainty to the representation computed by the pretrained Transformer.
% Claim of equivalence
Theoretically, the exact equivalence only occurs when this uncertainty is exactly zero, but empirically there is a practical range of non-zero uncertainty levels where the model's predictions are unchanged.  Continuing training of this initial model with NVIB regularisation would push this uncertainty higher, but training large models is computationally expensive and requires significant amounts of in-domain data. 
% Claim smooth adjustment of the regularisation
Instead, we only adjust hyperparameters of the initialisation to better satisfy the NVIB regulariser, without any backward-passes or paramter updates.  Increasing the uncertainty level does change the model's predictions, but interestingly it does not degrade the model's performance.  Instead it adds a form of post-training regularisation, which accesses a space of models with different predictions with comparable performance.
% Claim OOD generalisation
These model's can often improve over the original Transformer in out-of-domain generalisation. 
%We argue that this indicates that NVIB's model of the information content of a representation (versus NVIB's model of noise) is the same as the real information content of a pretrained Transformer's representations.  
In addition, this successful Bayesian reinterpretation of pretrained Transformers provides insights into how Transformers represent information.

\begin{figure}[t]
    \centering
    \includegraphics[width=0.55\textwidth]{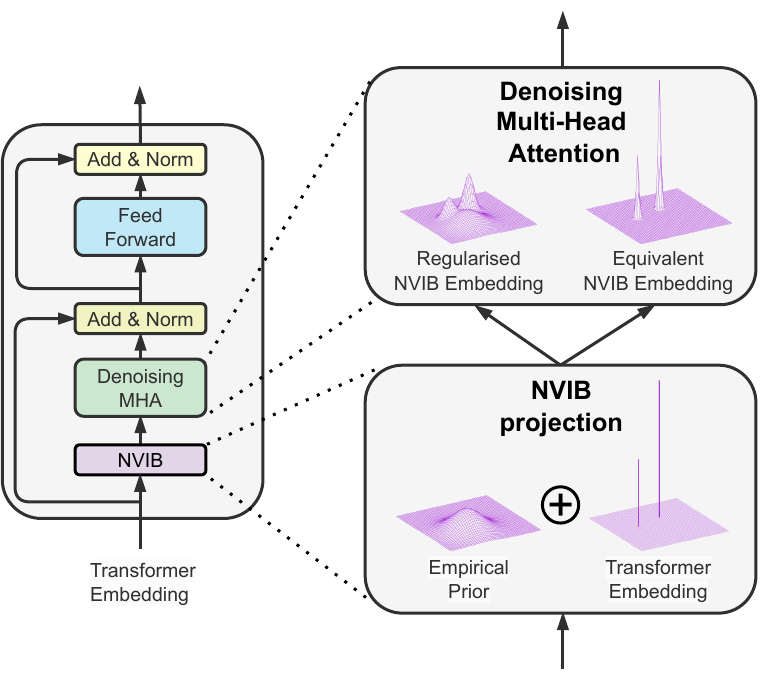}
    \caption{\textbf{Left:} Reinterpreted Transformer encoder layer with NVIB and denoising multi-head attention (MHA). \textbf{Right:} A NVIB projection layer which combines the empirical prior with the Transformer embeddings and produces a representation for denoising MHA for either post-training regularisation or exact equivalence.   
    }
    \label{fig:architecture}
\end{figure}

\paragraph{Contributions} 
In this work we propose multiple novel methods and provide empirical insights which contribute to representation learning. 
% Math in appendix
We extend NVIB regularisation beyond single-head cross-attention to all uses of attention in a Transformer: multi-head attention, encoder self-attention, and decoder causal self-attention, resulting in our proposed NV-Transformer architecture. 
% Empirical prior - data efficient
We define a novel empirical prior distribution estimated from a model's embeddings given a marginal amount of empirical data.  
% Equivalent model model with identity initialisation
We contribute a reinterpretation of pretrained Transformers as a variational Bayesian model by defining an equivalent initialisation of NV-Transformers.  
% Post-hoc regularisation
We propose a method to regularise a pretrained Transformer without re-training by adding uncertainty to the embeddings of this equivalent NV-Transformer.
% OOD improvements
Empirically, we show the usefulness of these proposals by smoothly varying the amount of this post-training regularisation and achieving improved performance in out-of-domain text summarisation.

\section{Nonparametric Variational Information Bottleneck}

% We first summarise NVIB and related work. 

% \subsection{Nonparametric Variational Information Bottleneck} 
% \fabio{In this section we need to explain original NVIB}

\citet{hendersonfehr2023} propose NVIB as an information-theoretic regulariser for attention layers.  It extends Variational Information Bottleneck \citep{alemi2017deep} to the unboundedly large representations supported by attention. The NVIB regulariser is used to define a Variational Auto-Encoder \citep{kingma2014autoencoding} for Transformers with NVIB applied to its cross-attention layer. \citet{behjati2023learning} apply NVIB to stacked Transformer self-attention layers in the encoder, which when trained learns an information-theoretic abstraction of the representations through the model.
% illustrated in Figure~\ref{fig:NVIB}.

% \begin{figure}
%     \centering
%     \includegraphics[width=0.8\textwidth]{figures/nvae_v5.2.png}
%     \caption{A Transformer Variational Auto-Encoder with NVIB regularisation.  Since in this paper we do not do any training, only the inference and the testtime denoising attention functions are used, without the sampling.}
%     \label{fig:NVIB}
% \end{figure}

\subsection{Latent Mixture Distributions} 
\label{sec:latent_mixtures}

A key insight of \citet{hendersonfehr2023} is that attention-based representations can be interpreted as nonparametric mixtures distributions.  The set of vectors which are accessed by attention are interpreted as specifying a mixture of impulse distributions.
The attention function can then be interpreted as Bayesian query denoising using this distribution as the prior.  The query is interpreted as being corrupted by Gaussian noise, 
%the prior's impulses are located at each vector with a mixture weight determined by its $L2$ norm, 
the attention weights are the posterior's mixture weights given the query observation, and the output of the attention function is the expected value of this posterior distribution. 
 % (See below for details.)  
 They then generalise the attention function to query denoising with any mixture distribution as the representation being accessed, called \textit{denoising attention}.  
 % Since the number of vectors in a Transformer embedding grows with the length of the input text, the space of mixture distributions needs to include arbitrarily large mixtures, and thus be nonparametric. Thus, in this generalised model, the latent space is the space of nonparametric mixture distributions.
 Since a Transformer embedding is a set of vectors which grows in size to accommodate the complexity of the input text, the latent space of equivalent mixture distributions is nonparametric in nature.

\paragraph{Denoising attention}
To show that the denoising attention function is a generalisation of the standard attention function, \citet{hendersonfehr2023} provide a constructive proof of exact equivalence.
The standard attention function projects the accessed set of vectors $\boldsymbol{Z} \in \mathbb{R}^{n \times d}$ via weight matrices $\boldsymbol{W}^K, \boldsymbol{W}^V \in \mathbb{R}^{d \times d}$ to keys and values, respectively, and projects the accessing input vector $\boldsymbol{u^\prime} \in \mathbb{R}^{1 \times d}$ via the weight matrix $\boldsymbol{W}^Q \in \mathbb{R}^{d \times d}$ to a query.  It uses the keys' dimensionality $d$ for scaling.
This scaled dot product attention function can be regrouped into a core dot product attention function $\text{Attn}(\boldsymbol{u},\boldsymbol{Z})$ in which all operations are done in the space of $\boldsymbol{Z}$.
\begin{align*}
\text{Attention}(\boldsymbol{u^\prime}, \boldsymbol{Z} \;;~ \boldsymbol{W}^Q,\boldsymbol{W}^K,\boldsymbol{W}^V) =  \text{Attn}(\boldsymbol{u^\prime} \boldsymbol{W}^Q (\boldsymbol{W}^K)^\top,~ \boldsymbol{Z})~ \boldsymbol{W}^V = \text{Attn}(\boldsymbol{u},\boldsymbol{Z})~ \boldsymbol{W}^V
\nonumber
\end{align*}
where $\boldsymbol{u} = (\boldsymbol{u}^\prime \boldsymbol{W}^Q (\boldsymbol{W}^K)^\top) \in \mathbb{R}^{1\times d}$. The function $\Attn(\boldsymbol{u},\boldsymbol{Z})$ can then be defined both in terms of a sum over the vectors $\boldsymbol{z}_i$ in $\boldsymbol{Z}$, or in terms of an integral over a distribution which is only nonzero at the $\boldsymbol{z}_i$:
%
%\vspace{-2ex}
\begin{align}
  \label{eq:attn}
  \text{Attn}(\boldsymbol{u}, \boldsymbol{Z})
  &= \text{softmax}\left( \tfrac{1}{\sqrt{d}} \boldsymbol{u} \boldsymbol{Z}^\top \right) \boldsymbol{Z} = \text{DAttn}(\boldsymbol{u}\;;~ F_{\boldsymbol{Z}})
  \\
  \label{eqn:probZ}
  F_{\boldsymbol{Z}} &= \sum_{i=1}^n ~\frac{\exp( \tfrac{1}{2\sqrt{d}}||\boldsymbol{z}_i||^2 )}{\sum_{i=1}^n \exp( \tfrac{1}{2\sqrt{d}}||\boldsymbol{z}_i||^2 )}~ \delta_{\boldsymbol{z}_i}
  \\
  \label{eq:dattn}
  \text{DAttn}(\boldsymbol{u};~ F) &=
  \int_{\boldsymbol{v}} ~\frac{ f(\boldsymbol{v}) ~g(\boldsymbol{u};~ \boldsymbol{v},\sqrt{d}\boldsymbol{I})
  }{ \int_{\boldsymbol{v}}  f(\boldsymbol{v}) ~g(\boldsymbol{u};~ \boldsymbol{v},\sqrt{d}\boldsymbol{I}) ~d\boldsymbol{v}
  }~ \boldsymbol{v} ~d\boldsymbol{v}
\end{align}
where $\delta_{\boldsymbol{z}_i}$ is an impulse distribution at $\boldsymbol{z}_i$,
$f(\cdot)$ is the probability density function for distribution $F$, and $g(\boldsymbol{u};~ \boldsymbol{v},\sqrt{d}\boldsymbol{I})$ is the multivariate Gaussian function with diagonal variance of $\sqrt{d}$.  $\text{DAttn}(\boldsymbol{u};~ F_{\boldsymbol{Z}})$ can be interpreted as Bayesian query denoising where $F_{\boldsymbol{Z}}$ is the prior and $g(\boldsymbol{u};~ \boldsymbol{v},\sqrt{d}\boldsymbol{I})$ is a noisy observation.

This construction shows that any Transformer embedding $\boldsymbol{Z}$ has an equivalent mixture of impulse distributions, namely $F_{\boldsymbol{Z}}$, where denoising attention $\text{DAttn}(\boldsymbol{u};~ F_{\boldsymbol{Z}})$  gives us exactly the same result as attention $\text{Attn}(\boldsymbol{u}, \boldsymbol{Z})$, for all queries $\boldsymbol{u}$.

% \paragraph{Denoising attention at training time}
% During training, the set of vectors $\boldsymbol{Z} \in \mathbb{R}^{n \times p}$ and their log-probability values $\text{log}(\boldsymbol{\pi}) \in \mathbb{R}^{1 \times n}$ are both sampled and output by the NVIB layer, thereby specifying the sampled mixture distribution $F$. 
% For each use of denoising attention, the query $\boldsymbol{u^\prime} \in \mathbb{R}^{1 \times p}$ is projected by the grouped matrices $\boldsymbol{W}^Q, \boldsymbol{W}^K \in \mathbb{R}^{p \times d}$ to $\boldsymbol{u} = (\boldsymbol{u}^\prime \boldsymbol{W}^Q (\boldsymbol{W}^K)^\top)$.  The keys' dimensionality $d$ is used for scaling.  Denoising attention can then be computed as:
% \begin{align}
%   \text{DAttn}(\boldsymbol{u}\;;\; F) 
%   &= \text{softmax}\left( \tfrac{1}{\sqrt{d}}\boldsymbol{u}{\boldsymbol{Z}}^\top +\log(\boldsymbol{\pi}) -\tfrac{1}{2\sqrt{d}}\|\boldsymbol{Z}\|^2 \right) \boldsymbol{Z}
%   \nonumber
% \end{align}
% We define this for $\boldsymbol{u} \in \mathbb{R}^{1 \times p}$, but this can easily be extended to multiple queries.

\subsection{Distributions over Mixture Distributions}

The generalisation of attention-based representations to nonparametric mixture distributions allows Bayesian nonparametrics to be used to define distributions over the latent space.
% , which can then be used to define a Bayesian model.
\citet{hendersonfehr2023} propose to use Dirichlet processes (DPs) to define distributions over mixture distributions.  Given a prior DP and an input text, NVIB computes a DP representation of that text using exact inference of the posterior from a set of pseudo-observations output by the Transformer encoder, taking advantage of the fact that DPs are conjugate priors.  

This posterior ``$q$'' DP is specified by a base distribution $G^q_0$ for generating vectors and a pseudo-count $\alpha^q_0$ for generating mixture weights.  The mixture distribution $G^q_0$ consists of one component for the prior distributions plus one Gaussian component for each vector input to the NVIB layer from its Transformer encoder.
These vectors are mapped to the parameters $(\boldsymbol{\mu}^q, \boldsymbol{\sigma}^q, \boldsymbol{\alpha}^q)$ of the DP posterior's pseudo-count $\alpha^q_0 = \sum_i {\alpha}^q_i$ and base distribution 
$G^q_0 = \sum_i \frac{{\alpha}^q_i}{\alpha^q_0} \mathcal{N}(\boldsymbol{\mu}^q_i,\boldsymbol{I}(\boldsymbol{\sigma}^q_i)^2)$.

\paragraph{Denoising attention during evaluation}
During evaluation, \citet{hendersonfehr2023} do not sample a mixture distribution $F$ from the DP, but instead use the mean of the posterior DP distribution, which is its base distribution $G^q_0$.  Denoising attention applied to $G^q_0$ can be computed simply because it is a mixture of Gaussians.  It differs from standard attention in that the component weights $\frac{\boldsymbol{\alpha}^q}{\alpha^q_0}$ replace the key biases implicit in the query-key dot product, and an interpolation between queries and values replaces the values in the attention-weighted average.
For convenience let $(\boldsymbol{\sigma}^r_i)^2=(\sqrt{d}+(\boldsymbol{\sigma}^q_i)^2)$.
Evaluation denoising attention can then be computed as:
% \begin{align*}
%  \text{DAttn}(\boldsymbol{u}\;;\; G^q_0) \\
% &\!\!{=}\!\!\!& \text{softmax}\left(\!\! \boldsymbol{u} \left(\frac{\boldsymbol{\mu}^q}{(\boldsymbol{\sigma}^r)^2}\!\right)^{\!\!T} 
% % 
% + \text{log}(\frac{\boldsymbol{\alpha}^q}{\alpha^q_0})
% % 
%     -\left(\!\tfrac{1}{2}\left\|\frac{\boldsymbol{\mu}^q}{\boldsymbol{\sigma}^r}\right\|^2\right)^{\!\!T}
%     % 
%     -\boldsymbol{1}_p\left(\text{log}(\boldsymbol{\sigma}^r)\right)^\top
%     \!\right)
%     % 
%     \left( \frac{(\boldsymbol{\sigma}^q)^{2}}{(\boldsymbol{\sigma}^r)^2}\odot(\boldsymbol{1}_n^\top\boldsymbol{u}) 
%     % 
%     +\frac{\sqrt{d}}{(\boldsymbol{\sigma}^r)^2}\odot\boldsymbol{\mu}^q \!\right)
% \end{align*}
\begin{align}
 \text{DAttn}(\boldsymbol{u}; G^q_0) &=  \text{softmax}\left(\boldsymbol{u} \bigg(\frac{\boldsymbol{\mu}^q}{(\boldsymbol{\sigma}^r)^2}\right)^\top
 \underbrace{ \addstackgap[10pt] + \; \boldsymbol{b}}_{\text{Bias}}
    \bigg)
    \bigg( \underbrace{\frac{(\boldsymbol{\sigma}^q)^{2}}{(\boldsymbol{\sigma}^r)^2}\odot(\boldsymbol{1}_n^\top\boldsymbol{u}) 
    +\frac{\sqrt{d}}{(\boldsymbol{\sigma}^r)^2}\odot\boldsymbol{\mu}^q}_{\text{Query-Value interpolation}} \bigg) 
    \label{eqn:evalDAttn}
    \\
    % \text{where} \\
    \boldsymbol{b} &= \text{log}(\frac{\boldsymbol{\alpha}^q}{\alpha^q_0})
    -\left(\!\tfrac{1}{2}\left\|\frac{\boldsymbol{\mu}^q}{\boldsymbol{\sigma}^r}\right\|^2\right)^{\top}
    -\boldsymbol{1}_d\left(\text{log}(\boldsymbol{\sigma}^r)\right)^\top \label{eqn:evalDAttn_bias}
\end{align}
where $\boldsymbol{1}_d$ is a row vector of $d$ ones. When the attention bias term $\boldsymbol{b}$ is large across the non-prior components and thus down-weights the prior component, we can have an identical softmax to attention. When the posterior variance $(\boldsymbol{\sigma}^q)^{2}$ is approximately zero, we get no interpolation between the query and value and get an identical value computation to attention. \citet{hendersonfehr2023} propose this formulation of attention only for the case of single-head cross-attention during evaluation.

\subsection{Related Work}

% Sparsity + Quantisation
Related work in post-training regularisation such as quantisation \citep{dettmers_8bit_2022, yao_quantisation2022, xiao2023smoothquant,frantar-gptq} and sparsity \citep{Hubara2021, frantar-sparsegpt, frantar2022obc} focus on regularising the the model weights and not the latent representation. Similar to \citet{frantar-sparsegpt}, who propose a data-driven way to sparsify a model in one-shot without any retraining, we propose a data-driven form of soft sparsity for attention, without any retraining.
% and both techniques use data to refine the model. 
In contrast, our method's regularisation uses information theory to regularise post-training. 

% DOmain generalisation 
% We learn a domain specific prior to adapt to the new domain more similar to 

% Variational OOD generalisation
Our work shares similarity to the out-of-distribution generalisation literature in learning variational models with invariant features \citep{ilse2020diva} and parameter sharing in the embedding space to adapt to domain-shifts \citep{MuaFukSriSch2017, Li2017DeeperBA, Blanchard2021}. We take a Bayesian approach and define our prior distribution based on empirical dataset statistics which allows for properties which are robust to domain shift.
% Pretrained LLMs - VAE with fine tuning
\cite{park-lee-2021-finetuning} proposes a way to define pretrained language models as variational models by fine-tuning. Our work, much like low-rank adaptions \citep{hu2021lora} does not fine-tune or update the original model weights. Similar to adapter modules \citep{pmlr-v97-houlsby19a} we include a NVIB layer and keep the rest of the model weights frozen. Our initialisation of this layer reinterprets the model as a variational Bayesian model.

\section{Pretrained Transformers as Nonparametric Variational Models}
\label{sec:pretrained}

% Roadmap:
% Application to cross, self and causal attention and mulithead.
% Exact equivalence contribution with initialisation 
% Empirical prior

We propose a method for reinterpreting pretrained Transformers as nonparametric variational Bayesian models. To support this construction, we contribute the following extensions to NVIB: adaption to multihead denoising attention (Appendix \ref{sec:multiDattn}); adaptation to encoder denoising self-attention (Appendix \ref{sec:selfDattn}); and adaptation to decoder denoising causal attention (Appendix \ref{sec:causalDattn}).  This allows us to apply NVIB to every form of attention used in standard Transformers, resulting in our proposed NV-Transformer reinterpretation.
% , which we call NVIB Transformers.

% Identity Initialisation
Using these extensions, we propose an identity initialisation which allows us to achieve an equivalence with the standard attention mechanisms in Transformers (Section \ref{sec:pretrained_identity_initialisation}). 
% Empirical prior
We then define a novel empirical prior distribution which introduces uncertainty in the latent representations in a way which captures the implicit uncertainty in pretrained Transformers
%the prior distribution given the pretrained model as an empirical estimation over our data 
(Section \ref{sec:empirical_prior}).
In the following sections, we evaluate this reinterpretation of pretrained Transformers as the nonparametric variational Bayesian model, NV-Transformers.

% In the next section, we evaluate the equivalence of these models, and some extensions made possible by this reinterpretation.

% \subsection{Denoising is All You Need}

% To reinterpret pretrained Transformers as nonparametric models, we first need to extend NVIB to all the uses of attention in these Transformers.

% ...
% A caveat of this derivation is that it applies only for single-head attention and is not trivial to extend for multi-head attention. ADD IN MULTIHEAD and mention the IMPLICIT GRADIENTS reparameterisation trick here. Also the EXP activation function.

% \fabio{Include how we would do mulithead attention, self-attention and causal attention.}

% \paragraph{Improved NVIB implementation}
% Following from the work of \cite{hendersonfehr2023}, who propose Nonparametric Variational Information Bottleneck (NVIB) for attention layers and apply it to cross-attention in Transformers, we make implementation improvements which allow for an exact equivalence with pretrained models and improved stablility in training. For the equivalence with pretrained models we use an implementation of NVIB compatible with multihead attention and use exponential activation functions. 

\subsection{Pretrained Identity Initialisation}
\label{sec:pretrained_identity_initialisation}

We define an identity initialisation for NVIB such that denoising attention is equivalent to standard attention. The construction in Section~\ref{sec:latent_mixtures} was previously used to show that denoising attention subsumes standard attention.
% Here we use it to convert a standard attention layer into an NVIB layer.  
Given a set of vectors $\boldsymbol{Z}$ input to the attention layer, we convert it to the parameters $(\boldsymbol{\mu}^q, \boldsymbol{\sigma}^q, \boldsymbol{\alpha}^q)$
%\frac{\boldsymbol{\alpha}^q}{\alpha^q_0}
of a posterior DP distribution.  Then we apply the evaluation denoising attention function (Equation \ref{eqn:evalDAttn}) to the resulting base distribution $G^q_0$.  We want this function to be equivalent to standard attention over $\boldsymbol{Z}$.

Our proposed NVIB layer takes the set of vectors $\boldsymbol{Z}$ computed by the previous layer and projects it to the DP parameters $(\boldsymbol{\mu}^q, \boldsymbol{\sigma}^q, \frac{\boldsymbol{\alpha}^q}{\alpha^q_0})$ (excluding the prior component) as follows:
\begin{align}
    \boldsymbol{\mu} \ &= \ \mu(\boldsymbol{Z}) \ = \boldsymbol{Z} \boldsymbol{W}^{\mu} + \boldsymbol{b}^{\mu} \\
    \boldsymbol{\sigma}^2 &= \sigma^2(\boldsymbol{Z}) = \text{exp}(\boldsymbol{Z} \boldsymbol{W}^{\sigma} + \boldsymbol{b}^{\sigma}) \\
    \boldsymbol{\alpha} \ &= \ \alpha(\boldsymbol{Z}) \ =  \text{exp}(\boldsymbol{Z}^2 \boldsymbol{w}_1^{\alpha} + \boldsymbol{Z} \boldsymbol{w}_2^{\alpha} + b^{\alpha})
\end{align}
To complete our model specification we need to set the parameters of this mapping, $\boldsymbol{W}^{\mu}, \boldsymbol{W}^{\sigma} \in \mathbb{R}^{d \times d}$, $\boldsymbol{b}^{\mu}, \boldsymbol{b}^{\sigma}\in \mathbb{R}^{d}$, $\boldsymbol{w}_1^{\alpha}, \boldsymbol{w}_2^{\alpha} \in \mathbb{R}^{d}$ and $b^{\alpha} \in \mathbb{R}$.

%The constructive proof described above suggests a way to map the output of a pretrained Transformer encoder into a specification of a mixture of impulse distributions.  Based on this mapping, we develop a method for mapping the output of a pretrained Transformer encoder into a specification of a DP distribution over these mixtures.  This requires proposing a specification of the uncertainty in the DP's distribution, which we develop with the goal of modelling the uncertainty implicit in the pretrained Transformer.  We evaluate this property in Section~\ref{sec:eval}.   Key to this model of uncertainty is the use of an empirical distribution over latent representations to define the DP prior.

For equivalence between standard attention and denoising attention we must remove the influence of the prior component and reduce the uncertainty around the other components of the mixture. See Figure \ref{fig:architecture} for a visual depiction.  Conceptually, the mixture of Gaussians $G^q_0$ can be equivalent to the mixture of impulse distributions $F_{\boldsymbol{Z}}$ in Equation~\ref{eqn:probZ}, if the weight of the prior component is very small and the other Gaussian components have so little uncertainty that they behave like the impulses of $F_{\boldsymbol{Z}}$.  To achieve this, 
%To converting pretrained Transformers into nonparametric variational Bayesian model we initialise the NVIB parameters so that the base mixture distribution is equivalent to standard attention. This is achieved by initialising: the mean mixture distribution to with the weights of the pretrained model; the uncertainty around that mean to be very small; and the pseudo-count weights both be large relative to the prior and form a constant in the attention bias (Equation \ref{eqn:evalDAttn_bias}). We propose the following intialisation for the NVIB parameter projections:
%
the mean projection $\mu(.)$ can simply be the identity, and the variance projection $\sigma^2(.)$ can map to approximately zero values (exactly zero is not possible with the exponential activation function).  For the pseudo-count projection $\alpha(.)$, the use of a log-quadratic projection allows the psuedo counts to be proportional to L2-norm term in Equation~\ref{eqn:probZ}.  In addition, we would like these non-prior-component pseudo-counts to be large to down-weight the influence of the prior component.

%% The [] shows how many lines you want to wrap
%\begin{wrapfigure}[12]{r}{0.45\textwidth}
%  \centering
%  %\vspace{-10pt}  % Controls up and down of the figure
%    %\raisebox{3ex}{
%   \includegraphics[width=0.45\textwidth]{figures/posthoc-reg_v1.png}
%   %}
%   %\includegraphics[width=0.2\textwidth]{figures/pretrainedNVIB-pretrained_only.pdf}
%  \caption{Illustration of converting a Transformer embedding into an NVIB embedding, either for exact equivalence or with post-hoc regularisation.}
%  \label{fig:posthoc}
%\end{wrapfigure}
We define initialisation hyperparameters which allows to control the transition between equivalence and a smooth regularisation of embeddings (illustrated in Figure~\ref{fig:architecture}). The hyperparameters $\tau_{\alpha}^i$ 
 and $\tau_{\sigma}^i$ control the level of uncertainty of the mixture distributions by changing the initialisation of the pseudo-counts and variance, respectively. The indicator $i$ is for the different sections of the model which allows for independent control of the encoder's self-attention $(e)$, decoder's cross-attention $(c)$ and decoder's causal self-attention $(d)$. 
% To allow a range of models which include exact equivalence but also allow us to increase the uncertainty in the mixture distributions as a form of regularisation (illustrated in Figure~\ref{fig:architecture}), 
% we define the initialisation weights in terms of initialisation hyperparameters, $\tau_{\alpha}^i$ 
 % and $\tau_{\sigma}^i$, which control the level of regularisation in different sections of the model where $i$ represents the encoder's self-attention $(e)$, decoder cross-attention $(c)$ and decoder causal self-attention $(d)$, independently. 
 Empirically, this allows for more flexibility than a single hyperparameter and practically easier to tune than defining a hyperparameter per layer. In general we define the layer projection weights as follows:
%
% To initialise the NVIB layer to an identity we define the weight matrices
%
% of the projections to the mean $\boldsymbol{W}^\mu = \boldsymbol{I}$, $\boldsymbol{b}^\mu = \boldsymbol{0}$. 
% % 
% The projection to the posterior variance
% $\boldsymbol{W}^\sigma = \boldsymbol{0}$, $\boldsymbol{b}^\sigma = \text{log}(\sigma^2_{init}) \odot \boldsymbol{1}$, where $\sigma_{init} = (\boldsymbol{\sigma}^p \tau_{\sigma} + 1e^{-14})$ is set to be very small. The psuedo-count parameters projection are quadratic to get an equivalence with the squared norm term in denoising attention 
% $\boldsymbol{w}_1^\alpha = \frac{1}{2\sqrt{d}} \odot \boldsymbol{1}$, $\boldsymbol{w}_2^\alpha = \boldsymbol{0}$, $\boldsymbol{b}^\alpha = \log(\alpha_{init})$ where $\alpha_{init} = 1e^{14}$ is a scaling factor which affects the total $\alpha_0$ which we intialise to be large to ingore all noise even when sampling at training time.
%
% \vspace{-0.73cm} % If no space then it breaks the minipage.
\centerline{
\begin{minipage}[t]{.3\textwidth}
    \begin{align*}
        \boldsymbol{W}^{\mu} &= \boldsymbol{I} \\ 
        \boldsymbol{W}^{\sigma} &= \boldsymbol{0} 
    \end{align*}
\end{minipage}
\begin{minipage}[t]{.3\textwidth}
    \begin{align*}
        \boldsymbol{w}^{\alpha}_1 &= \frac{1}{2\sqrt{d/h}} \odot \boldsymbol{1} \\
        \boldsymbol{w}^{\alpha}_2 &= \boldsymbol{0}
    \end{align*}
\end{minipage}
% \vfill
\begin{minipage}[t]{.3\textwidth}
    \begin{align*}
        \boldsymbol{b}^{\mu} &= \boldsymbol{0} \\ 
        \boldsymbol{b}^{\sigma} &= \text{log}((\boldsymbol{\sigma}^p * \tau_{\sigma}^i)^2 ) \\ 
        {b}^{\alpha} &= \epsilon^\alpha \tau_{\alpha}^i \\
    \end{align*}
\end{minipage}
% \vspace{-4ex}
}
where $d$ and $h$ denote the model projection size and number of attention heads.  As discussed in the next section, the empirical distribution is used to define the prior standard deviation $\boldsymbol{\sigma}^p$, and the constant $\epsilon^\alpha$, which denotes the empirical standard deviation of the scaled L2-norm in log space per layer.

% Alpha parameter
The hyperparameter $\tau_\alpha^i$ controls the relative weight of the prior component to the non-prior components in the mixture distribution $G^q_0$.
Since $\epsilon^\alpha$ is a standard deviation, it reflects the normal range of values for the non-prior-component pseudo counts, so we can use $\tau_\alpha^i$ to control the weight given to the prior component with respect to this range. 
When $\tau_\alpha^i = 0$, the non-prior pseudo counts are their scaled L2-norms and the prior pseudo count is the the expected scaled L2-norm. When we increase or decrease the $\tau_\alpha^i$ it adjusts the magnitude of the non-prior L2-norms proportionately, which relatively decreases or increases the weight on the prior proportionately to the standard deviation of the scaled L2-norm. 
% We use the standard deviation of the L2-norms  $\epsilon^\alpha$ to shift the magnitude in a meaningful way. 

% Sigma parameter
The hyperparameter $\tau_{\sigma}$ controls the interpolation between the query and value (Equation \ref{eqn:evalDAttn}), which we set proportionately to the variance of the prior distribution.
%We define $\tau_{\sigma}$ to control the standard deviation of the prior, which
When $\tau_{\sigma}\approx0$, there is effectively no interpolation, as with standard attention. When $\tau_{\sigma}=1$ this can be interpreted as increasing the uncertainty to the level of the empirical prior distribution.

\subsection{Empirical Priors} \label{sec:empirical_prior}

% The prior is a DP and it makes sense to use a empirical data
The prior distribution is a Dirichlet Process which treats the latent representations as sampled mixture of impulse distributions coming from the base distribution of the prior. Therefore, we can estimate the base distribution of the prior empirically by observing the distribution of vectors given forward passes of the model. 

% what we know prior to seeing the input text, as required by our Bayesian model.
Taking a Bayesian approach, the NV-Transformer's prior should reflect the distribution over vectors which the pretrained Transformer knows before seeing the input text.  This is the distribution observed during training.  We don't have the original pretraining corpus, but we do have the corpora which our pretrained Transformers were fine-tuned on. Therefore, we compute statistics from those corpora and use them to define our priors.  We estimate the prior as the best fit of an isotropic Gaussian distribution to the empirical distribution over latent vectors computed when embedding this training corpus.

% Because the $\boldsymbol{\alpha}$ parameters computed above need to be large to get exact equivalence, the weight given to the prior component in the mixture $G^q_0$ is small, which makes it ignored in the softmax normalisation.  But we nonetheless want the prior to reflect what we know prior to seeing the input text, as required by our Bayesian model.  This becomes more important when we relax the exact equivalence requirement and regularise the latent representations towards the prior.

% The prior distribution is a DP, which is an exchangeable distribution, which treats each vector in the sampled mixture of impulse distributions as coming from the same base distribution of the prior. We can therefore estimate the base distribution of the prior empirically by observing the distribution of vectors.  

\paragraph{Data-informed prior}
%Given the Bounded Dirichlet Process Prior $BDP(G^p_0, \alpha^p_0, \kappa)$ 
We can estimate the empirical parameters for an isotropic Gaussian $G^p_0 \sim \mathcal{N}(\boldsymbol{\mu}^p,\boldsymbol{I}(\boldsymbol{\sigma}^p))$ and the prior's pseudo-count $\alpha_0^p$
%and the number of samples $\kappa=1$. This is done 
by using the latent vectors $\boldsymbol{Z}$ as follows: 
% Equations
$\boldsymbol{\mu}^p = \frac{\sum_i^N \boldsymbol{Z}_i}{N}$, $(\boldsymbol{\sigma}^p)^2 = \frac{\sum_i^N (\boldsymbol{Z}_i - \boldsymbol{\mu}^p)^2}{N-1}$ and 
% $\alpha^p_0 = \text{exp}(\frac{\sum_i^d (\boldsymbol{\mu}^p)^2}{2\sqrt{d/h}})$,
$\text{log}(\alpha^p_0) = \sum_i^N(\frac{\sum_j^d (\boldsymbol{\boldsymbol{Z}_{ij}})^2}{2\sqrt{d/h}})/N$,
where $N$ is the total number of tokens in the training corpus, $d$ is the dimension of the embedding and $h$ is the number of attention heads. This allows the prior mean to be the least informative representation in the center of the latent embeddings vector space. The variance is estimated from this mean. The empirical pseudo-count is kept in log space for numerical stability and is the expected scaled L2-norm of the latent vectors.

% used to weight the attention scores.  

% \paragraph{Informative Prior}
% Given the empirical prior formulation above, we can create an \textit{informed prior} through changing the data to the target distribution. In the presence of a domain shift we can use a sample of the new target distibution to adapt the prior.
% We show that initialisation of the NVIB layer with the empirical prior actually leads to minor improvements in performance post hoc. And controllable focus of the attention maps which smooths the distributions. 

\paragraph{Empirical prior analysis}

% Understand the distribution of different data in these models? - motivates grouping the hyperparametyers
% How much data is require by this prior? - motivates potential for low resource scenarios. als

% To motivate the empirical prior we analyse latent distribution of embeddings given data from different domains and a low resource scenarios. 
To further understand the distributions of our latent embeddings, we calculate the distribution of the empirical priors across all layers of the encoder and decoder attention mechanisms (Appendix \ref{sec:empirical_prior_analysis}). Given a model $\theta_x^*$ that has been trained on $x$ we consider generating priors from both in-domain and out-of-domain distributions.
% Grouping the parameters 
We find that the distribution of embeddings for the groups: encoder self-attention, decoder cross-attention and decoder causal self-attention, behave differently between and similarly within these attention groups. This observation motivates the choice of grouping the hyperparameters $\tau_\alpha^i$ and $\tau_\sigma^i$ where $i$ is an indicator for encoder, cross or decoder attention mechanisms. 
% Analysis of whats happening through the layer of the encoder and decoder (appendix?)
% The distribution of the encoder's embeddings are consistently centered around zero with low variance. The final layer of encoder latent embeddings used for cross-attention has a lower variance and lower expected log pseudo-count. Consider the decoder embeddings, we notice significantly larger means, variances and log pseudo-counts. 
% Data efficient 
We also consider the data requirements of the empirical prior by calculating the prior from a sub-sample of the training data. We found that the empirical prior can be created from as few as 0.1$\%$ of the training data ($\approx$ 200 examples) and achieve similar performance (Appendix \ref{sec:empirical_prior_data}). Hence, the empirical prior is data-efficient as it requires a minimal amount of training data for creation.  

\section{Post-Training NVIB Regularisation of Pretrained Models}
\label{sec:post-hoc-regularisation}
%\section{Applications to Pretrained Models}
\label{sec:eval}

% We can show a smooth regularisation with the empirical prior. Show the maps. Also show the experiment with a gaussian prior. Basically show usefulness of the empirical prior. 
% Claims smoother representations for transition between the prior. (heat map plots)

% Sparser representations by looking at the prior (attention maps!)

% What is this section about?
To evaluate our Bayesian reinterpretation of Transformers, we first show that our proposed identity initialisation results in empirical equivalence to pretrained Transformers.
%(Section \ref{sec:equivalence}).  
We then investigate how this reinterpretation allows us to apply an information theoretic post-training regularisation by changing the initialisation of the NVIB layers. 
% What is the advantage?
Once the empirical prior is estimated, 
adapting a pretrained Transformer to a new domain simply requires a hyperparameter selection on data from the new domain, which can be done solely with forward-passes of the model. 
% Why do we care?
This is an important contribution as the cost of backward-passes, regularisation and parameter updates becomes increasingly expensive as the models scale.  
% What problem does it solve?
The problem of domain generalisation is prominent in large language models and critical in the deployment of these models \citep{gulrajani2021in}. In the presence of a domain-shift, these models would normally need to be fine-tuned, which may become infeasible at scale.
% How we define our experiments?
Hence, we evaluate our nonparametric variational (NV) regularisation entirely post training, both in-domain and out-of-domain.

% We consider encoder-decoder Transformer such that we may influence all distributions of latent embeddings throughout the model. 

% The influence of the prior vector in the attention maps and the uncertainty  parameter in the attention calculation. 

% Road map what are we going to show.
For this evaluation, we consider the task of text summarisation.
%, and out-of-domain (OOD) generalisation. 
This task naturally requires the model to compress the information in the latent representation and would intuitively benefit from regularisation in the presence of a domain-shift.  
We first evaluate in-domain, showing that our identity initialisation results in empirical equivalence (Section \ref{sec:equivalence}) and that increasing the uncertainty results in a smooth space of models with different but still accurate predictions (Section \ref{sec:smooth_reg}). 
We then evaluate out-of-domain (OOD) generalisation, demonstrating that by changing the initialisation of our model we are able to achieve an information-theoretic post-training regularisation which improves OOD generalisation (Section \ref{sec:OOD_generalisation}).
This successful generalisation supports the argument that this Bayesian reinterpretation  
captures essential properties of
pretrained Transformer embeddings, as discussed more in Section~\ref{sec:discussion}.
% 
% We show that the empirical prior may be used as a novel form of domain adaption (\ref{sec:domain-adaption}). 

% \paragraph{Datasets} 
% To evaluate our summarisation models in-domain we use CNN/DailyMail (CNN/DM) \cite{cnndm_see-etal-2017-get} and Xsum \cite{narayan-etal-2018-dont}, and for out-of-domain on the Curation Corpus (CC) \citep{curationcorpusbase:2020} and SAMsum \citep{gliwa-etal-2019-samsum}.
% % and WikiHow \citep{wikihow2018}. 
% We have further dataset statistics in Appendix \ref{sec:dataDetails}.

% footnote{\hyperlink{https://huggingface.co/facebook/bart-large-cnn}{https://huggingface.co/facebook/bart-large-cnn}} and Xsum\footnote{\hyperlink{https://huggingface.co/facebook/bart-large-xsum}{https://huggingface.co/facebook/bart-large-xsum}}

\paragraph{Models}
We consider two models that are already pretrained and fine-tuned, available on HuggingFace\footnote{\url{https://huggingface.co/facebook/bart-large-cnn}}
\footnote{\url{https://huggingface.co/facebook/bart-large-xsum}}. In our experiments, we do not fine-tune or update the original model weights with data, instead only adapting these models with post-training techniques.  For baselines, we consider 16-bit, 8-bit and 4-bit quantisation models, which regularise the linear projections post-training \footnote{\url{https://github.com/TimDettmers/bitsandbytes}}.\footnote{We did not consider sparsity techniques as they were not easily available without further training or fine-tuning.}
For our proposed model, we set hyperparameters in the initialisation of our NVIB layers, which regularise the attention functions post-training.
For a comprehensive application of our attention regulariser, we consider encoder-decoder Transformer models, which accounts for: encoder self-attention, cross-attention and decoder causal-attention. For our summarisation tasks we consider BART models  \citep{lewis-etal-2020-bart}. We provide further implementation details regarding the models in Appendix \ref{sec:modelDetails}.

\subsection{Equivalence with the Identity Initialisation} \label{sec:equivalence}

% \fabio{First claim: Including and initialising NVIB into pretrained models leads to equivalence}

% This is done by taking a pretrained and fine-tuned model readily available from Huggingface and including the NVIB layer and denoising attention. Its important to note we do not fine-tune or update the original model weights whatsoever and only initialise our NVIB layer. For completeness we consider NVIB initialisation in encoder-decoder Transformer models which use self, cross and causal attention. 

% We consider typical summarisation datasets CNN/DailyMail \cite{cnndm_see-etal-2017-get} and Xsum \cite{narayan-etal-2018-dont} and 
% For Translation we consider WMT14 \cite{bojar-EtAl:2014:W14-33} datasets from French-English and German-English and readily available fine-tuned Transformer models\footnote{\hyperlink{https://huggingface.co/Helsinki-NLP/opus-mt-fr-en}{https://huggingface.co/Helsinki-NLP/opus-mt-fr-en}} \footnote{\hyperlink{https://huggingface.co/Helsinki-NLP/opus-mt-de-en}{https://huggingface.co/Helsinki-NLP/opus-mt-de-en}}.

% \jamie{Somewhere we need to say what the values of $d$ and $h$ are for these BART models.  Maybe here, maybe in the first appendix.}

We show empirically that pretrained Transformer models can be interpreted as NV models by showing that our identity initialisation (Section \ref{sec:pretrained_identity_initialisation}) does not change their predictions. We set the NVIB intialisation parameters $\tau_\alpha^i {=}10$ and $\tau_\sigma^i \approx 0$ ($1e^{-38}$ for \texttt{Float32}) for all $i$, which, respectively, down-weights the influence of the prior and reduces the interpolation between the query and value in the attention mechanism (Equation~\eqref{eqn:evalDAttn}). 
%Figure \ref{fig:equivalent_transformer} provides a visual intuition of our initialisation.
Table~\ref{tab:equivalence} shows that the accuracy of the NV-Transformer initialisation is the same as that of the pretrained Transformer, and, more precisely, we find that comparing the predicted summaries gives us exactly the same outputs in both cases (leftmost points in Figure~\ref{fig:overlap_with_baseline} below).

%  The 10 here refers to how many lines we wrap with
%\begin{wrapfigure}[10]{r}{0.3\textwidth}
%  \centering
%  \vspace{10pt}  % Controls up and down of the figure
%    \includegraphics[width=0.2\textwidth]{figures/pretrainedNVIB-pretrained_only.pdf}
%  \caption{A pretrained Transformer encoder layer with the NVIB identity intialisation.}
%  \label{fig:equivalent_transformer}
%\end{wrapfigure}

% \captionsetup[table]{skip=0pt,singlelinecheck=on}
\begin{table}[h]
\centering
  \caption{NV-Transformer equivalence to Pretrained Transformers. Validation cross-entropy (CE), Rouge-1, Rouge-2 and Rouge-L scores for in domain text summarisation.
  % \vspace{-2ex}
  }
  \label{tab:equivalence}
  \begin{tabular}{l|llll}
    \toprule
     % \textbf{Data} & \multicolumn{4}{c}{\textbf{BART}} & \multicolumn{4}{c}{\textbf{NviBART}} \\
     Data & CE  & Rouge-1  & Rouge-2 & Rouge-L \\
    \midrule
    BART \scriptsize{(CNN/DM)} & 2.71 & 43.45 & 21.17 & 30.56 \\
    NV-BART \scriptsize{(CNN/DM)} & 2.71 & 43.45 & 21.17 & 30.56 \\
    \midrule
    BART \scriptsize{(Xsum)} & 2.30 & 44.03 & 21.90 & 36.47 \\
    NV-BART \scriptsize{(Xsum)} & 2.30 & 44.03 & 21.90 & 36.47 \\
    \bottomrule
  \end{tabular}
\end{table}

\paragraph{Attention maps}
Examining the attention maps of the encoder and decoder, we see that the prior component is included in the keys of attention, but the identity initialisation down-weights this prior to the point that there is no attention weight to this component (Appendix \ref{sec:attention_plots}).
% We also include attention maps given our identity initialisation in Appendix \ref{sec:attention_plots}. We can see the prior component is included in the keys of attention, but the identity initialisation down weights this prior such that there is no attention weight to this component.
%we get an equivalence with the pretrained model.  
%
In the next section we show that by altering this initialisation we are able to get a smooth transition from this attention distribution of the pretrained model to one with attention to our prior distribution component.

% \fabio{Practical computation impact: As a reviewer I would be interested in latency of inclusion here. How many more parameters does it add \% per layer and in the model overall \%? How much time does it add to computation? - measure time when equivalent for all datasets eg: its 10\% slower given our current implementations.}

% Moreover, we find a space between the pretrained model and the prior that achieves better performance due to the interpolation with the query and sparser representations by putting that extra weight on the prior.

%  \fabio{When running the full NVIB translation model, the alpha projections in the encoder layer, project some inputs to inf. This breaks our equivalence. For now I will leave translation and focus on summarisation.}
% \fabio{Key caching is now working! wow that was a catacomb of coding.}

% Hypothesis 2: Take pretrained model (or not) and fine-tune until its potential on Task. Compare that with NVIB with pretrained model (or not) and fine tune until its potential on task.

% NVIB should be better, eg: better translation, low resource or summarisation.

% low resource and high resource is about data, so we can run low resource experiments by restricting data.

% Out of distribution analysis: We want to show that NVIB generalises better due to the distributional assumptions. So we need a similar but different distribution task. Train on Arabic and then use to transfer to dialects. Dutch to Afrikaans,  German swiss german etc 

\subsection{In-Domain Regularisation} \label{sec:smooth_reg}

% Claim 1
We show that our reinterpretation allows for a controllable and smooth regularisation of the model's performance. 
Given that the original model has been fine-tuned on the in-domain task, we would normally expect that any deviation from the predictions of the original model would result in a degradation of accuracy, with progressively greater deviations resulting in progressively worse accuracy.  
% a more specific version of Claim 1?
Instead, our regularisation results in a range of models which make different predictions but have similar accuracies to the original model.
% Claim 2 
This suggests that the added uncertainty can be interpreted as the real uncertainty about in the data distribution which has been captured by the pretrained model.
%such that it does not collapse but finds a space of different but similarly performing models.

We specify our in-domain experiment as: a model $\theta_x^*$ that has been trained on data from domain $x$ and evaluated on the same domain $x$. For our experiments we consider BART summarisation models that have already been fine-tuned on either CNN/DailyMail data \citep{cnndm_see-etal-2017-get}, or Xsum data \citep{narayan-etal-2018-dont}, and we use empirical priors created from their respective training corpora. We then evaluate these models on the validation sets of their same domain. 

% \fabio{Show in domain regularisation. Regularise from pure equivalence to some collapse. I interpolate bewteen the best model I found. We compare rouge overlap with the baseline model and see that when we arent perfect we get good models. This shows that we can get smooth controllable regularisation even in domain but also that there exist a lottery of good  models sitting in this space. Then the next question is well given weird inputs (OOD) then perhaps there is a model that is good at solving it close by.}
\begin{figure}[h]
    \centering
\includegraphics[width=0.35\textwidth]{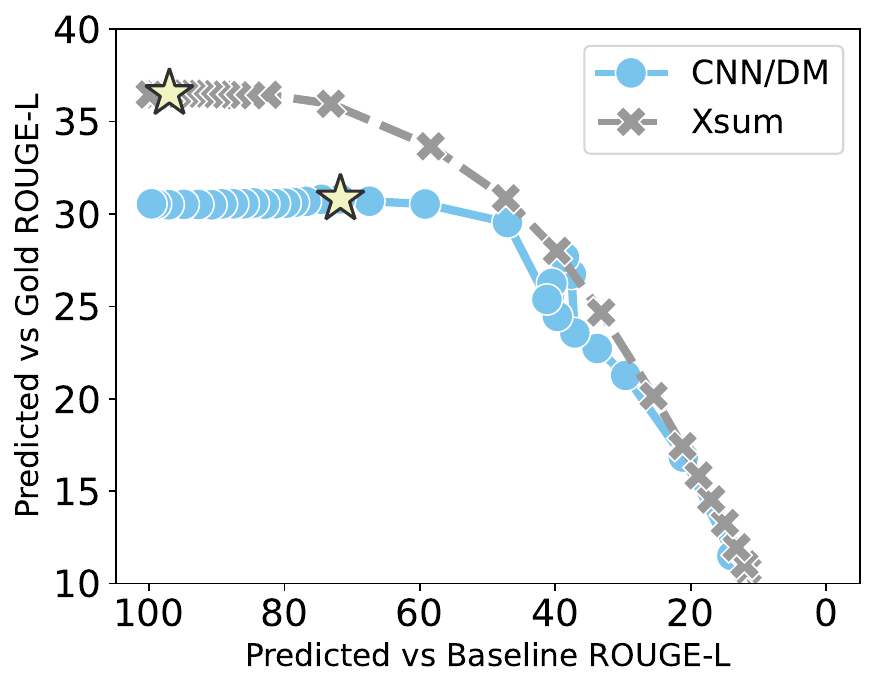}
    \caption{Each point represents a different NVIB intitialisation validated in-domain. The NV-Transformer's output is compared against the original non-regularised model ($x$-axis) and against the gold summary ($y$-axis). The stars indicate the highest performance.}
    \label{fig:overlap_with_baseline}
\end{figure}
Initially, for each NVIB hyperparameter individually, we find the point where the predictions are unchanged and the point where over-regularisation degrades performance. Thereafter we use a random uniform search of 50 samples in this range for each hyperparameter, and select the best model based on the in-domain validation set, which is shown as the same-domain model in Table~\ref{tab:test_OOD} below (more results are given in Appendix \ref{sec:validation_results}).  
To characterise the range of these regularised NV models, Figure \ref{fig:overlap_with_baseline} plots them in a line ordered by their
% projection onto the line of 
linear interpolations between the equivalent-initialisation and the over-regularised corners of this sample space.  The NV-Transformers are plotted by comparing their output using Rouge-L against the gold summary ($y$-axis) and against the original non-NV model's output ($x$-axis). The stars show models with slightly improved accuracy in-domain. As we increase the regularisation from the identity intialisation (left) we discover a space of models which are not only different from the original model ($x$-axis of $< 100$\% overlap), but also equally good (same $y$-axis performance). The inclusion of NVIB allows us to smoothly transition between models which make different predictions but have equivalent or similar accuracy.

\subsection{Out-of-Domain Generalisation} \label{sec:OOD_generalisation}

% Our claim
We hypothesize that our post-training information-theoretic regulariser improves out-of-domain summarisation performance. 
% This is an increasing important problem as pretrained models get more costly to fine-tune. 
To evaluate our regulariser in the presence of a domain-shift we define the following experiment: a model $\theta_x^*$ that has been trained on data $x$ and evaluated on the same task but data from a different domain $y$. We consider a BART \citep{lewis-etal-2020-bart} summarisation model $\theta_x^*$ already fine-tuned on data $x$ and we use this training data of $x$ to create our empirical prior, where as above this in-domain data is either CNN/DailyMail or Xsum\footnote{We also considered using an empirical out-of-domain prior but found similar or worse performance than the in-domain prior.}. We then evaluate these models on the validation sets of out-of-domain summarisation corpora from either Xsum or CNN/DailyMail, respectively, and from Curation Corpus (CC) \citep{curationcorpusbase:2020}, SAMsum \citep{gliwa-etal-2019-samsum} and WikiHow \citep{wikihow2018}. We provide further dataset details in Appendix \ref{sec:dataDetails}.

To evaluate on the test set we select the best model on the validation set (Appendix \ref{sec:validation_results}).  Table \ref{tab:test_OOD} reports the test set Rouge-L on the OOD text summarisation datasets for the post-training regularisation methods. For baselines, we consider the original 32-bit model without any post-training regularisation, and its quantisation models of 16-bit, 8-bit and 4-bit \citep{dettmers8bit2022}. 
\begin{table}[h]
\centering
  \caption{Post-training regularisation on OOD summarisation. We report test set Rouge-L.}
  \label{tab:test_OOD}
  \begin{tabular}{lcc|ccc}
    \toprule
     &
    \multicolumn{2}{c}{} & \multicolumn{3}{c}{Out-of-Domain} \\
    \textbf{Model} &
    \textbf{CNN/DM} & \textbf{Xsum} & \textbf{CC} & \textbf{SAMsum}
    & \textbf{WikiHow} \\
    \midrule
    BART \scriptsize{(CNN/DM)}  & \textbf{29.99} & 13.12 & 24.99 & 22.42 & 9.26 \\
    BART-16bit \scriptsize{(CNN/DM)}  & 29.97 & 13.13 & 24.99 & 22.36 & 9.27 \\
    BART-8bit \scriptsize{(CNN/DM)}  & 29.55 & 13.13 & 24.67 & 22.13 & 9.36 \\
    BART-4bit \scriptsize{(CNN/DM)}  & 29.78 & 13.13 & 24.70 & 21.87 & 9.24 \\
    % NviBART-16bit \scriptsize{(CNN/DM)} & - & - & - & - & - \\
    NV-BART \scriptsize{(CNN/DM)} & 29.33
    & \textbf{13.99} & \textbf{25.04} & \textbf{22.60} & \textbf{9.41}  \\
    \midrule
    BART \scriptsize{(Xsum)} &  16.61 & 36.42 & 14.37 & 18.33 & 13.43 \\
    BART-16bit \scriptsize{(Xsum)} &  16.60 & \textbf{36.44} & 14.38 & 18.24 & 13.43 \\
    BART-8bit \scriptsize{(Xsum)} &  16.56 & 36.25 & 14.33 & 18.05 & 13.53 \\
    BART-4bit \scriptsize{(Xsum)} &  16.71 & 35.10 & 14.49 & 16.33 & 13.15 \\
    % NviBART-16bit \scriptsize{(Xsum)} & - & - & - & - & - \\
    NV-BART \scriptsize{(Xsum)} & \textbf{19.42} & 36.25 & \textbf{17.61} & \textbf{21.94} & \textbf{15.25} \\
    \bottomrule
  \end{tabular}
\end{table}

% What do we see?
Considering the original model trained on CNN/DailyMail, we notice that with NVIB regularisation the OOD performance on the test set improves over the baselines consistently, although minorly. In contrast, the original model trained on Xsum is substantially improved when NVIB regularisation is applied to OOD text summarisation. We speculate the larger improvement in the model trained on Xsum is due to the abstractive nature of the training data, whereas the model trained on CNN/DailyMail is more extractive. This shows that our information-theoretic, post-training regularisation can improve OOD generalisation, both over the original model and over quantisation.\footnote{We also considered a combining  NVIB regularisation with quantisation and found improvements over the baselines models (Appendix Table \ref{tab:validation_OOD}). This shows potential for improving compressed models in the presence of a domain shift.} 

\begin{figure}[h]
\centering
\begin{minipage}[t]{.24\textwidth}
    \includegraphics[width=\textwidth]{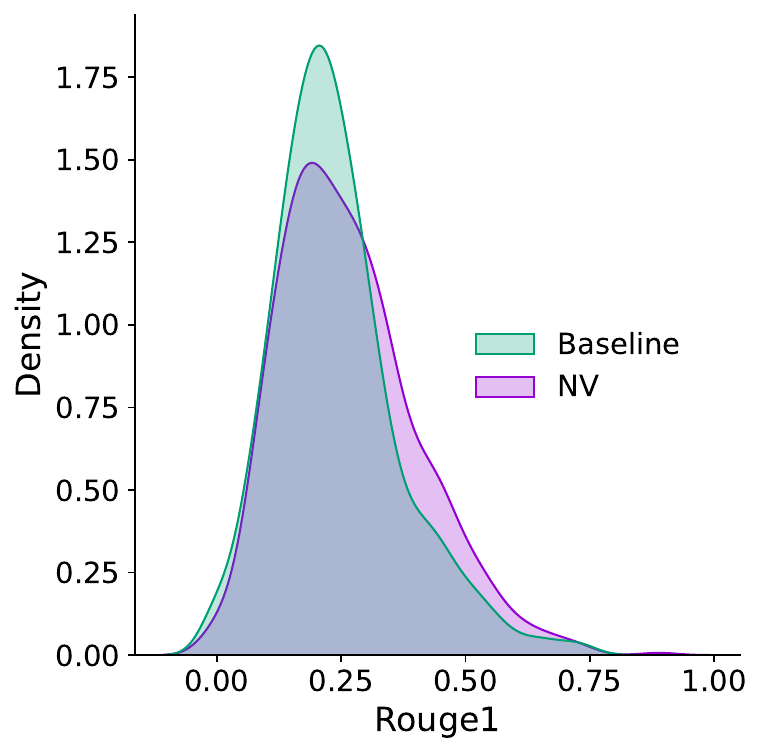}
\end{minipage}
\hfill
\begin{minipage}[t]{.24\textwidth}
    \includegraphics[width=\textwidth]{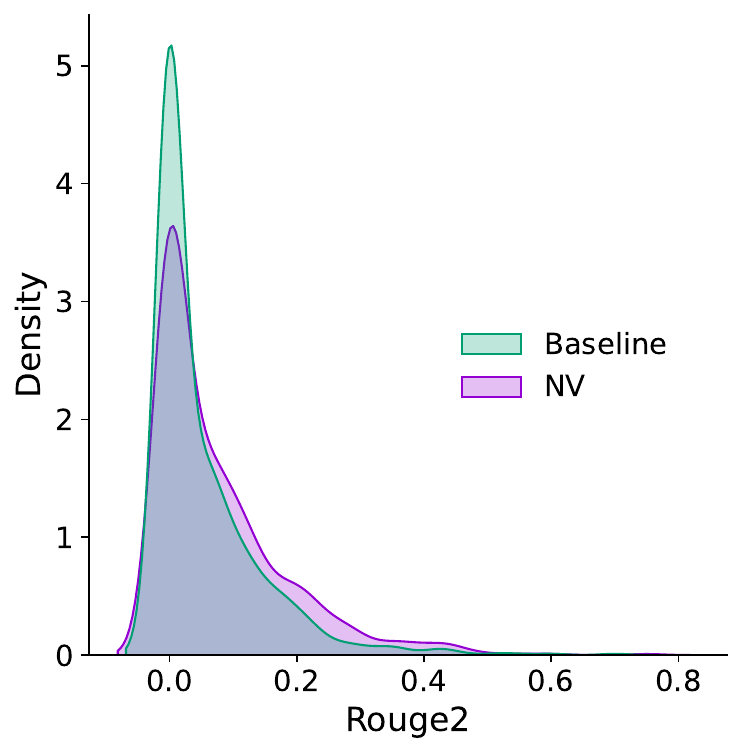}
\end{minipage}
\hfill
\begin{minipage}[t]{.24\textwidth}
\includegraphics[width=\textwidth]{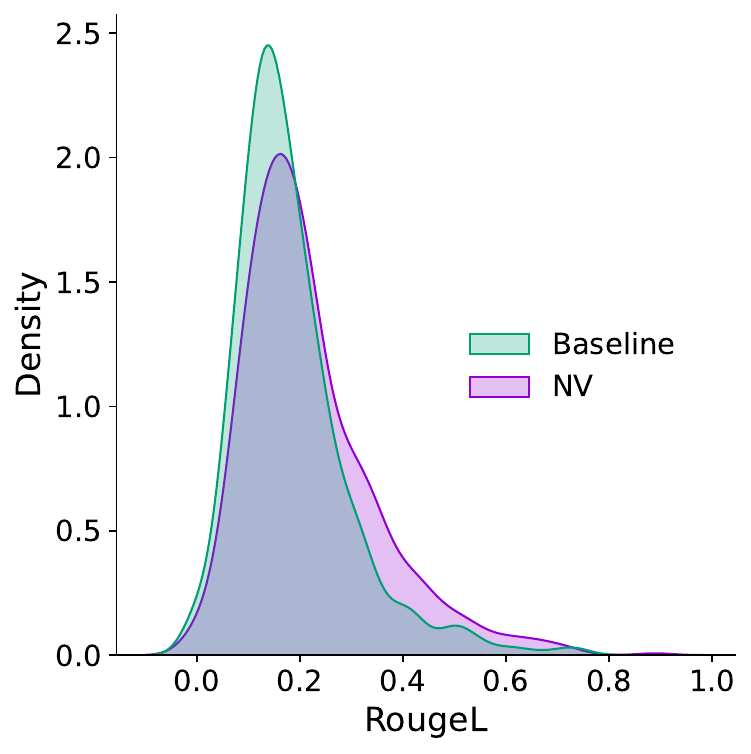}
\end{minipage}
\begin{minipage}[t]{.24\textwidth}
\includegraphics[width=\textwidth]{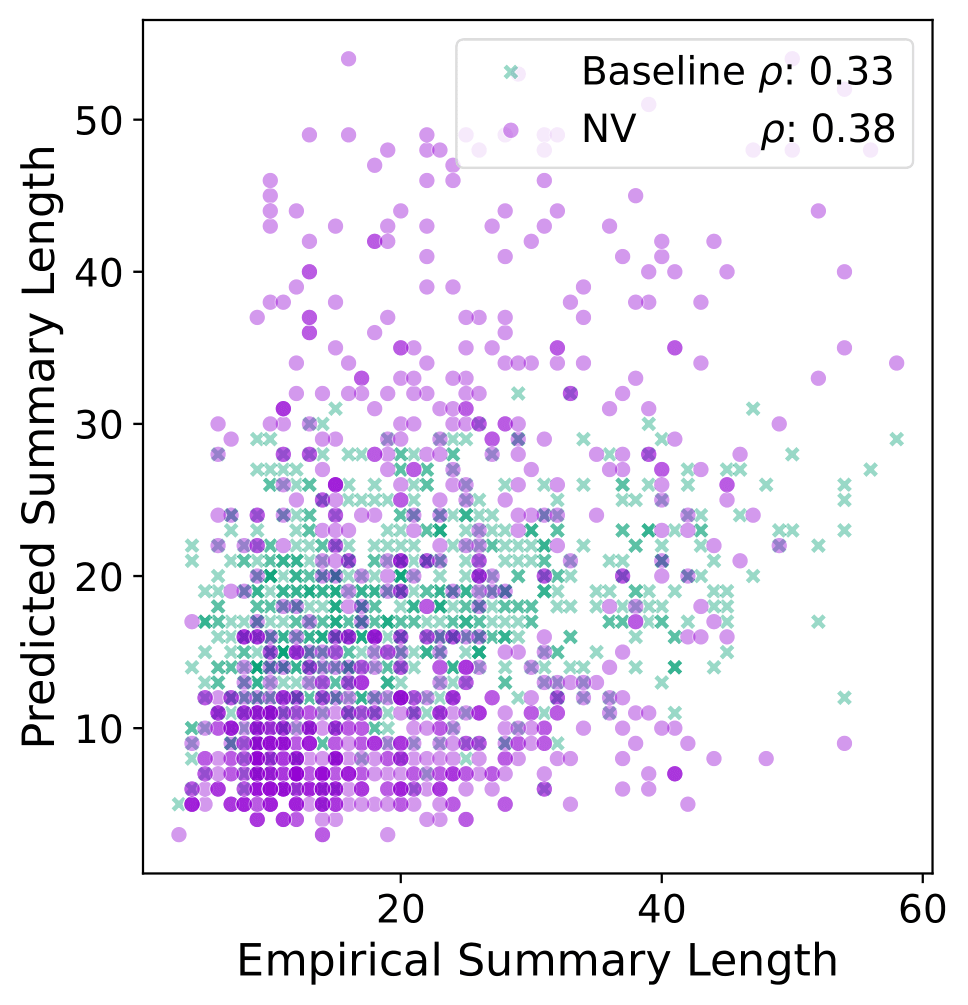}
\end{minipage}
\caption{A BART model trained on Xsum is compared against the reinterpreted model with NVIB regularisation on the SAMsum test dataset. \textbf{Left-Right}: Rouge-1, Rouge-2, and Rouge-L score distributions across test summaries; Empirical vs predicted summary length with Spearman's correlation.}
\label{fig:rouge_length_samsum_test}
\end{figure}

% Further analysis - What is happening to get these improvements.
To provide further analysis of the reasons for this OOD improvement, in Figure~\ref{fig:rouge_length_samsum_test} we compare the output summaries of a baseline BART model trained on Xsum data and a reinterpreted NV-Transformer model, for the OOD SAMsum test dataset. We provide similar validation set plots for all out-of-domain datasets in Appendix Figure \ref{fig:rouge_length_all}.

We conjecture that the length of the summary is a strong proxy for the information within the document. Since the baseline model has been trained to produce a summary of a certain length, the summaries on OOD data can be improved by adapting the length of the summary to the information content. Figure \ref{fig:rouge_length_samsum_test} (left) shows consistent improvements across the distributions of Rouge-1, Rouge-2 and Rouge-L overlap metrics against the gold summaries.  The modes of the distributions are smaller and the high-scoring tails are larger.
Figure \ref{fig:rouge_length_samsum_test} (right) plots the predicted summary lengths against the true empirical summary lengths and the spearman's correlation coefficient. We notice that NV-Transformer has not only decreased the average length of the outputs, but has a higher correlation with true empirical summaries. This suggests that information variation captured in the summary length is better captured by the reinterpreted NV model.  
We provide further  examples of generated summaries of all our summarisation models and in- and out-of-domain datasets in Appendix \ref{sec:generated_outputs}. 
These results support our hypothesis that our post-training, information-theoretic regulariser improves out-of-domain generalisation of pretrained Transformers.

\section{Discussion} \label{sec:discussion}

As well as being potentially useful, this surprising improvement from post-training regularisation suggests that our variational Bayesian reinterpretation of pretrained Transformers is an accurate model of how information is being captured in Transformer embeddings.  When we start adding uncertainty, thereby removing information, it does not degrade performance, even in-domain, presumably because the removed information is not relevant to the accuracy of the model. This interpretation is supported by Figure \ref{fig:overlap_with_baseline} and generated outputs in Appendix \ref{sec:generated_outputs}, which show that increasing uncertainty leads to a space of equally good but different models. Moreover, the attention maps support this interpretation because they remove weight on areas that are not relevant for the summary, such as punctuation or the start of sequence token (Appendix Figures \ref{fig:attention_encoders} and \ref{fig:attention_cross}).
Then at some point more uncertainty starts to also remove useful information, shown in the slow degradation on the right of Figure~\ref{fig:overlap_with_baseline}.

In the out-of-domain case, increasing uncertainty actually improves generalisation (Table~\ref{tab:test_OOD}), presumably because the removed information was the result of over-fitting to the training data on the source domain. 
%In this case the hyperparameters have been chosen based on the target domain, but these are just six hyperparameters which specify the amount of uncertainty to add; nothing else has been learned about the target domain at domain adaptation.
% Since this domain adaptation is done post-training, the pretrained model must already know what information is likely to generalise and what information is not.
Since this domain adaptation is done post-training, this suggests the pretrained model has a notion of the information useful for the task in general,
and that NVIB regularisation provides a method for identifying and removing the non-general information. In this sense, including NVIB helps us understand how pretrained Transformers are representing information in their embeddings.
In particular, our results suggest that the information in a given embedding dimension value is relative to the distribution over that dimension during training, with the mean value carrying no information and the scale being relative to the variance. Also, we speculate that the patterns in attention distributions often reflect an attempt to avoid information, in which case the attention weight can be shifted to the prior distribution without affecting performance.
We anticipate that these insights into how Transformer embeddings represent information
%, provided by NVIB's variational Bayesian framework, 
will help in the understanding and development of future improvements to Transformer architectures.

% \fabio{To discuss: (1) The model is capturing the distribution over the latent space. In-domain there exists equivalent models but OOD this removes the overfit-information. (2) No other framework allows for an information-theoretic regularisation of attention functions. (3) Fully post-training which shows that the information is captured by the model and NV-regularisation gives access to it.}

%***"This successful generalisation supports the argument that this Bayesian reinterpretation captures essential properties of pretrained Transformer embeddings, as discussed more in Section~\ref{sec:discussion}."

\section{Conclusion} \label{sec:conclusion}

This work contributes both to the development of novel models which use Nonparametric Variational Information Bottleneck regularisation, and to the understanding of pretrained Transformers.
We contribute multiple extensions to NVIB layers, for multihead attention (Appendix \ref{sec:evalDattn}), encoder self-attention (Appendix \ref{sec:selfDattn}) and decoder causal self-attention (Appendix \ref{sec:causalDattn}).  This development of the Nonparametric Variational Transformer architecture allow us to propose a Bayesian reinterpretation of pretrained Transformers.  We empirically show that this reinterpretation's identity initialisation of NV-Transformers (Section \ref{sec:pretrained}) is equivalent to its pretrained Transformer (Section \ref{sec:equivalence}). We also define a novel, data-efficient empirical prior defined by the model's empirical  distribution over embeddings (Section \ref{sec:pretrained_identity_initialisation} and Appendix \ref{sec:empirical_prior_data}).  With this empirical prior, increasing the uncertainty in the initialisation results in a smooth, information-theoretic, post-training regularisation of the model (Section \ref{sec:smooth_reg}). Empirically, we show the usefulness of this regularisation by improving the performance of existing pretrained Transformers in out-of-domain text summarisation (Section \ref{sec:OOD_generalisation}). 

% Future work - So the reviewers don't ask for more experiments
In future work we plan to evaluate our post-training regularisation on different tasks, such as machine translation and encoder-only and decoder-only applications.
% , and adapting our prior for low-resource scenarios. 
The success of NVIB regularisation in this post-training scenario suggests that even better generalisation could be achieved by fine-tuning on source-domain data with NVIB regularisation, for which our proposed mapping to a regularised NV-Transformer should be an effective initialisation.

\bibliography{references}

\appendix

\section{Denoising Multihead Attention}
\label{sec:multiDattn}

In this section we provide the details for \textit{denoising multihead attention} at training and evaluation time.  We define the set of Transformer latent embedding vectors as $\boldsymbol{Z} \in \mathbb{R}^{n \times d}$ and set of pre-projected queries as $\boldsymbol{U^\prime} \in \mathbb{R}^{m \times d}$. We assume the latent vectors are square 
% ($\mathbb{R}^{d \times d}$ and not $\mathbb{R}^{p \times d}$)
such that $\boldsymbol{W}^Q, \boldsymbol{W}^K,  \boldsymbol{W}^V \in \mathbb{R}^{d \times d}$ and biases $\boldsymbol{b}^Q, \boldsymbol{b}^K, \boldsymbol{b}^V \in \mathbb{R}^{d}$ are used to linearly project to the queries, keys and values, respectively. We define the standard attention weights before the softmax as follows: 

\begin{align}
\boldsymbol{A} = \tfrac{1}{\sqrt{d}} \underbrace{(\boldsymbol{U^\prime} \boldsymbol{W}^Q + \boldsymbol{b}^Q)}_{\boldsymbol{Q}}  \underbrace{(\boldsymbol{Z} \boldsymbol{W}^K + \boldsymbol{b}^K)^\top}_{\boldsymbol{K}^\top} \in \mathbb{R}^{m \times n}
\end{align}

Typically, for multihead attention the projected query $\boldsymbol{Q}$ and keys $\boldsymbol{K}$ are split into heads. In this definition, we split the linear projections by a divisible number of heads $h$ such that
$\boldsymbol{W}^Q, \boldsymbol{W}^K,  \boldsymbol{W}^V \in \mathbb{R}^{h \times d \times \frac{d}{h}}$ and biases $\boldsymbol{b}^Q, \boldsymbol{b}^K, \boldsymbol{b}^V \in \mathbb{R}^{h \times \frac{d}{h}}$, so that $\boldsymbol{Q} \in \mathbb{R}^{h \times m \times \frac{d}{h}}$ and
$\boldsymbol{K} \in \mathbb{R}^{h \times n \times \frac{d}{h}}$.  We can then specify multihead attention by defining a matrix of attention scores $\boldsymbol{A}\in \mathbb{R}^{h \times m \times n}$, for each head $i$:
\begin{align*}
\boldsymbol{A}_i = \tfrac{1}{\sqrt{d / h}}((\boldsymbol{Q}_i (\boldsymbol{W}^K_i)^\top \boldsymbol{Z}^\top + \boldsymbol{Q}_i (\boldsymbol{b}^K_i)^\top)
% mxd/h d/hxd dxn + mxd/h d/h
\end{align*}
where the bias term $\boldsymbol{Q}_i (\boldsymbol{b}^K_i)^\top \in \mathbb{R}^{m}$ is added across all $n$ keys, and thus is normalised out in the softmax below.
The scaling term also considers the heads and is division by $\sqrt{d / h}$. 
%This gives us $\boldsymbol{A}\in \mathbb{R}^{h \times m \times n}$.

For denoising attention, each head's query is projected into the space of the original set of vectors $\boldsymbol{Z}$, namely $\boldsymbol{U}_i {=} \boldsymbol{Q}_i (\boldsymbol{W}^K_i)^\top$, and so is still in $\mathbb{R}^{m \times d}$.  Thus each head can be viewed as doing denoising attention in the same way as single-head attention, with the only difference being that the variance of the theoretical query noise is now $\sqrt{d/h}\boldsymbol{I}$.  

\subsection{Training denoising attention}

In this work we do not do any training, but for completeness we include the equations for multihead attention at training time. 
The NVIB layer outputs the isotropic Gaussian parameters $\boldsymbol{\mu} \in \mathbb{R}^{(n+1) \times d}, \boldsymbol{\sigma} \in \mathbb{R}^{(n+1) \times d}$ and Dirichlet parameters $\boldsymbol{\alpha} \in \mathbb{R}^{(n+1)}$, which include the $(n{+}1)^\text{th}$ component for the prior. During training, these are used for sampling such that $\boldsymbol{\pi} \sim \text{Dir}(\boldsymbol{\alpha})$ and $\boldsymbol{Z} \sim \mathcal{N}(\boldsymbol{\mu}, \boldsymbol{\sigma})$.
Given these sampled weights and vectors,
the training-time denoising attention function is the same as the standard attention function with two changes: (1) the keys come from the sampled vectors $\boldsymbol{Z} \in \mathbb{R}^{(n+1) \times d}$, which include a vector sampled from the prior component; and (2) each key has an attention bias $\boldsymbol{b} \in \mathbb{R}^{(n+1)}$ which is determined by its weight $\boldsymbol{\pi} \in \mathbb{R}^{(n+1)}$.  Summing over heads $i$, the training-time denoising attention function is:
\begin{align*}
\text{DAttention}(.) 
  = \sum_i \text{Softmax}(\boldsymbol{A}_i + \underbrace{\log(\boldsymbol{\pi}) -\tfrac{1}{2\sqrt{d/h}}\|\boldsymbol{Z}\|^2}_{\boldsymbol{b}}) \underbrace{\addstackgap[7pt] (\boldsymbol{Z} \boldsymbol{W}^V_i + \boldsymbol{b}^V_i)}_{\boldsymbol{V}_i})
\end{align*}
The biases $\boldsymbol{b}$ are defined by adding the log of the sampled weights $\log(\boldsymbol{\pi}) \in \mathbb{R}^{(n+1)}$ from the NVIB layer and subtracting the scaled squared-L2-norms of the sampled vectors $\frac{1}{2\sqrt{d/h}}\|\boldsymbol{Z}\|^2 \in \mathbb{R}^{(n+1)}$. 
For multihead attention we only need to reuse the same biases $\boldsymbol{b}$ for each head, just like we reuse the same vectors $\boldsymbol{Z}$ for each head.

\subsection{Evaluation denoising attention} \label{sec:evalDattn}

During evaluation, as for training, the NVIB layer outputs the isotropic Gaussian parameters $\boldsymbol{\mu} \in \mathbb{R}^{(n+1) \times d}, \boldsymbol{\sigma} \in \mathbb{R}^{(n+1) \times d}$ and Dirichlet parameters $\boldsymbol{\alpha} \in \mathbb{R}^{(n+1)}$.
% \times 1
For evaluation, the denoising attention function is similar to the training-time denoising attention function with three changes: (1) the parameters are used directly without sampling;
% (1) the NVIB layer includes and the additional key vector for the prior thus $\boldsymbol{Z} \in \mathbb{R}^{(n+1) \times d}$, which for notational simplicity, we will omit; 
(2) there is a different attention bias $\boldsymbol{c} \in \mathbb{R}^{(n+1)}$ for each of the $(n{+}1)$ input vectors, including the prior; and (3) the attention values are computed with an interpolation between the query and value. We can write the denoising attention scores $\boldsymbol{A} \in \mathbb{R}^{h \times m \times (n+1)}$, for each head $i$, as follows:
% During inference we have an added term to the bias from training function c. this is just split by heads and included. Then the interpolation between the query and value. They define for a single query we need to show for multiple and multihead with biases.
\begin{align*}
\boldsymbol{A}_i = \boldsymbol{Q}_i (\boldsymbol{W}^K_i)^\top (\frac{{\boldsymbol{\mu}}}{ \sqrt{d / h} + {\boldsymbol{\sigma}}^2})^\top + \tfrac{1}{\sqrt{d / h}} \boldsymbol{Q}_i (\boldsymbol{b}^K_i)^\top
% \tfrac{1}{\sqrt{d / h}}
\end{align*}
where the bias term $\boldsymbol{Q}_i (\boldsymbol{b}^K_i)^\top \in \mathbb{R}^{m}$ is added across all $n$ keys, and thus is normalised out in the softmax below. 
%giving us $\boldsymbol{A} \in \mathbb{R}^{h \times m \times (n+1)}$.

To this attention score matrix $\boldsymbol{A}$, multihead evaluation denoising attention adds the same key biases $\boldsymbol{c} \in \mathbb{R}^{h \times (n+1)}$ across all $m$ queries and $h$ heads. For ease of notation we define $\boldsymbol{\sigma}^2_r=(\sqrt{d/h}+\boldsymbol{\sigma}^2)$ and note that $\alpha_0 = \sum^d_{j=1} {\alpha}_j$. 
% Not $\sqrt{{d}/{h}}$ as we are not in head space.
\begin{align*}
\boldsymbol{c} = \log(\frac{\boldsymbol{\alpha}}{\alpha_0}) -\tfrac{1}{2}\| \frac{\boldsymbol{\mu}}{\boldsymbol{\sigma}_r^2} \|^2 -\boldsymbol{1}_{d}(\log(\boldsymbol{\sigma}^r))^\top
\end{align*}
where $\boldsymbol{1}_{d}$ is a row vector of $d$ ones.

% @JH

Now we define the interpolation of each head's projected query $\boldsymbol{U}_i = \boldsymbol{Q}_i (\boldsymbol{W}^K_i)^\top \in \mathbb{R}^{m \times d}$ with the inference time value vectors 
${\boldsymbol{\mu}} \in \mathbb{R}^{(n+1) \times d}$.
The interpolation weights, $\frac{\boldsymbol{\sigma}^{2}}{(\sqrt{d/h}+\boldsymbol{\sigma}^2)}$ and $\frac{\sqrt{d/h}}{(\sqrt{d/h}+\boldsymbol{\sigma}^2)}$, are specific to values but not to queries, and the attention matrix, $\text{Softmax}(\boldsymbol{A}_i + \boldsymbol{c})$, maps values to queries, so we can accommodate the interpolation between queries and values by multiplying in the values before the dot product with the attention matrix and multiplying in the queries after this dot product.
%$\boldsymbol{Z}^\top + \boldsymbol{Q} (\boldsymbol{b}^K)^\top)$
\begin{align*}
\ignore{*
\text{Softmax}(&\boldsymbol{A}_i + \boldsymbol{c}) 
\boldsymbol{V}^\prime \boldsymbol{W}^V_i =
\\
=& \ \text{Softmax}(\boldsymbol{A} + \boldsymbol{c}) \left( \frac{\boldsymbol{\sigma}^{2}}{\boldsymbol{\sigma}_r^2}\odot \boldsymbol{1}_n \boldsymbol{U}_i +\frac{\sqrt{d/h}}{\boldsymbol{\sigma}_r^2}\odot\boldsymbol{\mu} \right) \boldsymbol{W}^V_i
\\
*}
\text{DAttention}(.) 
\hspace*{-10ex} &&
\\ =&& \sum_i \left( \left( \text{Softmax}(\boldsymbol{A}_i + \boldsymbol{c}) \frac{\boldsymbol{\sigma}^{2}}{\boldsymbol{\sigma}_r^2} \right) \odot\boldsymbol{U}_i + \text{Softmax}(\boldsymbol{A}_i + \boldsymbol{c})\left( \frac{\sqrt{d/h}}{\boldsymbol{\sigma}_r^2}\odot\boldsymbol{\mu} \right) \right) \boldsymbol{W}^V_i + \boldsymbol{b}^V_i
% \left(\text{Softmax}(\boldsymbol{A}_i + \boldsymbol{c}) \left( \frac{\boldsymbol{\sigma}^{2}}{\boldsymbol{\sigma}_r^2} \right) \odot\boldsymbol{U} + \text{Softmax}(\boldsymbol{A} + \boldsymbol{c})\left( \frac{\sqrt{d/h}}{\boldsymbol{\sigma}_r^2}\odot\boldsymbol{\mu} \right) \right) \boldsymbol{W}^V
\end{align*}

% \textbf{Should be something like:}
% \begin{align*}
% \text{DAttention}(.) =& 
% \sum_i
% \left(\text{Softmax}(\boldsymbol{A}_i + \boldsymbol{c}) \left( \frac{\boldsymbol{\sigma}^{2}}{\boldsymbol{\sigma}_r^2} \right) \odot\boldsymbol{U}_i + \text{Softmax}(\boldsymbol{A}_i + \boldsymbol{c})\left( \frac{\sqrt{d/h}}{\boldsymbol{\sigma}_r^2}\odot\boldsymbol{\mu} \right) \right) \boldsymbol{W}^V_i
% \end{align*}

\subsection{Denosing Self-Attention} \label{sec:selfDattn}

% \begin{wrapfigure}{r}{0.3\textwidth}
%   \begin{center}
%     \includegraphics[width=0.2\textwidth]{figures/NVIBSaTransformer.pdf}
%   \end{center}
%   \caption{Transformer encoder layer including NVIB and denoising self-attention.}
%   \label{fig:NVIBSA}
% \end{wrapfigure}
We build upon the contribution of \citet{behjati2023learning} of including NVIB and denoising attention for encoder self-attention. In that previous work the NVIB regulariser is applied to training single-head self-attention in the stacked layers of a Transformer encoder. 
% As seen in Figure \ref{fig:NVIBSA}, 
The queries for denoising self-attention are computed from the original $n$ vectors of the Transformer before they are projected to the $(\boldsymbol{\mu}, \boldsymbol{\sigma}, \boldsymbol{\alpha})$ parameters of NVIB, but the keys and values are computed from the vectors $\boldsymbol{Z} \in \mathbb{R}^{(n+1) \times d}$ of the NVIB layer. This allows every use of the attention function in self-attention to be done with denoising attention.
% Novelty is exponential activation and self-attention
From this previous work we maintain the exponential activation for the pseudo-counts. However, we remove the skip connection between pseudo-counts, because it is not part of the pretrained Transformer.
% We must give it in the supplimentary materials

\subsection{Denoising Causal Attention} \label{sec:causalDattn}

A causal mask is applied to self-attention in a Transformer decoder to simulate a next token prediction when using teacher-forcing \citep{vaswani2017}. The mask may be applied as before over the attention scores before the softmax. The prior component in the keys is never masked. This prior component acts like a new additional form of start-of-sequence token, without a positional embedding.  The bias terms do not depend on which other keys are masked, since the only term which depends on other keys is $\alpha_0$, and this term normalises out in the softmax.  Thus, the implementation is identical to denoising self-attention with the inclusion of a diagonal mask.

\section{Experimental Setup} \label{sec:ExperimentalSetup}

\subsection{Data} \label{sec:dataDetails}

In this work we use commonly available summarisation datasets from HuggingFace\footnote{\url{https://huggingface.co/datasets}}. We use the following datasets: \textbf{CNN/DailyMail} (\textbf{CNN/DM}) \citep{cnndm_see-etal-2017-get} which is one of the most widely used summarization corpora. It is based on news articles from the CNN and DailyMail websites. The summary sentences are a concatenation of human-generated "highlights" and bullet points. \textbf{Xsum} \citep{xsum_Narayan2018DontGM} is abbreviated for the "extreme summarization" dataset and is created from BBC news articles. The summaries are taken to be the first sentence of the article and the source document is the rest of the article. \textbf{Curation Corpus} (\textbf{CC}) \citep{curationcorpusbase:2020} is a dataset of professionally written summaries of news articles. This is the only freely available news summarization dataset with references that were written for the purpose of summarizing the article. For this dataset we use the version available on HuggingFace and split it manually into train, validation and test by a 50\%/25\%/25\% split. \textbf{SAMsum} \citep{gliwa-etal-2019-samsum} is an abstractive dialogue summarization dataset which is constructed to resemble the chats of a mobile messenger app. Each dialogue is written by a single linguist which can be formal or informal, and potentially contains slang, emoticons or typos. \textbf{WikiHow} was constructed from a knowledge base of how-to articles, explaining how to solve a task. We use the text as the document and the headline as the summary. Table \ref{tab:datasets} provides dataset statistics.
\begin{table}
\centering
  \caption{Dataset statistics 
  % \hspace*{2em}
  }
  \label{tab:datasets}
  \begin{tabular}{llll|ll}
    \toprule
    & \multicolumn{3}{c}{Examples} & \multicolumn{2}{c}{Mean words} \\
     \textbf{Dataset} & \textbf{Train} & \textbf{Val} & \textbf{Test} & \textbf{Document} & \textbf{Summary} \\
    \midrule
    CNN/DailyMail & 287K & 13.4K & 11.5K & 685 & 52 \\
    Xsum & 204K & 11.3K & 11.3K & 431 & 23  \\
    Curation Corpus & 15K & 7.5K & 7.5K & 504 & 83\\
    SAMsum & 14.7K & 0.8K & 0.8K & 94 & 20 \\
    Wikihow & 198K & 6K & 6K & 580 & 62 \\
    \bottomrule
  \end{tabular}
\end{table}
\subsection{Models}\label{sec:modelDetails}

% Intro to BART 
In this work we consider the BART encoder-decoder model \citep{lewis-etal-2020-bart}, which is a Transformer-based sequence-to-sequence model and is pretrained as a denoising autoencoder. This model has shown efficacy in a wide range of tasks including summarisation. In our work we consider BART (large)  summarisation models that are already pretrained and fine-tuned on CNN/DailyMail\footnote{\hyperlink{https://huggingface.co/facebook/bart-large-cnn}{https://huggingface.co/facebook/bart-large-cnn}} and Xsum\footnote{\hyperlink{https://huggingface.co/facebook/bart-large-xsum}{https://huggingface.co/facebook/bart-large-xsum}}, available on HuggingFace. We do not fine-tune or update the original model weights; we only initialise our NVIB layer as a form of post-training regularisation. 

The BART (large) model uses a 12 layer Transformer encoder and decoder with 16 attention-heads. The size for the word embedding vectors and model projections are 1024, and the size of the feed forward dimensions are 4096, which leads to models of approximately 406 million parameters. The inclusion of the NVIB projection layers, per attention mechanism,  results in an increase of approximately 11\% parameters. These projections are not trained and only initialised, which results in 459 million parameters in total.  It has an input context length of 1024 tokens. Where the tokens are created from a Byte-Pair-Encoding (BPE) tokenizer. 
% During inference we use a batch size of 4. 
Table \ref{tab:model_gen} provides specific autoregressive generation details for each of the models. 

% Currently inference is done on VGNG (Tesla V100-32GB) Original bart on XSUM takes 1hr. NVIB takes 3hrs 

% Each model's inference takes less than 3hrs to run on a single NVIDIA GeForce RTX 3090. However, this varies with the validation and test set size of the data, beam-search and max generation length.

\begin{table}[h]
\centering
  \caption{BART models specific autoregressive generation details.}
  \label{tab:model_gen}
  \begin{tabular}{l|cc}
    \toprule
     & BART \scriptsize{(CNN/DM)} & BART \scriptsize{(Xsum)} \\
    \midrule
    Number of beams & 4 & 6 \\
    Length penalty & 2 & - \\
    Max length & 142 & 62 \\
    Min length & 56 & 11 \\
    \bottomrule
  \end{tabular}
\end{table}

% Bart model details:
% . 
% What is d and what is h here? How many new tokens in generation etc?

\subsection{Validation Hyperparameters} \label{sec:validation_results}

To get final evaluation scores, we first decrease the search space of NVIB hyperparmeters by finding the points at which each hyperparameter individually has full equivalence and has degradation in performance. We record this space of parameters in Table \ref{tab:search_space}.

\begin{table}
\centering
  \caption{BART model's NVIB hyperparmeter selection space for random search.}
  \label{tab:search_space}
  \begin{tabular}{l|cccccc}
    \toprule
     & $\tau_\alpha^{e}$ & $\tau_\alpha^{c}$ & $\tau_\alpha^{d}$ & $\tau_\sigma^{e}$ & $\tau_\sigma^{c}$ & $\tau_\sigma^{d}$ \\
    \midrule
    min & -10 & -15 & 1 & $1e^{-38}$ & $1e^{-38}$ & $1e^{-38}$ \\
    max & 0 & 0 & 5 & 0.5 & 0.5 & 0.5 \\
    \bottomrule
  \end{tabular}
\end{table}

\begin{figure}[h]
    \centering
\includegraphics[width=\textwidth]{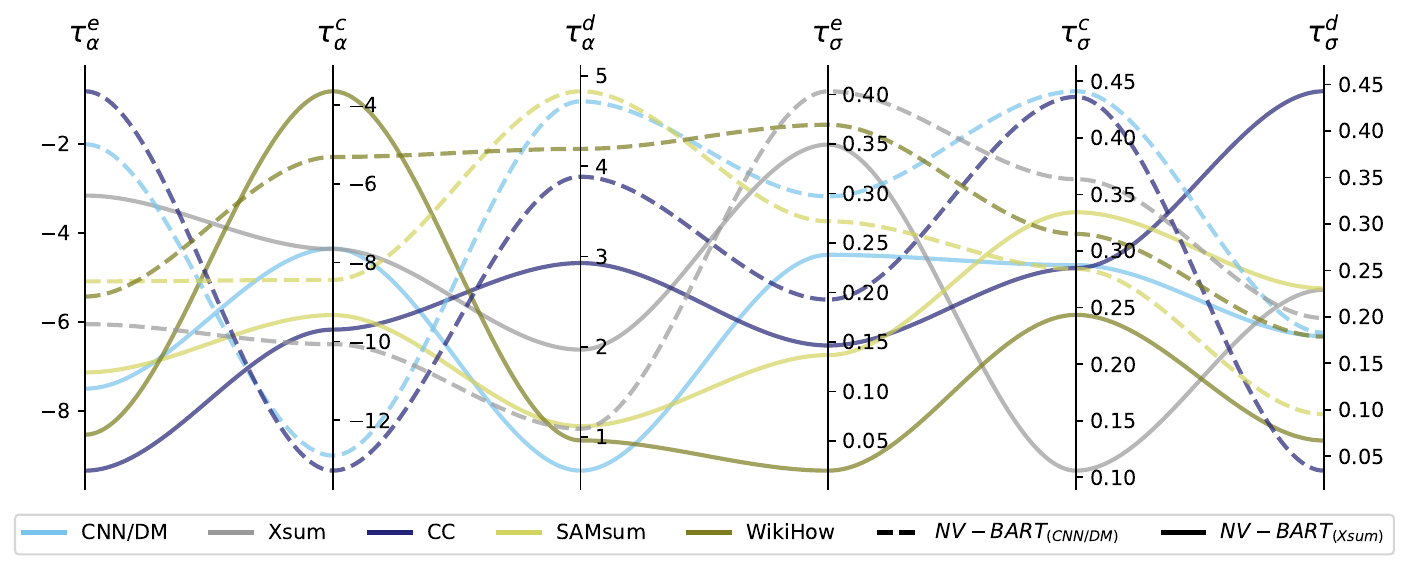}
    \caption{Parallel coordinate plots of best hyperparemeters across models and validation datasets. }
    \label{fig:parameter_plot}
\end{figure}

We notice that $\tau_\alpha^e$ and $\tau_\alpha^c$ can be decreased by several standard deviations before the noise affects the performance. We also notice that the $\tau_\alpha^d$ range shows that the decoder is more sensitive to this parameter. 
The interpolation parameters $\tau_\sigma$ have about the same sensitivity across the encoder, cross attention and decoder.
%similarly show less sensitivity in the encoder and cross attention and more sensitive to the decoder $\tau_\sigma^d$. 
The best hyperparamters for each model and each validation dataset are visualised in Figure \ref{fig:parameter_plot}.

\begin{table}[t]
  \caption{Post-training regularisation on OOD text summarisation. We report validation set Rouge-L.}
  \label{tab:validation_OOD}
  \centering
  \begin{tabular}{lcc|cccc}
    \toprule
     &
    \multicolumn{2}{c}{} & \multicolumn{3}{c}{Out-of-Domain} \\
    \textbf{Model} &
    \textbf{CNN/DM} & \textbf{Xsum} & \textbf{CC} & \textbf{SAMsum}
    & \textbf{WikiHow} \\
    \midrule
    BART \scriptsize{(CNN/DM)}  & 30.56 & 13.12  & 25.24  & 23.11 & 9.19 \\
    BART-16bit \scriptsize{(CNN/DM)} &  30.55 & 13.13 & 25.25 & 23.11 & 9.20 \\
    BART-8bit \scriptsize{(CNN/DM)} &  30.47 & 13.05 & 25.01 & 23.22 & 9.17 \\
    BART-4bit \scriptsize{(CNN/DM)} &  30.33 & 13.14 & 24.78 & 22.51 & 9.17 \\
    NV-BART-16bit \scriptsize{(CNN/DM)} & 29.70  & \textbf{14.05} & 25.38 & 23.38 & \textbf{9.40} \\
    NV-BART \scriptsize{(CNN/DM)} & \textbf{30.80} & 14.00 &  \textbf{25.46} & \textbf{23.57} & 9.37  \\
    \midrule
    BART \scriptsize{(Xsum)} & 16.57 & 36.47 & 14.41 & 18.68 & 13.35 \\
    BART-16bit \scriptsize{(Xsum)} &  16.57 & \textbf{36.48} & 14.42 & 18.67 & 13.35 \\
    BART-8bit \scriptsize{(Xsum)} &  16.53 & 35.78 & 14.42 & 17.84 & 13.14 \\
    BART-4bit \scriptsize{(Xsum)} &  16.39 & 35.05 & 14.46 & 16.45 & 13.05 \\
    NV-BART-16bit \scriptsize{(Xsum)} &  18.96 & 36.22 & 17.43 & \textbf{23.31} & \textbf{15.06}  \\
    NV-BART \scriptsize{(Xsum)} & \textbf{19.43} & 36.45 & \textbf{17.70} & 23.29 & 14.96  \\
    \bottomrule
  \end{tabular}
\end{table}

After finding the hyperparameter range, we perform a random search of 50 trials for each dataset to find the best regularised models. Table \ref{tab:validation_OOD} reports the validation results on the out-of-domain text summarisation task for BART models with post-training regularisation methods, including quantisation and NVIB regularisation. For quantisation we consider 32-bit (the uncompressed baseline), 16-bit, 8-bit and 4-bit. We also include a combination of NVIB regularisation with quantisation, but the implementation currently only supports 16-bit.

\begin{figure}
\centering
\begin{minipage}[t]{.24\textwidth}
    \includegraphics[width=\textwidth]{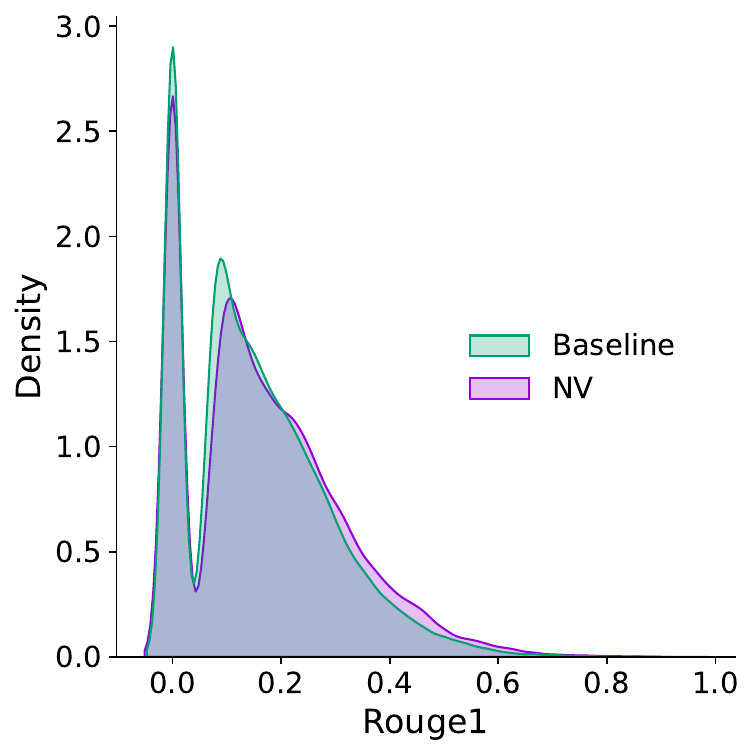}
\end{minipage}
\hfill
\begin{minipage}[t]{.24\textwidth}
    \includegraphics[width=\textwidth]{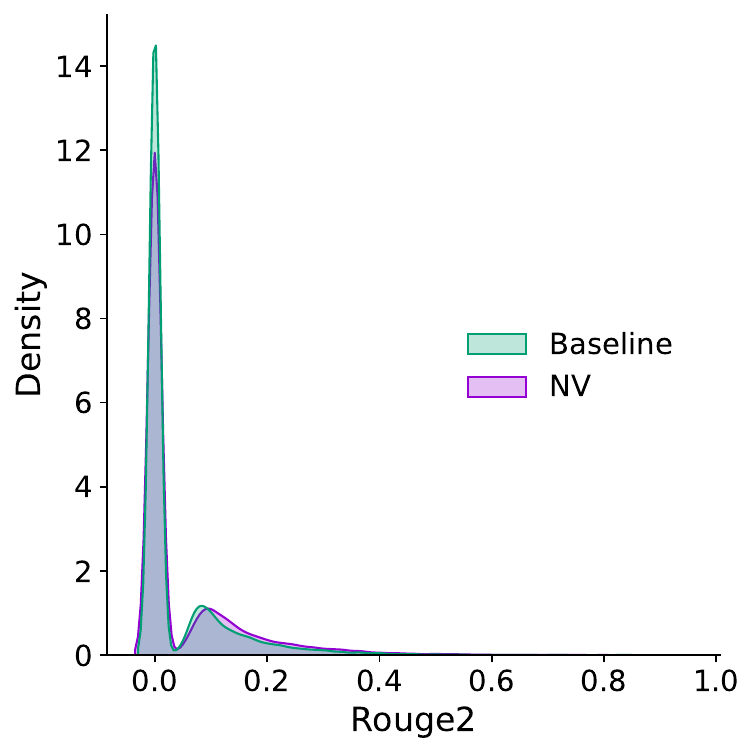}
\end{minipage}
\hfill
\begin{minipage}[t]{.24\textwidth}
\includegraphics[width=\textwidth]{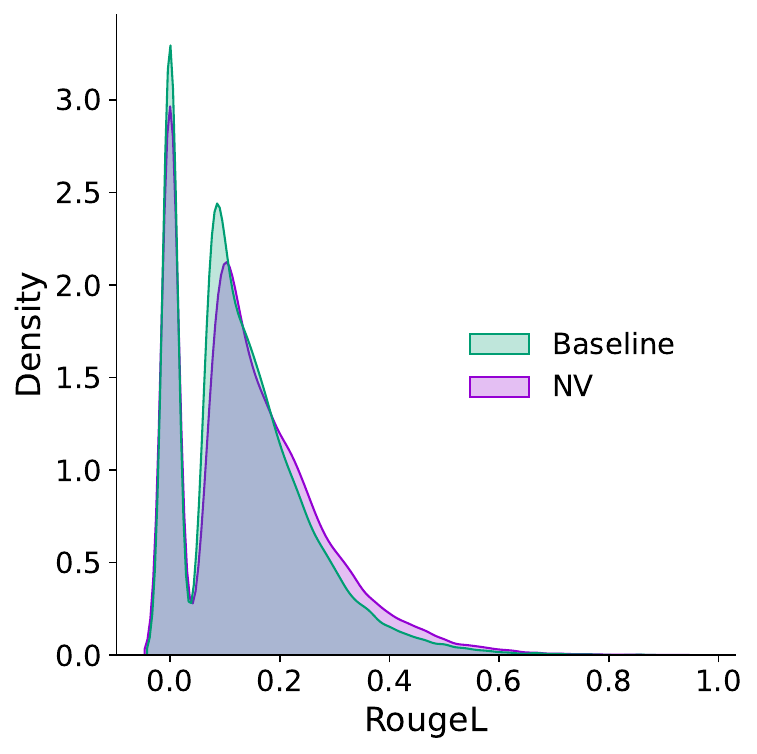}
\end{minipage}
\hfill
\begin{minipage}[t]{.24\textwidth}
\includegraphics[width=\textwidth]{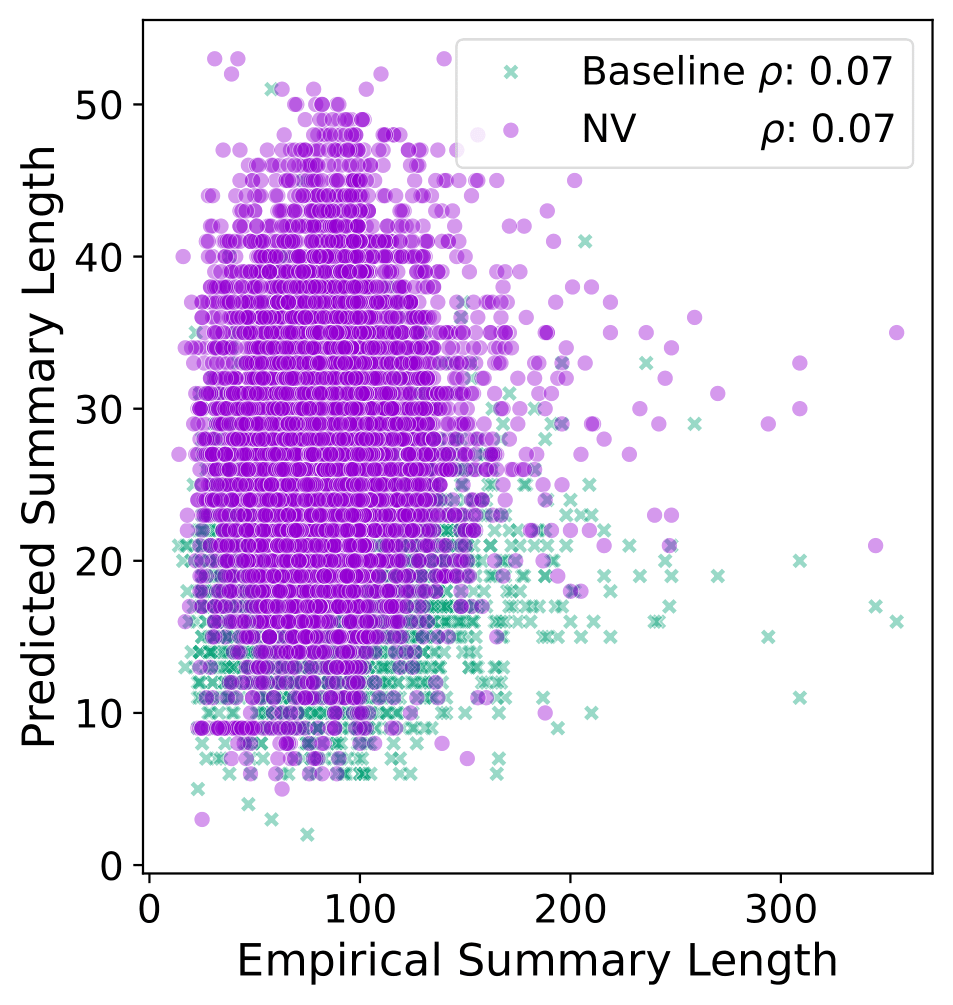}
\end{minipage}
\hfill
\begin{minipage}[t]{.24\textwidth}
    \includegraphics[width=\textwidth]{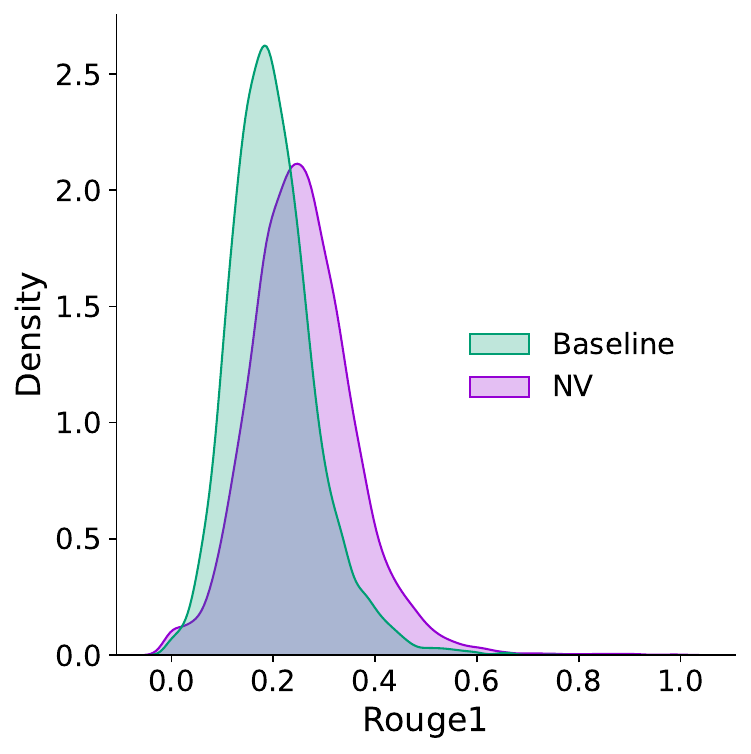}
\end{minipage}
\hfill
\begin{minipage}[t]{.24\textwidth}
    \includegraphics[width=\textwidth]{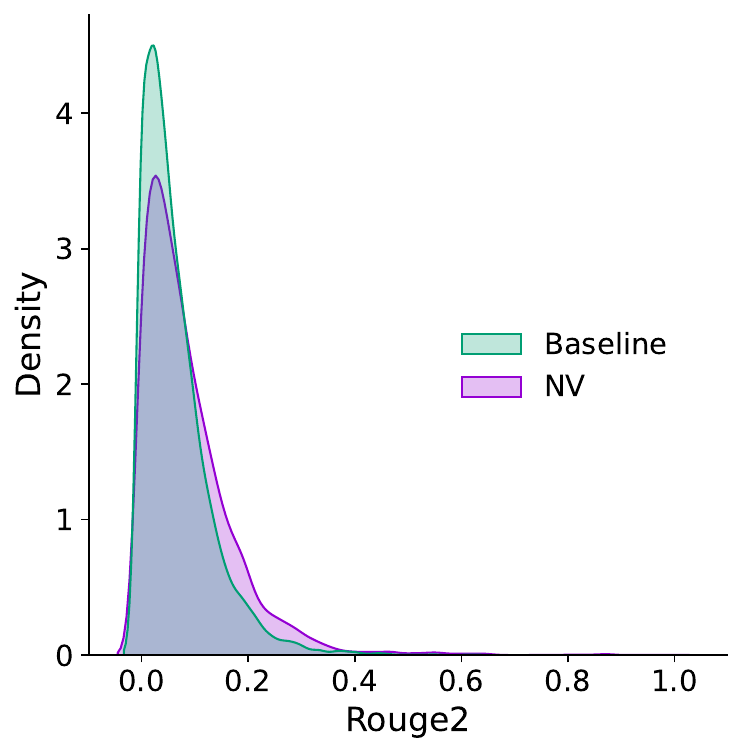}
\end{minipage}
\hfill
\begin{minipage}[t]{.24\textwidth}
\includegraphics[width=\textwidth]{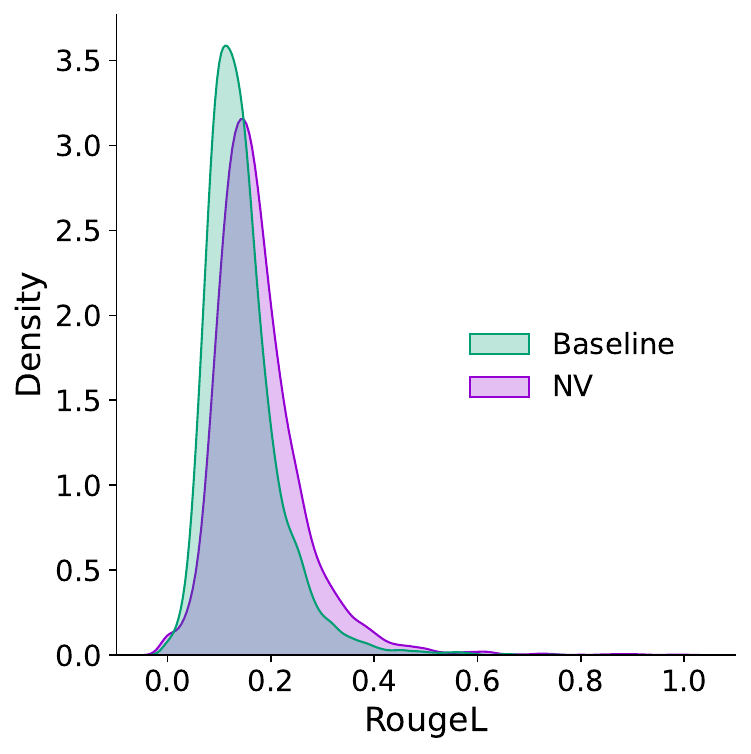}
\end{minipage}
\hfill
\begin{minipage}[t]{.24\textwidth}
\includegraphics[width=\textwidth]{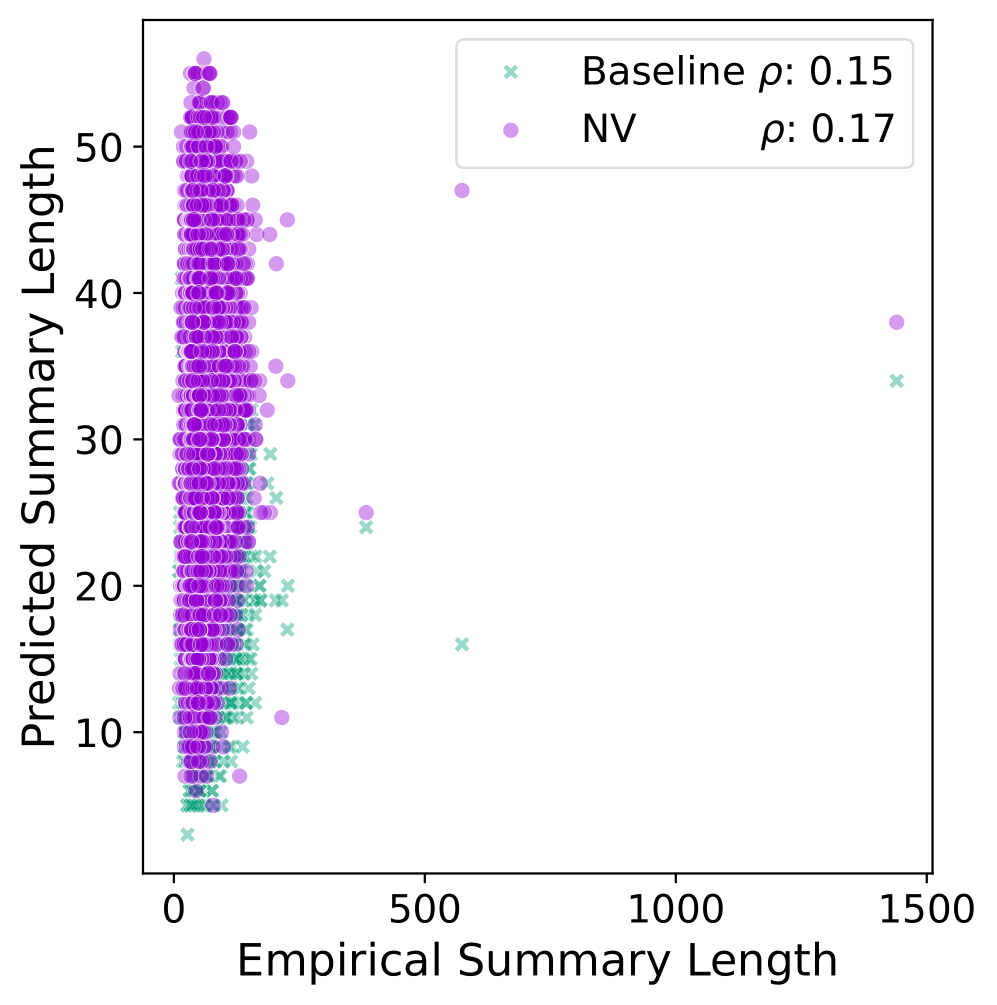}
\end{minipage}
\hfill
\begin{minipage}[t]{.24\textwidth}
    \includegraphics[width=\textwidth]{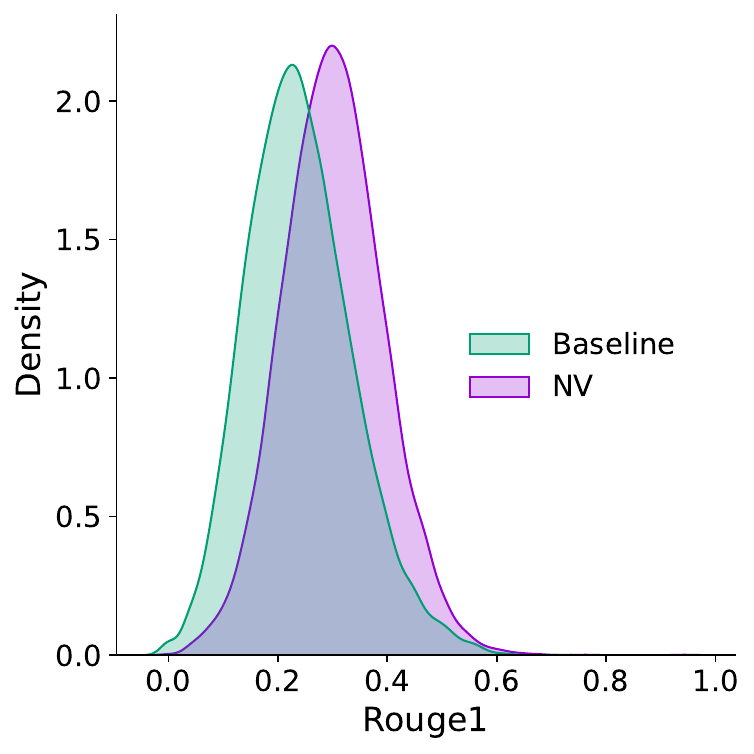}
\end{minipage}
\hfill
\begin{minipage}[t]{.24\textwidth}
    \includegraphics[width=\textwidth]{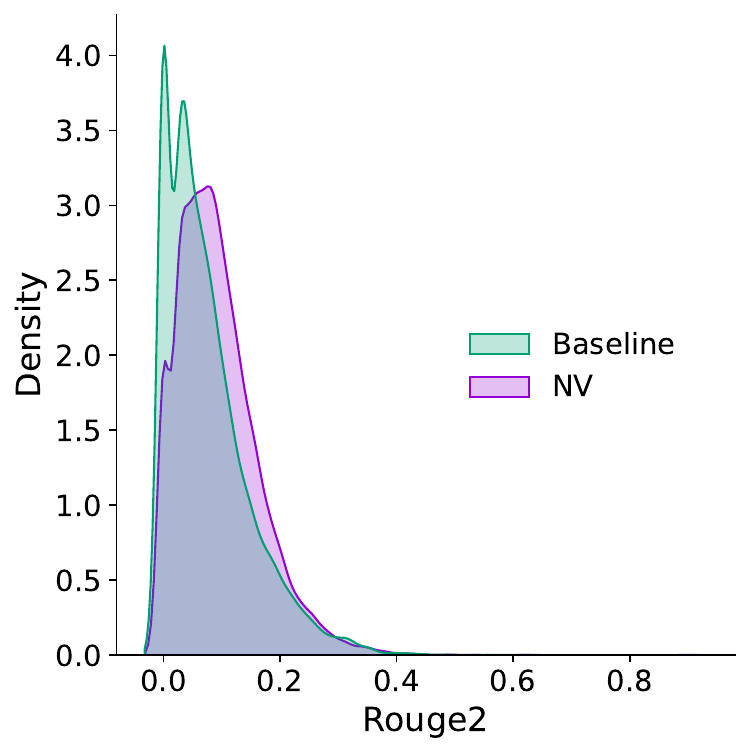}
\end{minipage}
\hfill
\begin{minipage}[t]{.24\textwidth}
\includegraphics[width=\textwidth]{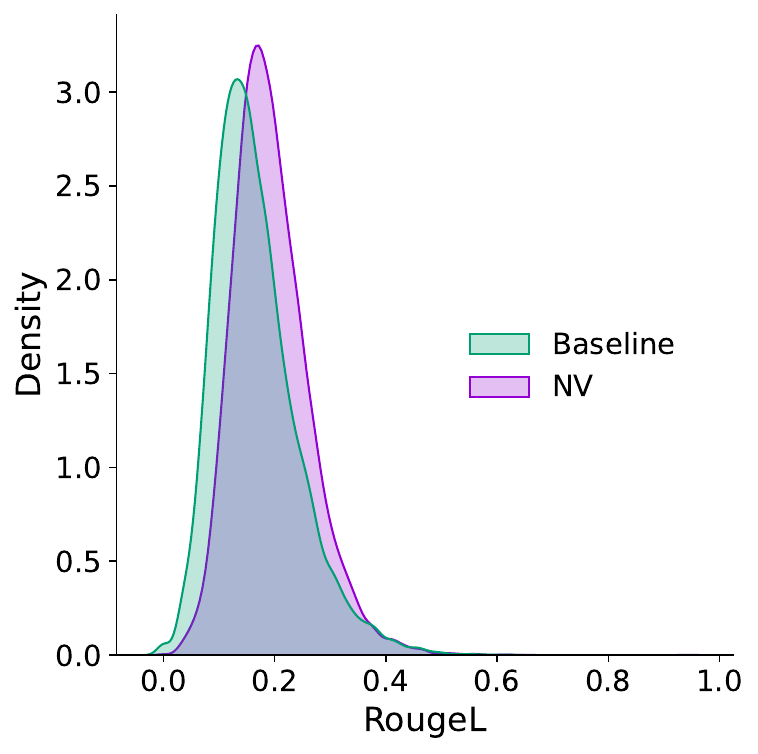}
\end{minipage}
\hfill
\begin{minipage}[t]{.24\textwidth}
\includegraphics[width=\textwidth]{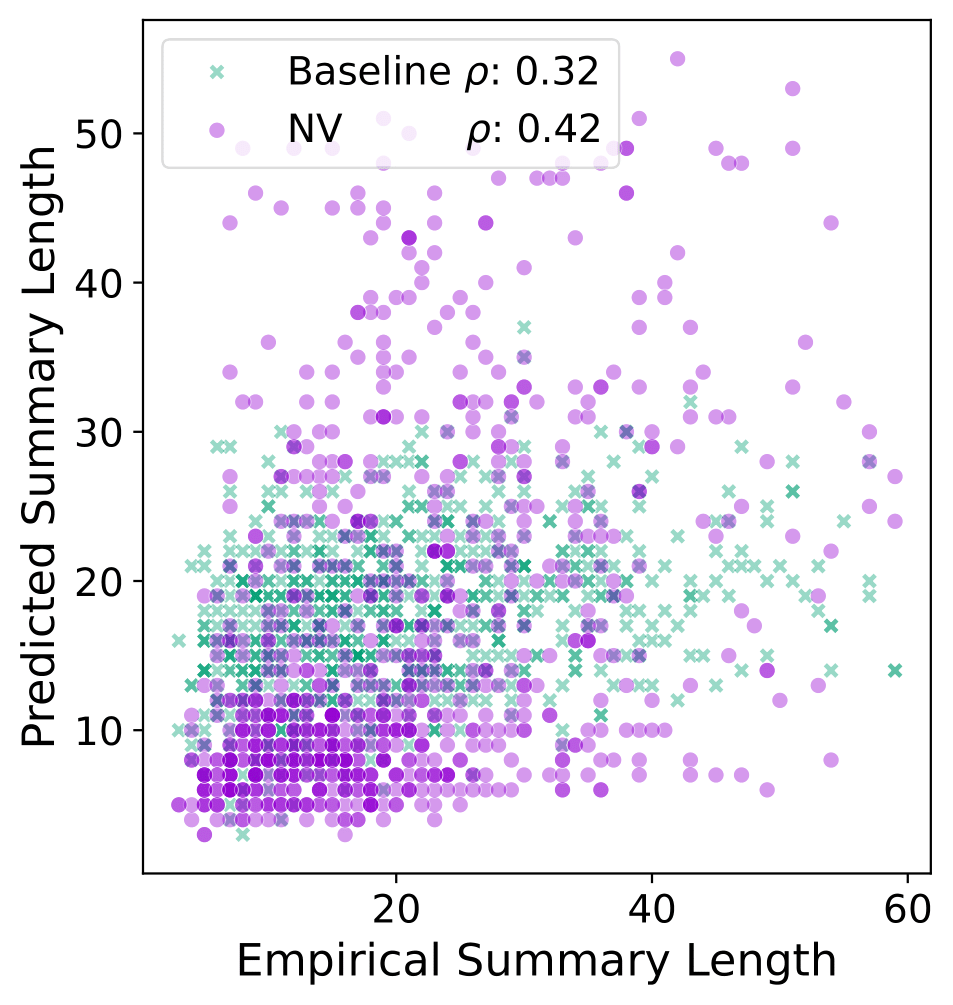}
\end{minipage}
\hfill
\begin{minipage}[t]{.24\textwidth}
    \includegraphics[width=\textwidth]{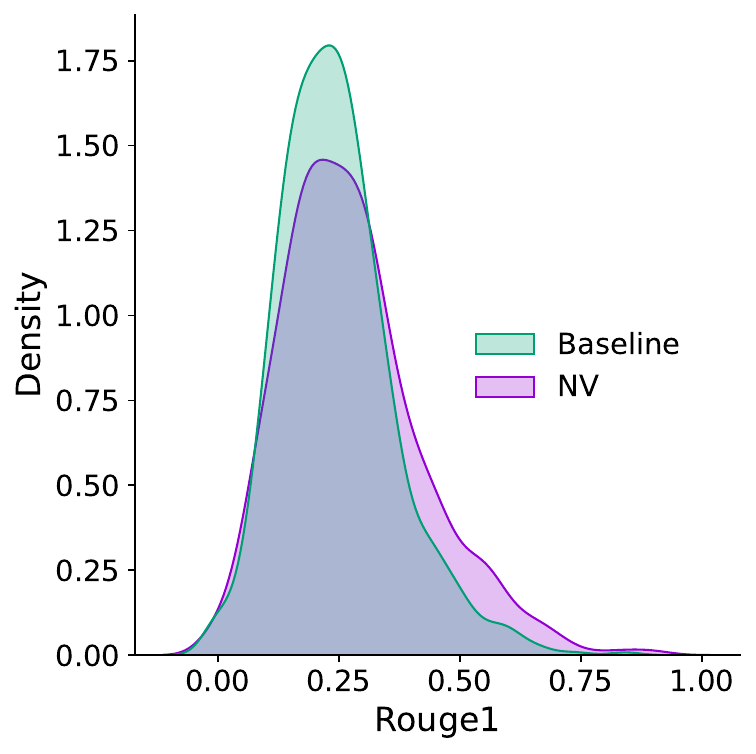}
\end{minipage}
\hfill
\begin{minipage}[t]{.24\textwidth}
    \includegraphics[width=\textwidth]{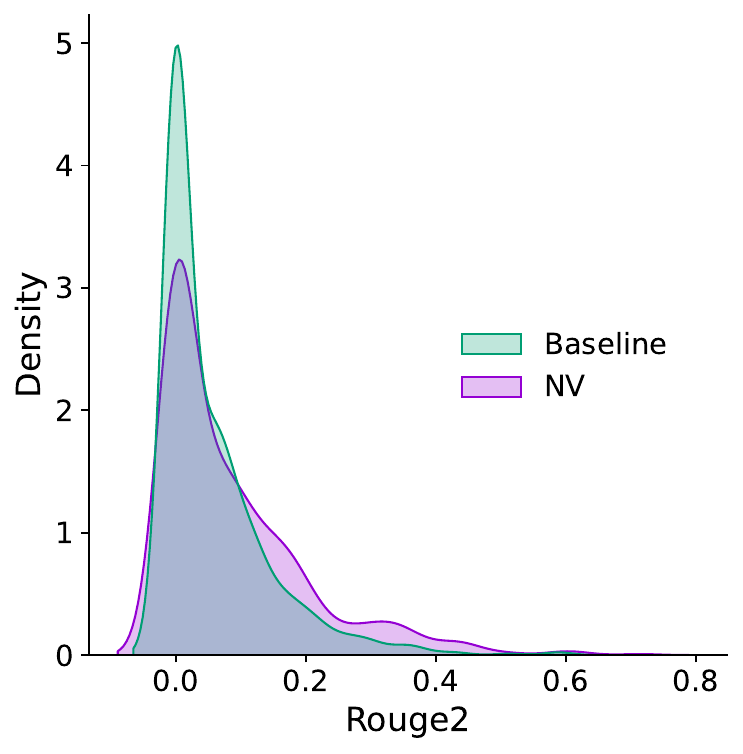}
\end{minipage}
\hfill
\begin{minipage}[t]{.245\textwidth}
\includegraphics[width=\textwidth]{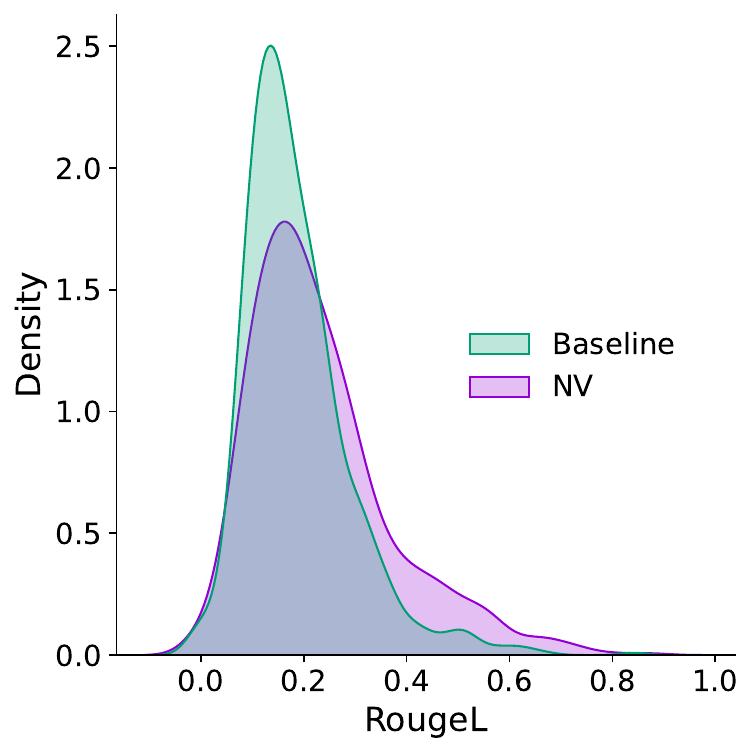}
\end{minipage}
\hfill
\begin{minipage}[t]{.24\textwidth}
\includegraphics[width=\textwidth]{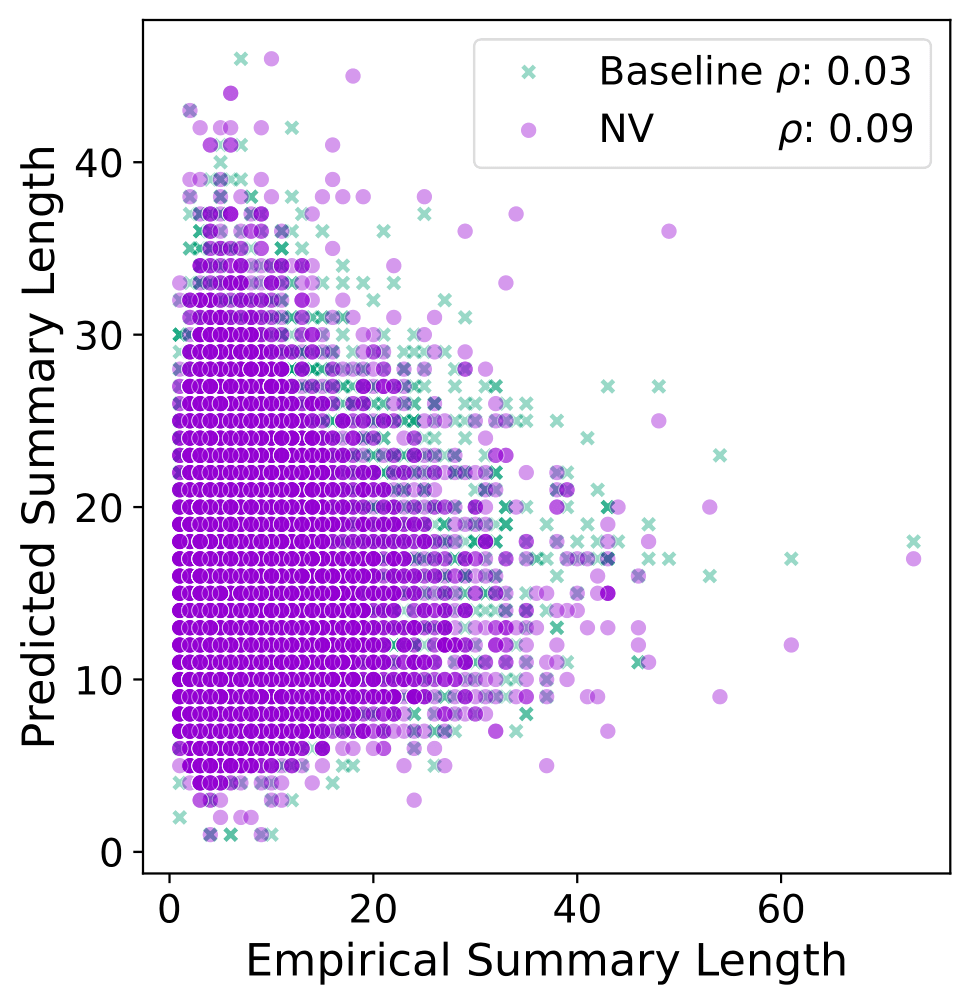}
\end{minipage}
\caption{A baseline BART (Xsum) is compared against the same reinterpreted model with NVIB regularisation on \textbf{Top-Bottom} Curation Corpus, CNN/DailyMail, SAMsum and WikiHow validation datasets. \textbf{Left-Right}: Rouge-1, Rouge-2, and Rouge-L score distributions; Empirical vs predicted summary length with Spearman's correlation.}
\label{fig:rouge_length_all}
\end{figure}

In Figure \ref{fig:rouge_length_all}, we compare each predicted validation summary to its corresponding gold empirical summary and report the distribution of Rouge overlap scores and correlations in length. Comparing these distributions for the baseline and for the NV-Transformer tells us in what way we are getting improvements. Comparing the lengths tells us how well the information content of the summary is being modelled. The summary is a compressed and high information version of the original document. The summary length is a measure of information in the document such that a longer summary means more information in a summary, given the compression ratio of the dataset. We compare the predicted and true summary length and compute their spearman's correlation coefficent. If the summary lengths are more correlated than the baseline, it shows the regularisation is adjusting the length according to the information in the article. 

Figure \ref{fig:rouge_length_all} shows plots for the model trained on Xsum. The first two rows show that NVIB regularisation increases the performance on the Curation Corpus and CNN/DailyMail datasets and is producing longer, more accurate summaries, but the adaptation to the information in the document is similar to the baseline. However, NVIB regularisation is resulting in longer summaries than the BART (Xsum) model, which better reflects in absolute terms the much longer summaries in the Curation Corpus dataset (See Table \ref{tab:mean_summary}). The last two rows show that NVIB regularisation increases the performance on the SAMsum and WikiHow datasets and produces shorter, more accurate summaries that are more adaptive to the information in the document. 

\begin{table}[h]
  \caption{Mean number of words in validation set summaries.}
  \label{tab:mean_summary}
  \centering
  \begin{tabular}{l|ccccc}
    \toprule
     & \multicolumn{4}{c}{Mean summary words} \\
    \textbf{Model} &
    \textbf{CC} &  \textbf{CNN/DM} & \textbf{SAMsum}
    & \textbf{WikiHow} & \textbf{Xsum} \\
    \midrule
    Empirical  & 85.18 & 57.91  & 20.28  & 6.46 & 21.12 \\
    \midrule
    BART \scriptsize{(Xsum)} & 19.26 & 20.71 & 17.43  & 16.42 & 18.90 \\
    NV-BART \scriptsize{(Xsum)}   & 28.17 & 30.79  & 16.62  & 14.42 & 18.63 \\
    \bottomrule
  \end{tabular}
\end{table}

\section{Additional Analyses}

\subsection{Empirical Prior Analysis} \label{sec:empirical_prior_analysis}

% \paragraph{What do the priors look like?}
% In distribution - the encoder means are around zero and variances are close approximately 0.1. The cross-attention means are close to zero with lower variance 0.02. The decoder mean still appear to be centered around zero but with higher variance that grows throughout the layers from 0.2 until 0.6.

% Out of distribution - When the BART CNN model is given out of distribution data (eg XSUM) the encoder self-attention and cross-attention behave similarly however the variance in the decoder grows substantially. For example the variance starts around 1 and gets to 12 in the final layers. This then makes the L2-norm of mean very large and it focuses heavily on the prior. Wikihow is even larger variance all the way up to 65! Curation is low. These large variances in the vectors mean that often the mean may have a large L2-norm and thus the alphas get large too. When the alpha is too large we get that overflow problem.

We conduct an analysis of how the distribution over vectors changes through the layers and with different datasets, by calculating the empirical prior parameters for each distribution. We consider the BART model trained on CNN/DailyMail and Xsum and calculate the empirical priors ($\boldsymbol{\mu}^p$, $\boldsymbol{\sigma}^p$, ${\alpha}^p_0$) given different in- and out-of-domain datasets.  The empirical prior used for NVIB regularisation is the one computed in-domain. We also include a standard normal Gaussian prior in the plots for reference. We plot the average embedding value across all layers, where the last layer of the encoder (Encoder layer 13) is the embedding for cross-attention.

\begin{figure}[h]
    \begin{minipage}{.31\textwidth}
\includegraphics[width=\textwidth]{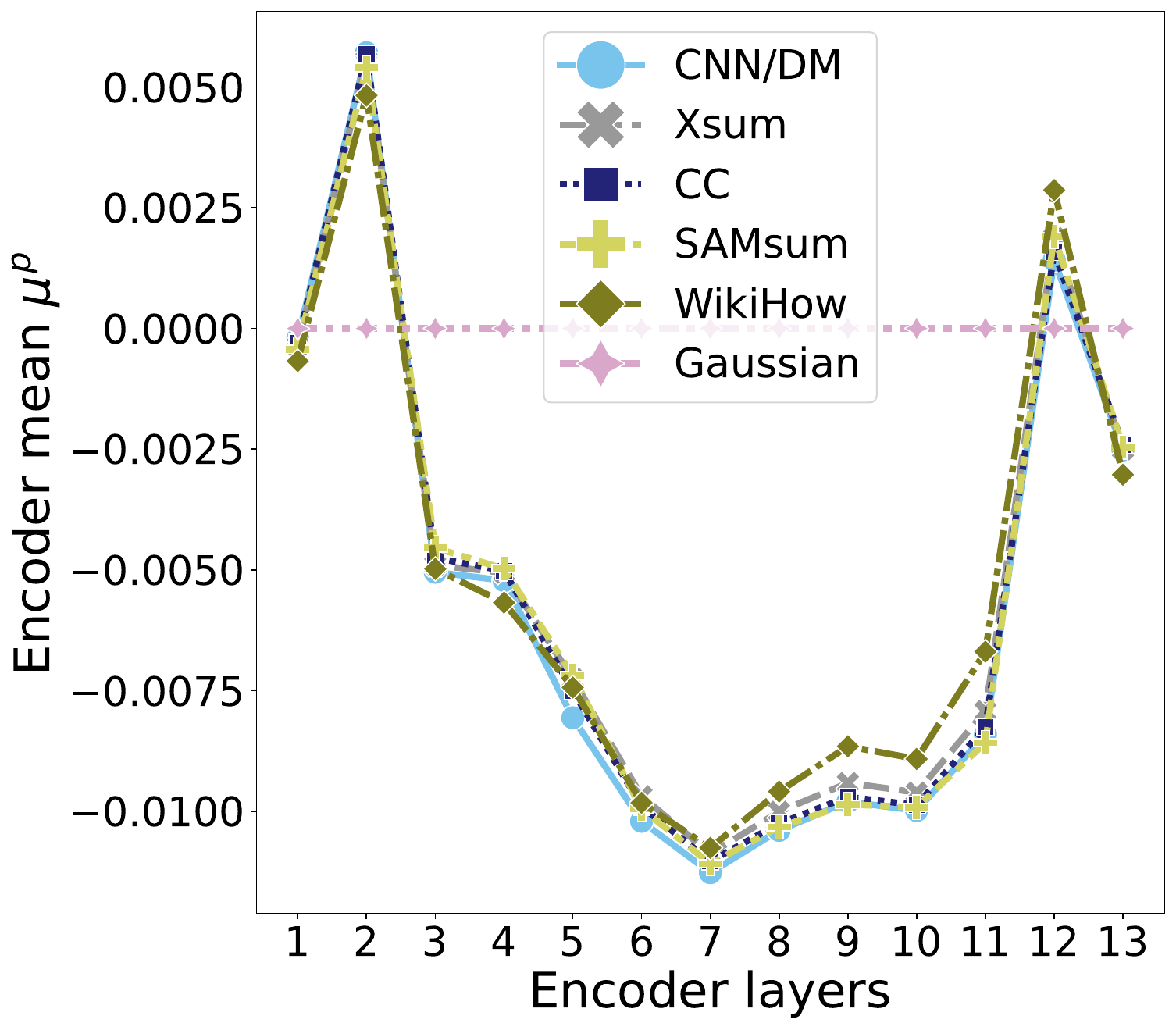}
\end{minipage}
\hfill
\begin{minipage}{.31\textwidth}
    \includegraphics[width=\textwidth]{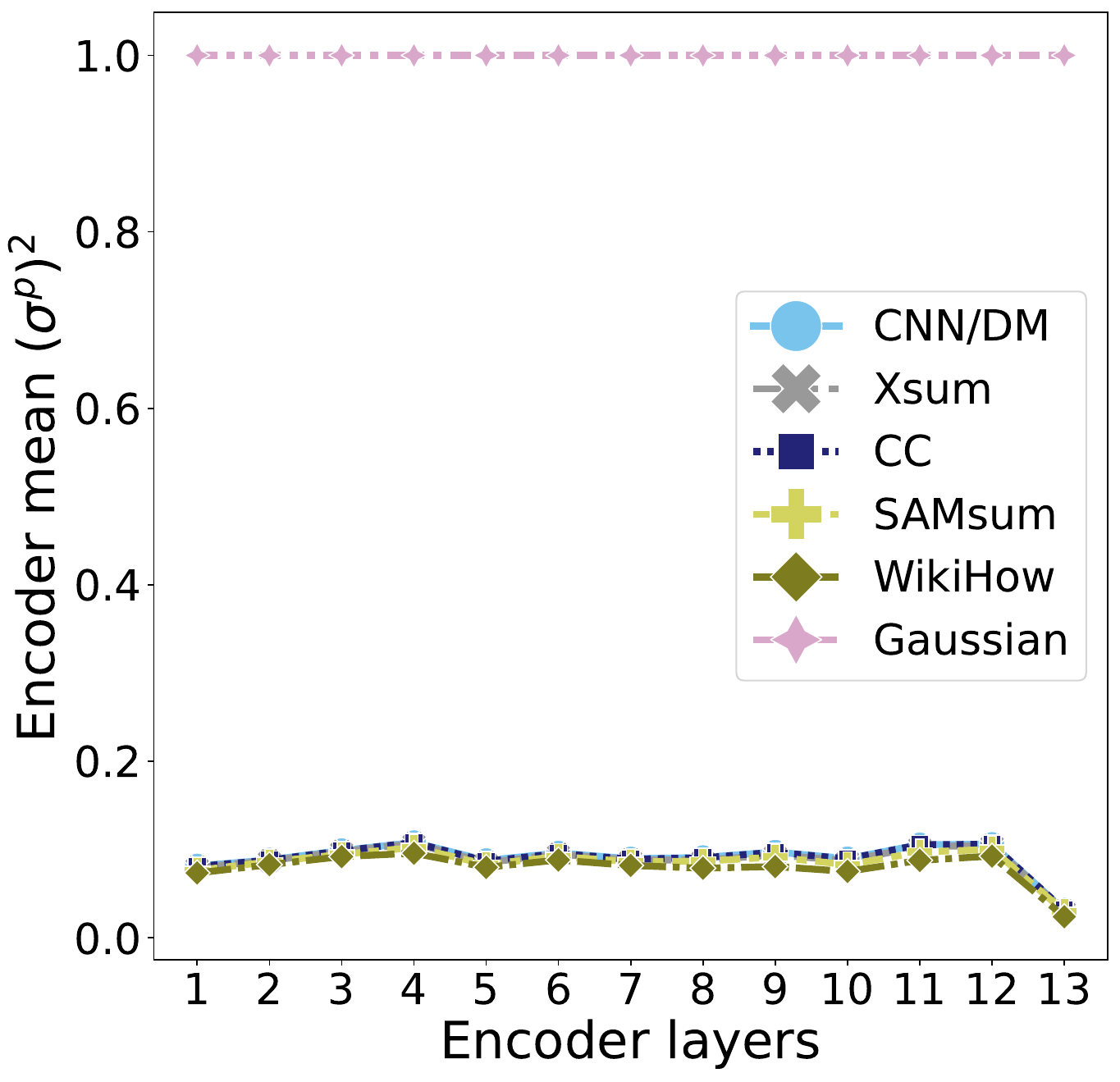}
\end{minipage}
\hfill
\begin{minipage}{.31\textwidth}
    \includegraphics[width=\textwidth]{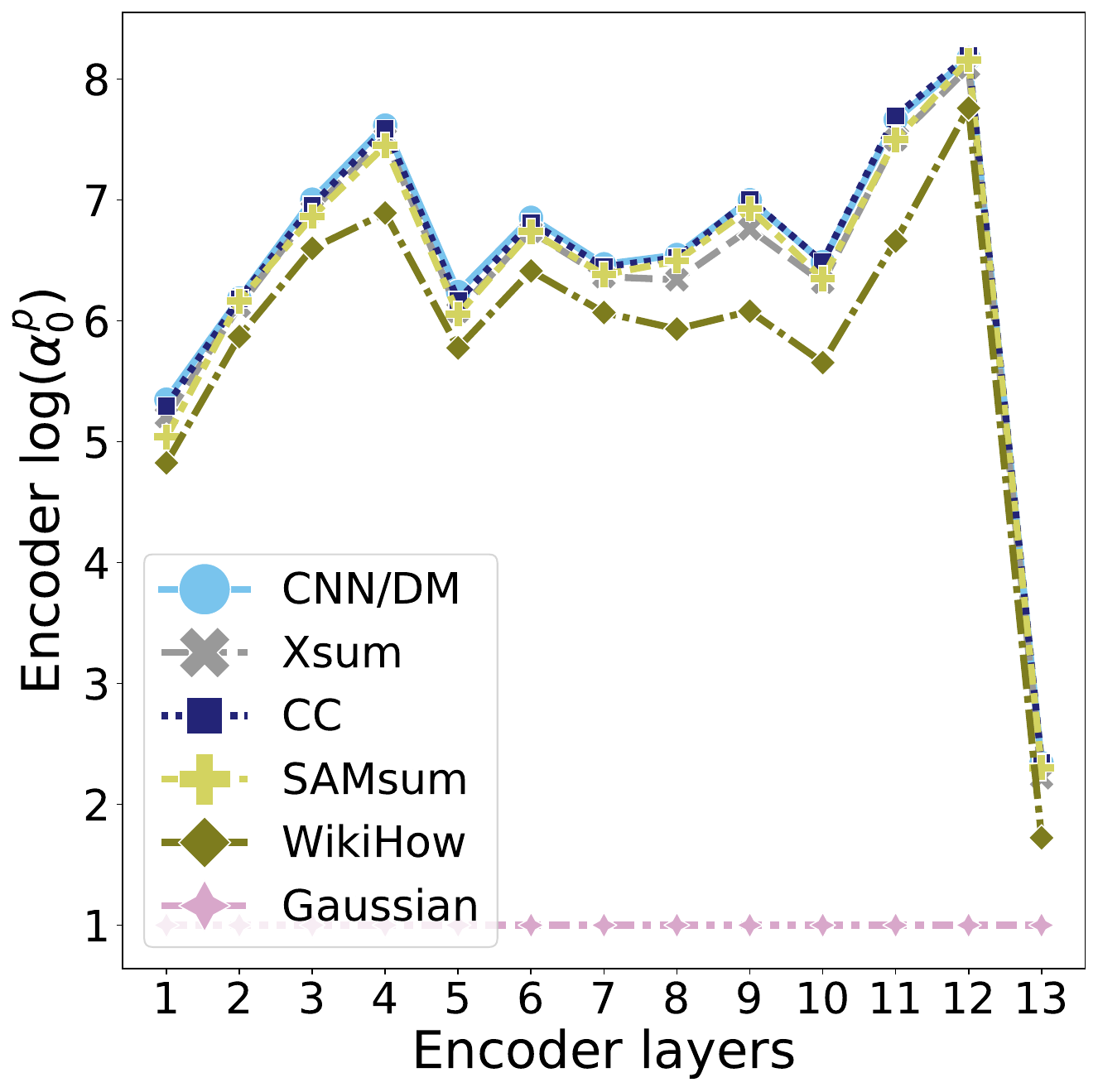}
\end{minipage}
\hfill
    \begin{minipage}{.31\textwidth}
\includegraphics[width=\textwidth]{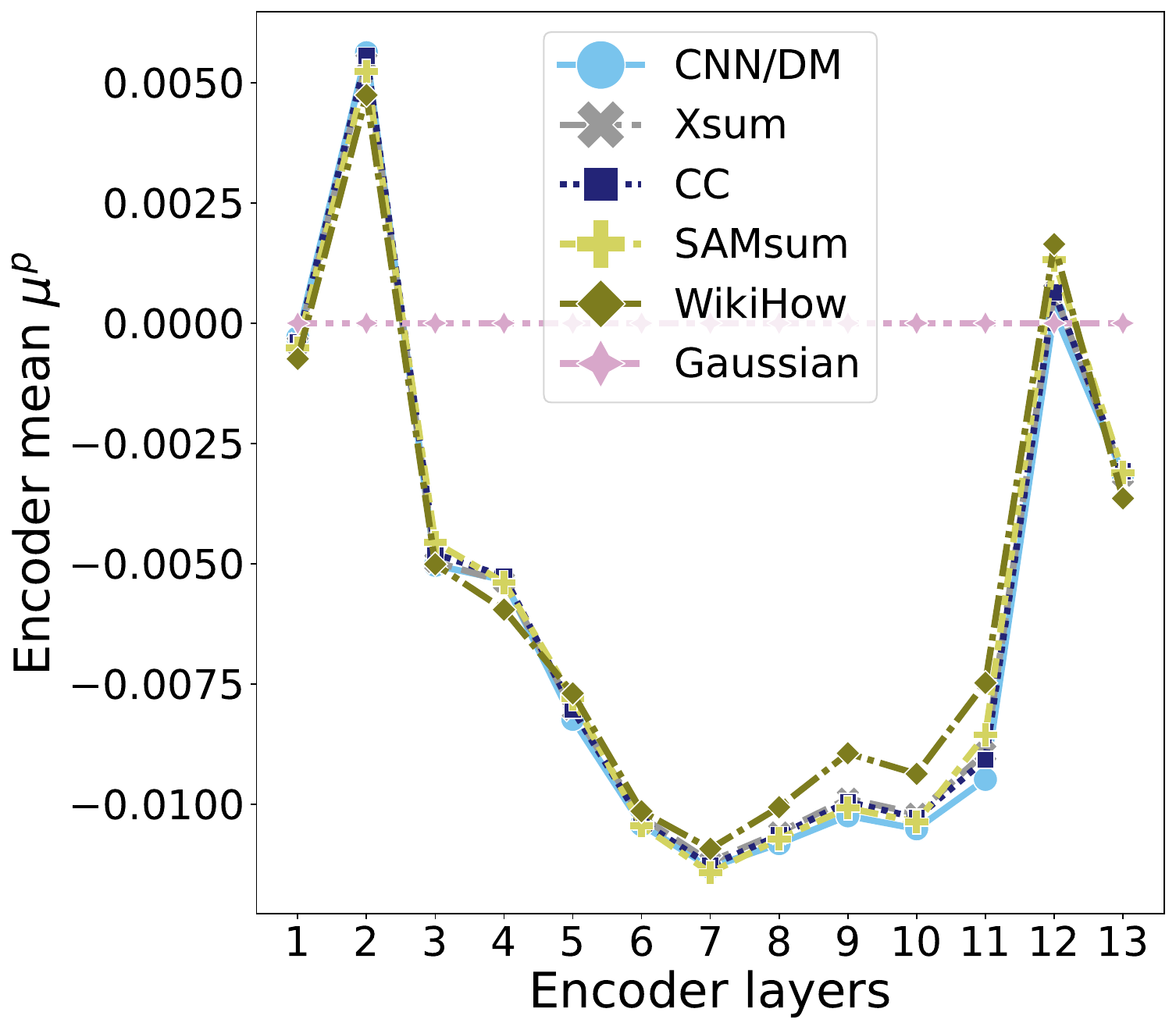}
\end{minipage}
\hfill
\begin{minipage}{.31\textwidth}
    \includegraphics[width=\textwidth]{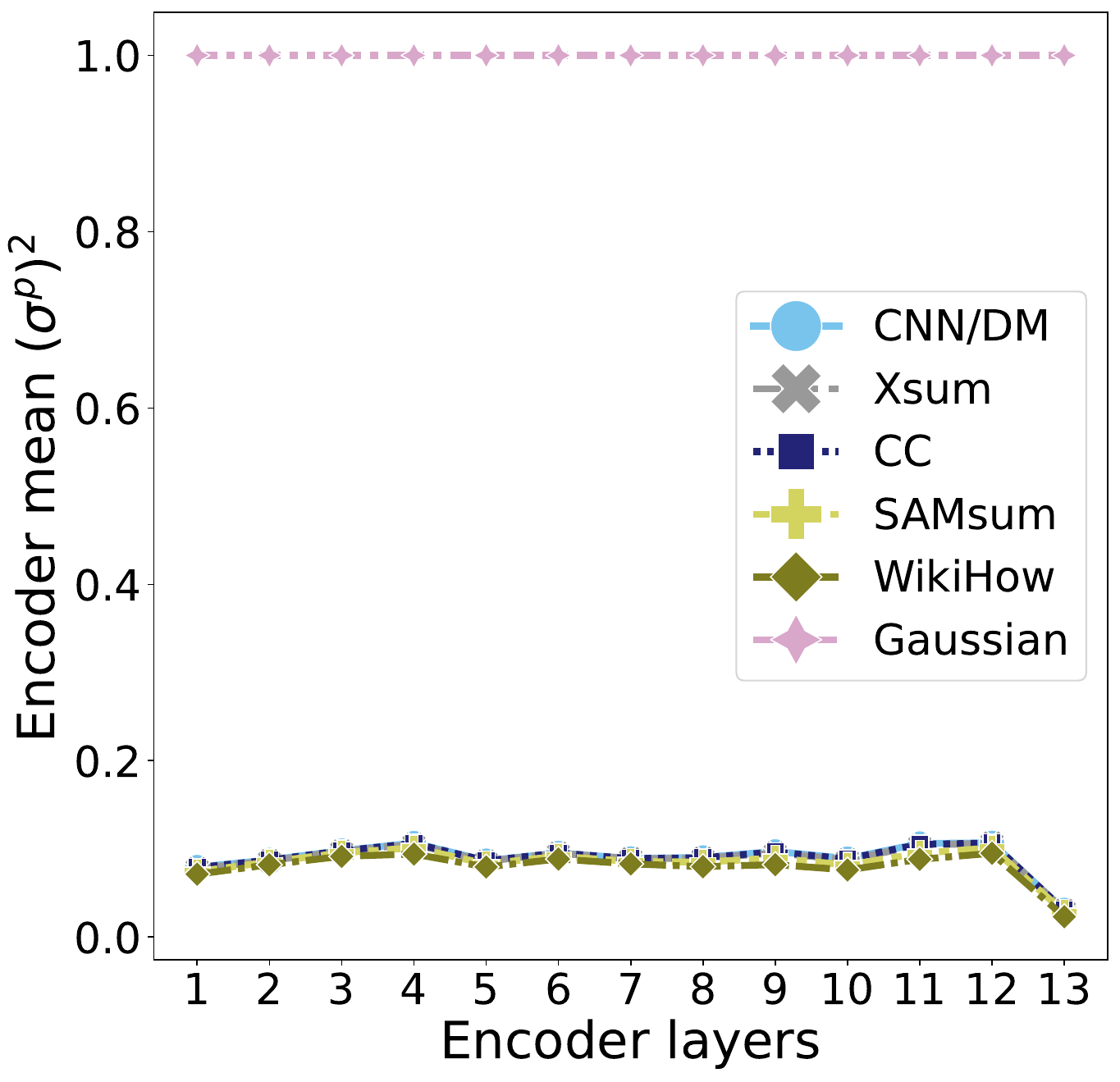}
\end{minipage}
\hfill
\begin{minipage}{.31\textwidth}
    \includegraphics[width=\textwidth]{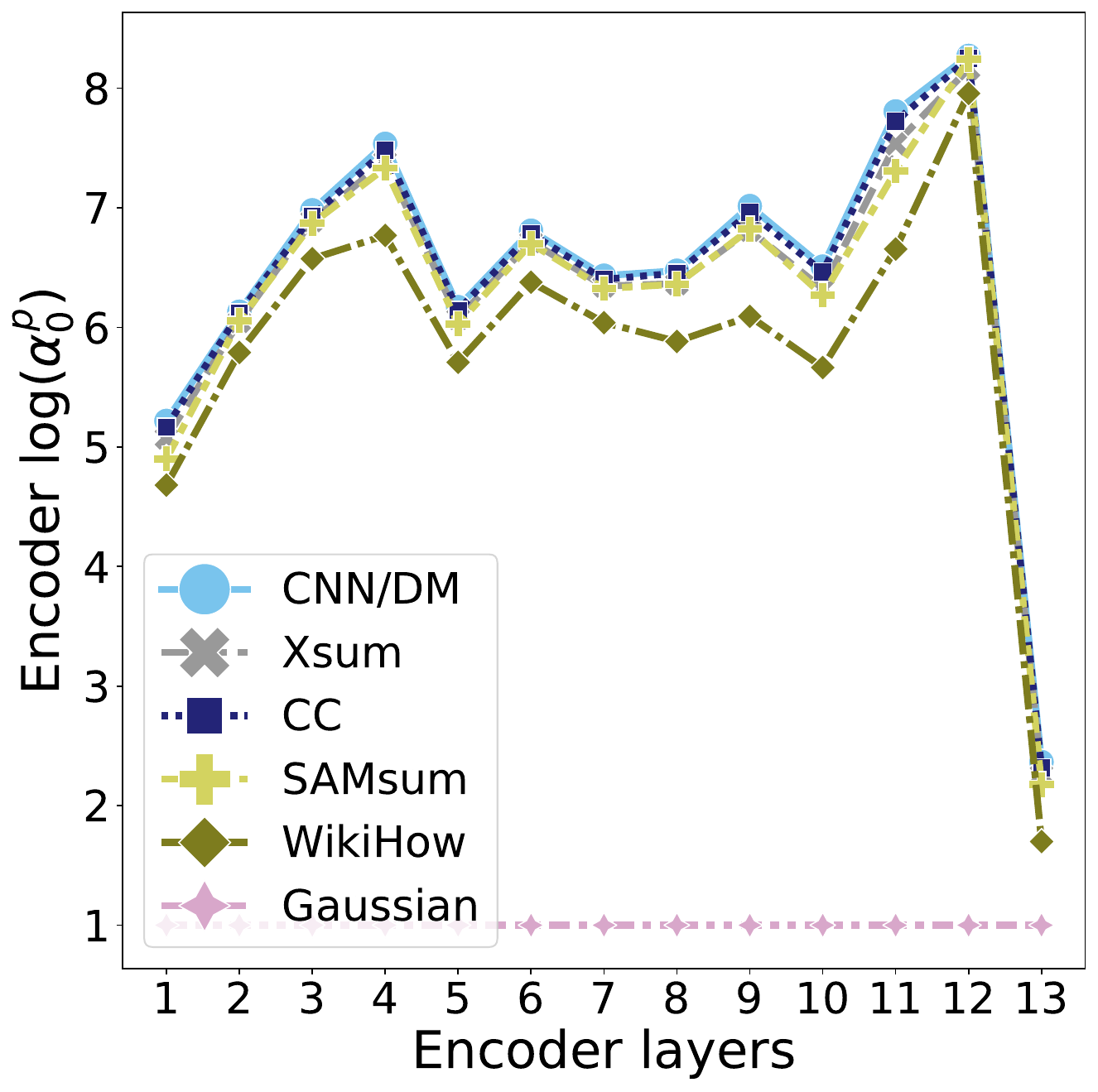}
\end{minipage}
\caption{\textbf{Top}: BART fine-tuned on CNN/DailyMail. \textbf{Bottom}: BART fine-tuned on Xsum. Averaged empirical embeddings per layer \textbf{Left:} encoder mean component $\boldsymbol{\mu}^p$. \textbf{Middle:} encoder variance $(\boldsymbol{\sigma}^p)^2$. \textbf{Right:} encoder logged pseudo-count log($\alpha^p_0$).}
\label{fig:empirical_encoder}
\end{figure}

% %%%%%%%%%%%%%%%%%%%

\begin{figure}[t]
    \begin{minipage}{.31\textwidth}
\includegraphics[width=\textwidth]{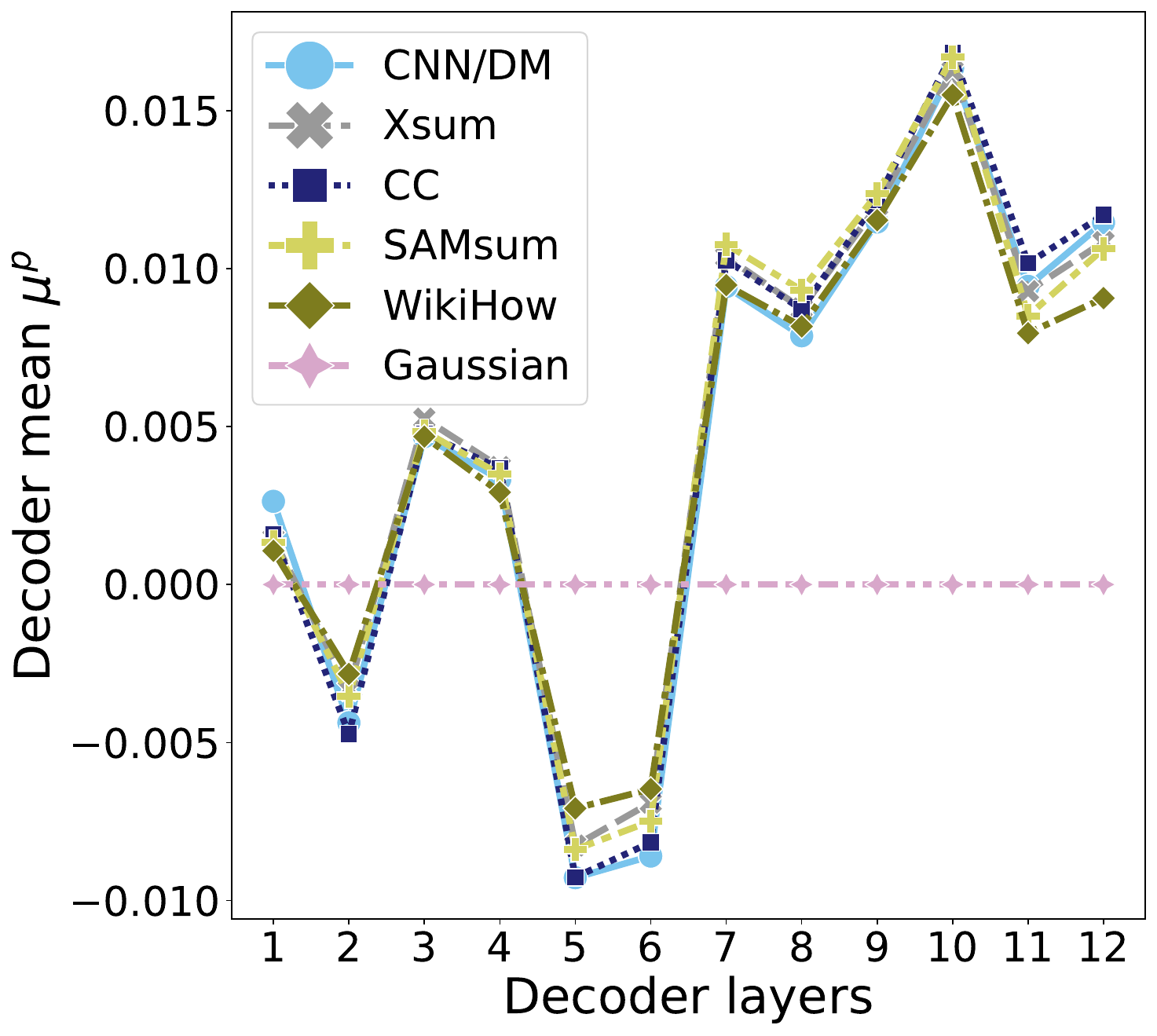}
\end{minipage}
\hfill
\begin{minipage}{.31\textwidth}
    \includegraphics[width=\textwidth]{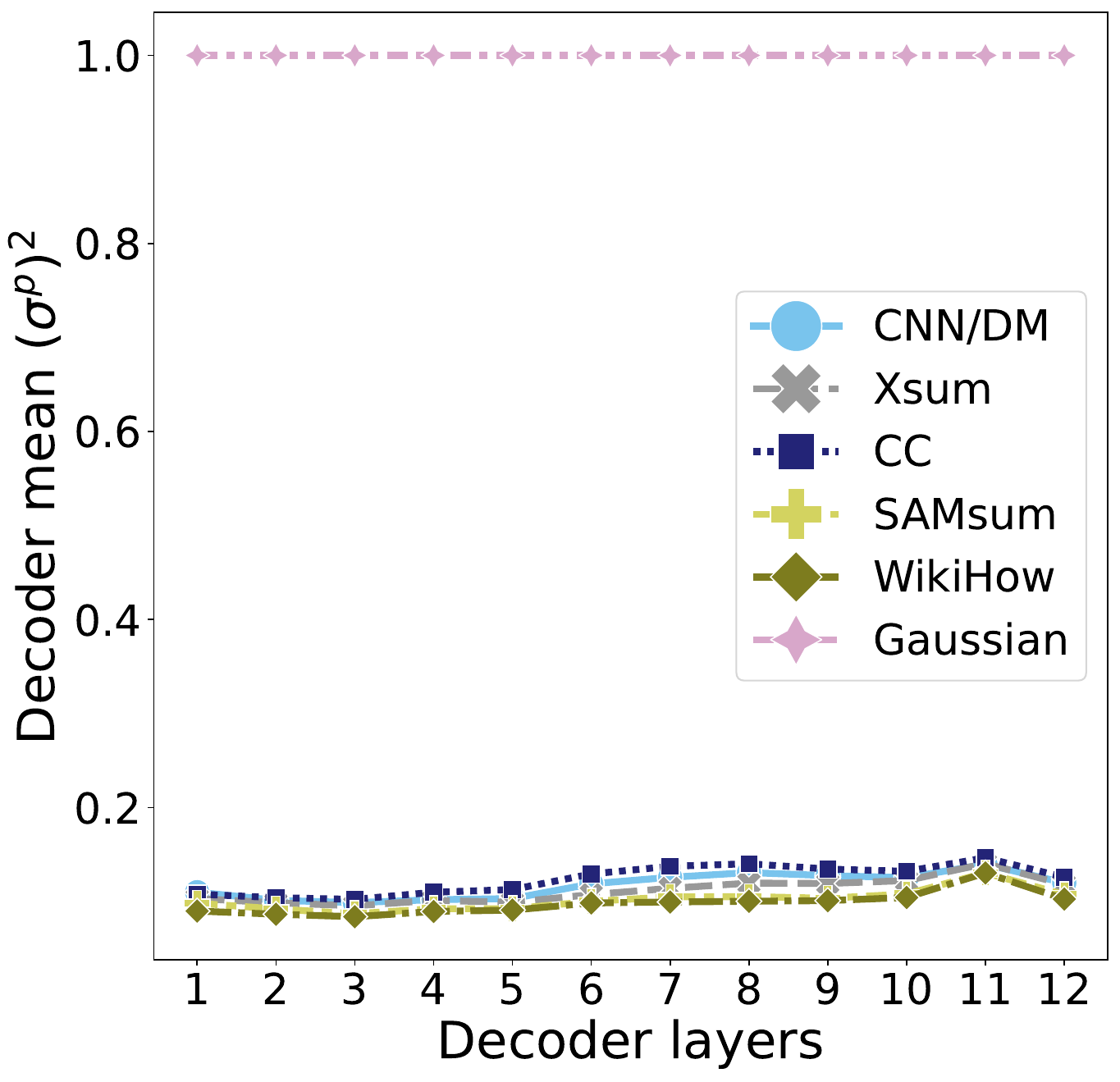}
\end{minipage}
\hfill
\begin{minipage}{.31\textwidth}
    \includegraphics[width=\textwidth]{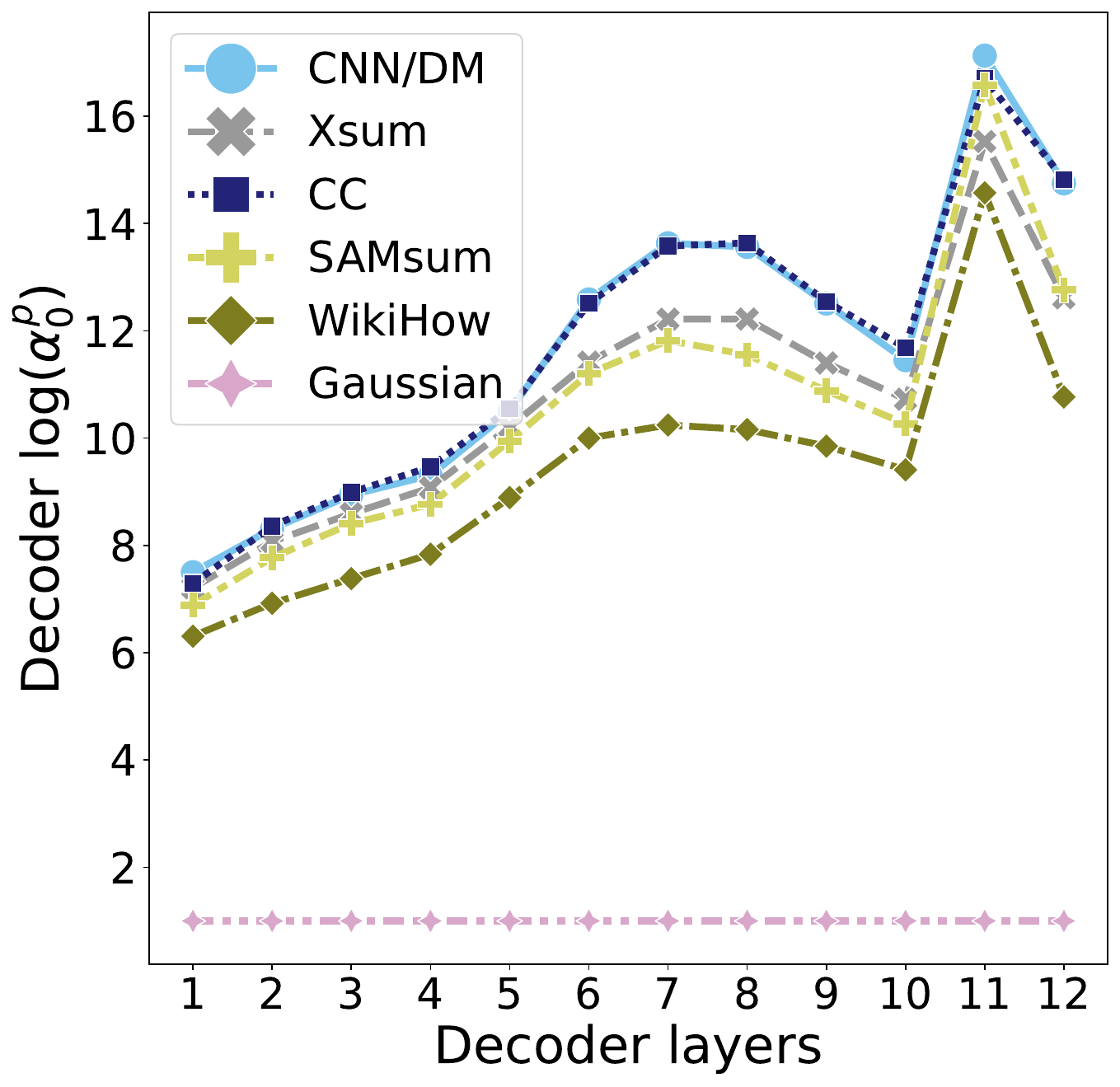}
    % \caption{}
\end{minipage}
\begin{minipage}{.31\textwidth}
\includegraphics[width=\textwidth]{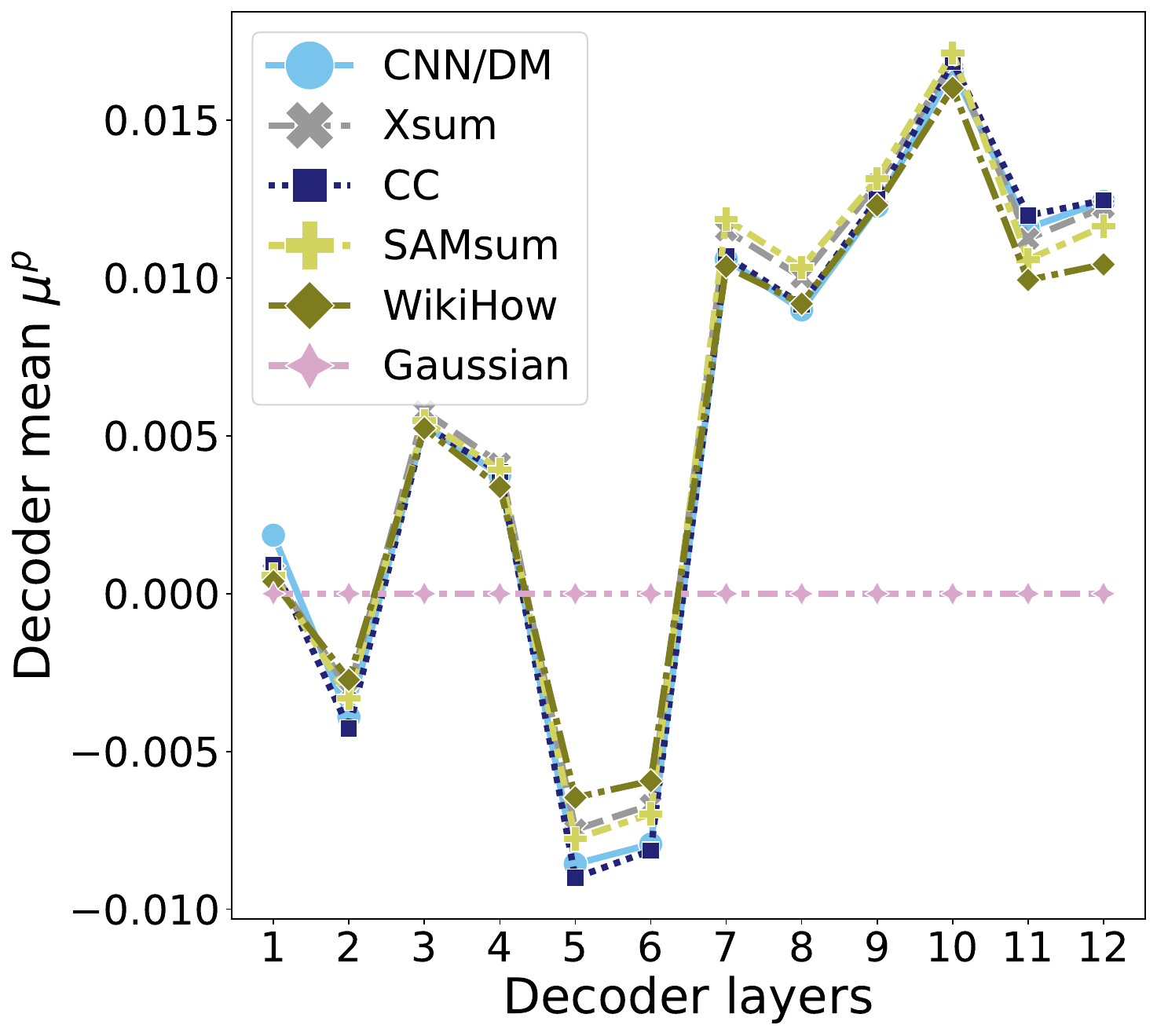}
\end{minipage}
\hfill
\begin{minipage}{.31\textwidth}
    \includegraphics[width=\textwidth]{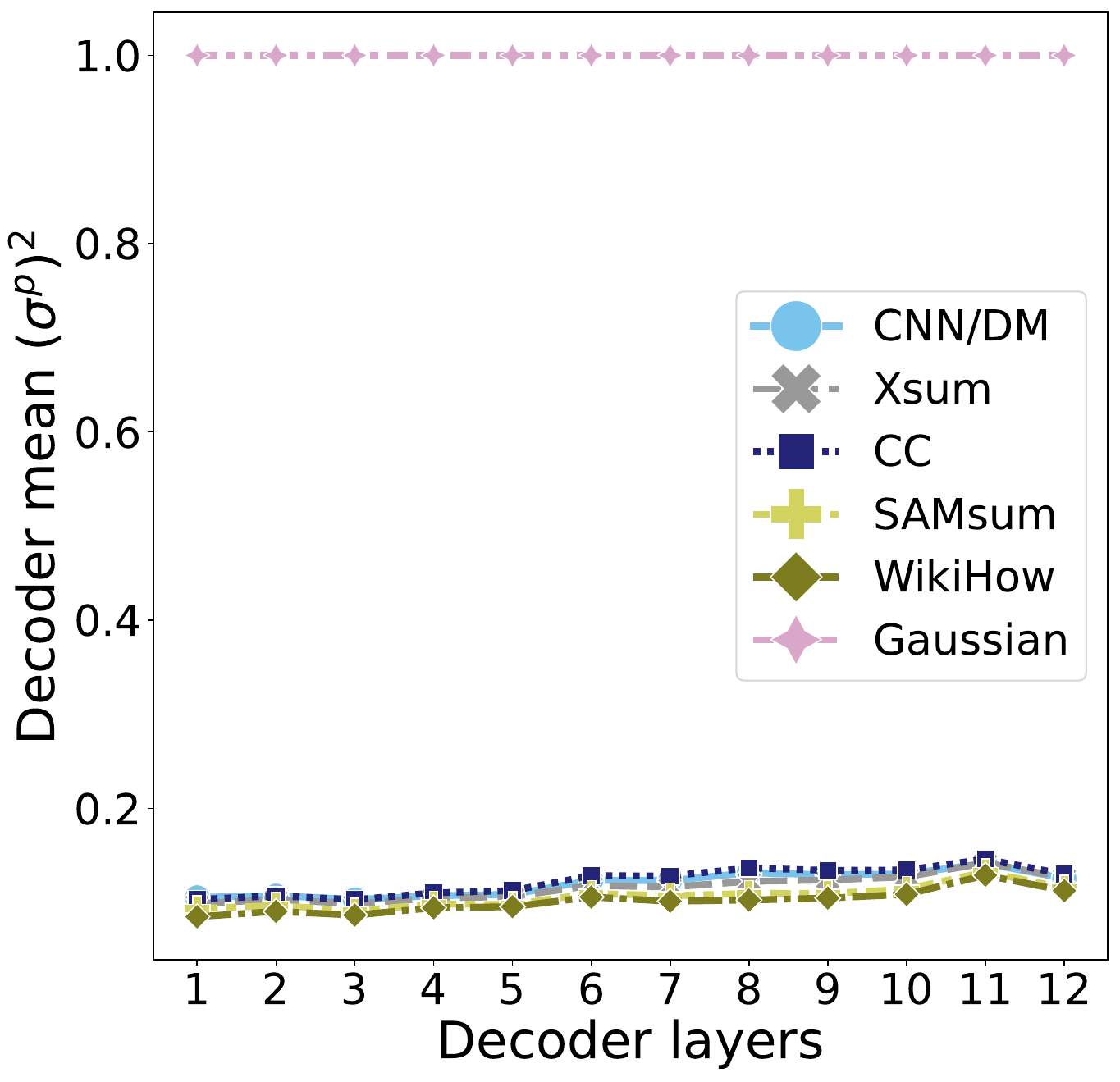}
\end{minipage}
\hfill
\begin{minipage}{.31\textwidth}
    \includegraphics[width=\textwidth]{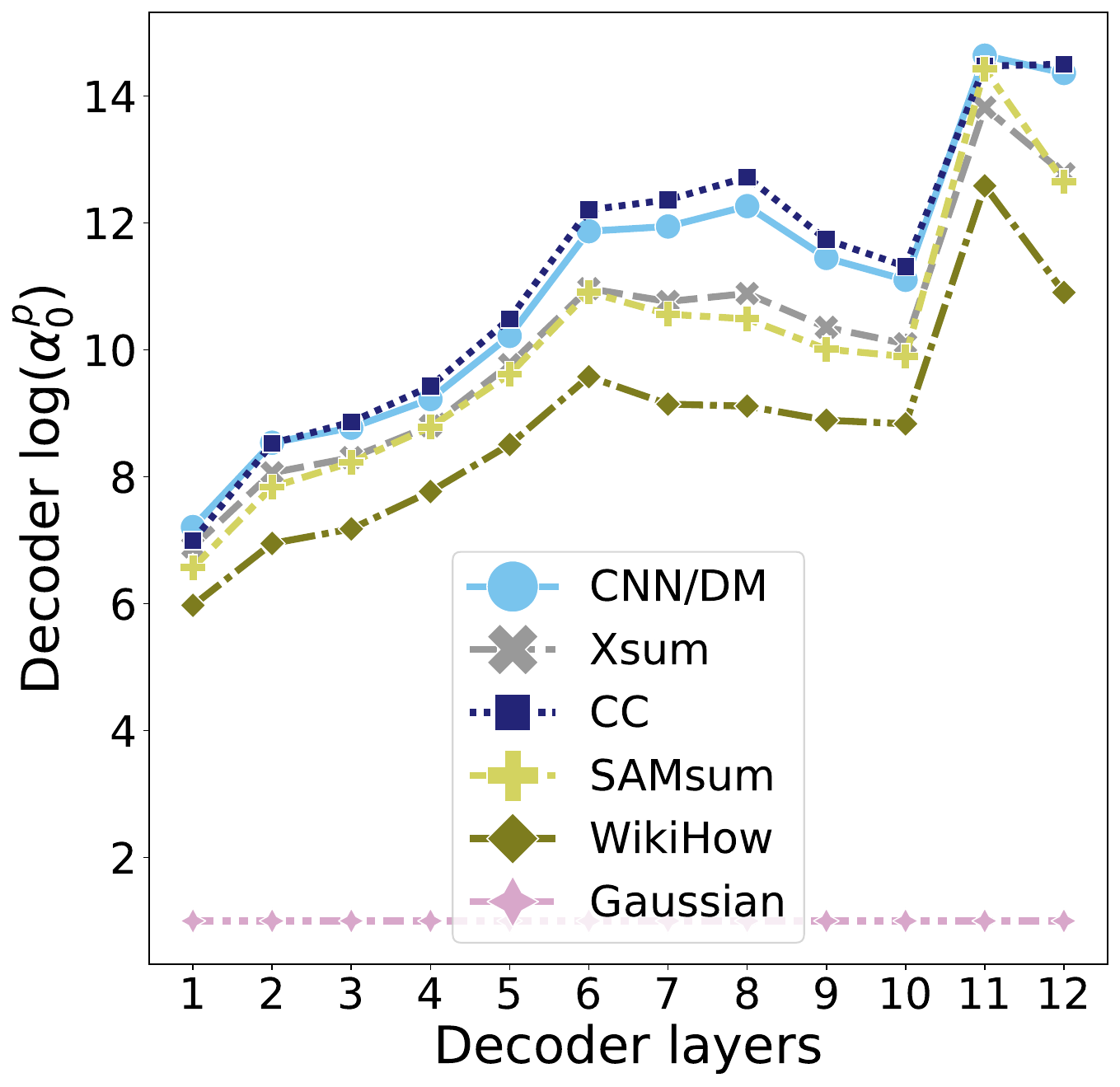}
    % \caption{}
\end{minipage}
\caption{\textbf{Top}: BART fine-tuned on CNN/DailyMail. \textbf{Bottom}: BART fine-tuned on Xsum. Averaged empirical embeddings per layer of \textbf{Left:} decoder mean component $\boldsymbol{\mu}^p$, \textbf{Middle:} decoder variance $(\boldsymbol{\sigma}^p)^2$, \textbf{Right:} decoder logged pseudo-count log($\alpha_0^p$).  "Gaussian" is a unit Gaussian, for reference.
}
\label{fig:empirical_decoder}
\end{figure}

Figure \ref{fig:empirical_encoder} displays the average empirical embedding across all encoder layers for a model fine-tuned on CNN/DailyMail and Xsum, respectively. We notice that the mean is approximately zero and very similar across datasets and across models. The variance is close to 0.1 with the cross-attention (Encoder layer 13) being lower at 0.02. The expectation of the logged pseudo-count is higher and approximately around 7 for encoder layers and lower around 2 for cross attention (Encoder layer 13) with consistent variation across datasets. Its clear the distribution of embeddings between fine-tuned BART encoders is similar. 

Figure \ref{fig:empirical_decoder} shows the average empirical embedding across all decoder layers for a model fine-tuned on CNN/DailyMail and Xsum, respectively. The decoder means have larger values in comparison to the encoder and similar low variance. Considering the logged pseudo-count we notice that they get exceedingly large through the layers which are near double the magnitude to the encoder in log-space.

\subsection{Empirical prior data requirements} \label{sec:empirical_prior_data}

In this section we restrict the amount of data used to create the empirical prior and see if it is still able to maintain the performance increase across the datasets. We plot the Rouge-L performance as a function of the amount of data used to create the prior. We consider the best selected hyperparameters from the validation set (Section \ref{sec:validation_results}) and show the relative performance for all models and all datasets. 

\begin{figure}
    \centering
    \includegraphics[width=0.3\textwidth]{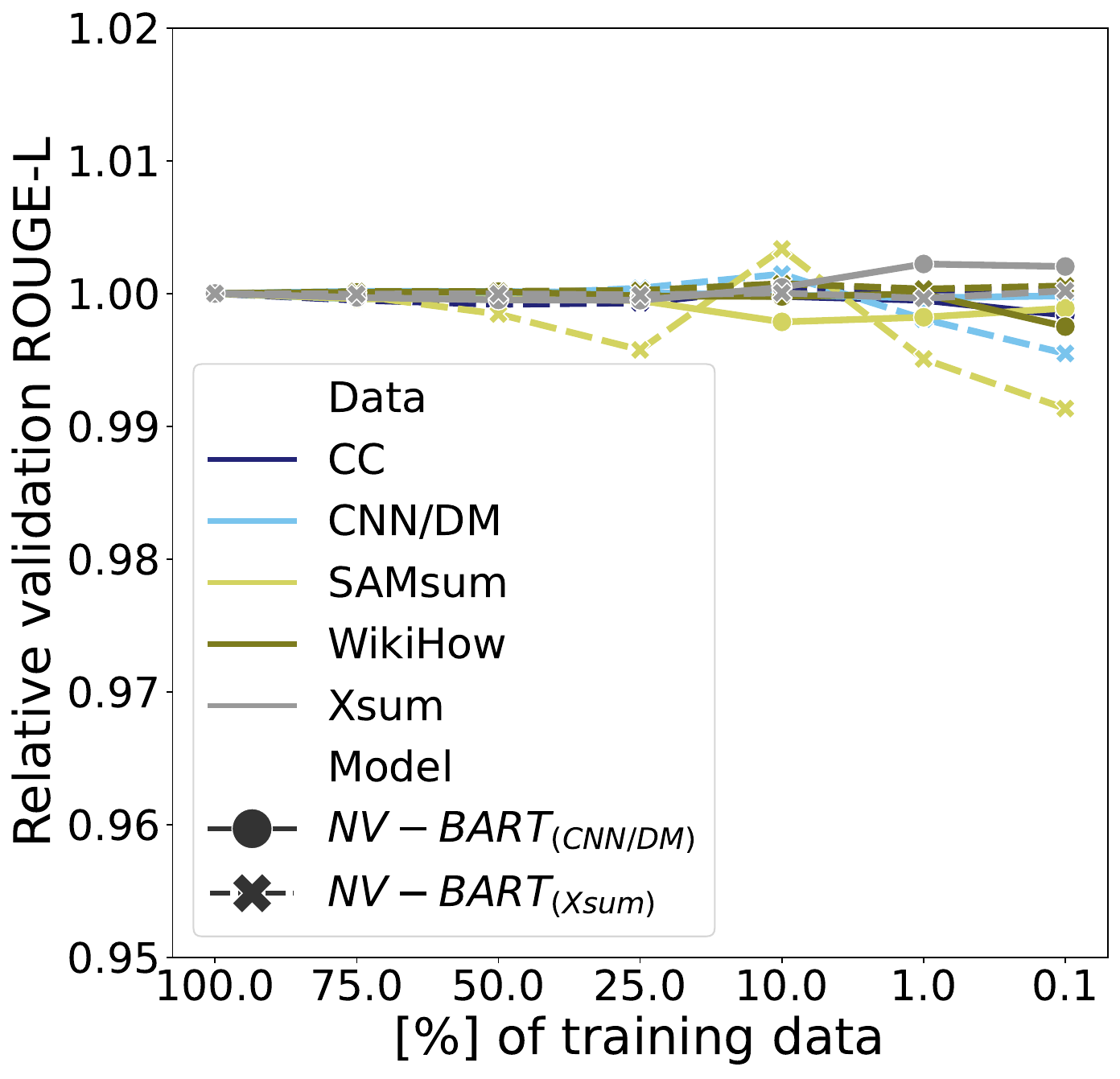}
    \caption{The relative validation performance of Rouge-L ($y$-axis) is compared for different amounts of training data used to create the empirical prior ($x$-axis).}
    \label{fig:low_data}
\end{figure}

Figure \ref{fig:low_data} shows that the empirical prior is data efficient as it requires few examples ($0.1\%$ of training data which is approximately $200$) to achieve good stable performance. 
% \fabio{In this section we want to answer: "How data hungry is the prior?" - If we can show we only need a minor amount of data this steers this towards low resource applications. "Data efficient"}

% Experiment: Calculate the in-domain prior with less data eg: 1\% (2K examples) or 0.1\% (200 examples) now look at its down stream performance with the same hyperparamters. On SAMsum we get very close performance with as few as 200 examples. DATA efficient! - I will make plot for this soon. 

% This begs the question does any prior work? Could we just use a Gaussian prior? Assuming we find the points it breaks. I will show this as a comparison ideally not.

\newpage

\subsection{Attention plots} \label{sec:attention_plots}

In this section we consider the attention plots of our reinterpreted model in the presence of different levels of regularisation. We selected a random validation example from the Curation Corpus dataset and shorten the article document for visualisation purposes. We manually selected the layers with the most regularisation towards the prior for best visualisation. The attention plots are averaged over all heads and the attention scores are displayed with light yellow for 1 and dark purple for 0.

We show an examples of encoder self attention,  decoder cross attention and decoder causal attention for both reinterpreted NV-Transformers. We consider three cases: firstly the identity initialisation $\tau_\sigma^i \approx 0$ and $\tau_\alpha^i = 10$; secondly with regularisation where the $\tau_\sigma^i$ and $\tau_\alpha^i$ are set to the best validation hyperparameters (Appendix \ref{sec:validation_results}); and finally an over-regularised example to show what happens in the presence of collapse to the prior component $[P]$, where the initialisation parameters are $\tau_\sigma^i \approx 0$ and $\tau_\alpha^i =-30$. The prior collapse case can be interpreted as the non-prior pseudo counts being 30 standard deviations smaller than the prior. 

\begin{figure}
    \begin{minipage}{.32\textwidth}
\includegraphics[width=\textwidth]{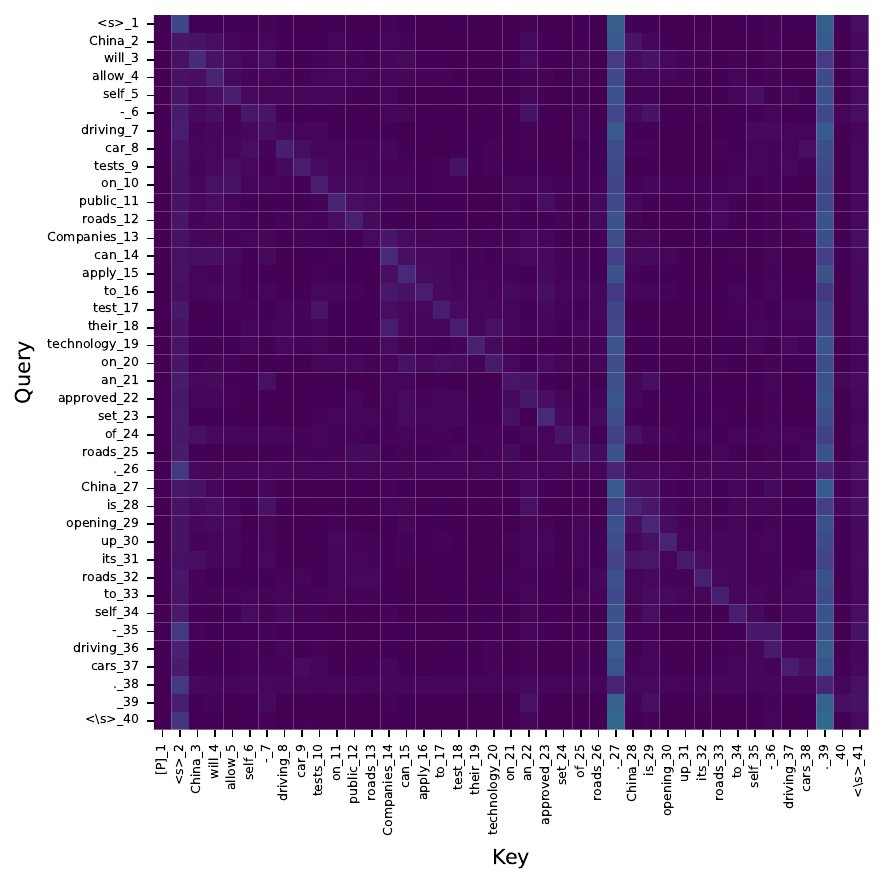}
\end{minipage}
\hfill
\begin{minipage}{.32\textwidth}
    \includegraphics[width=\textwidth]{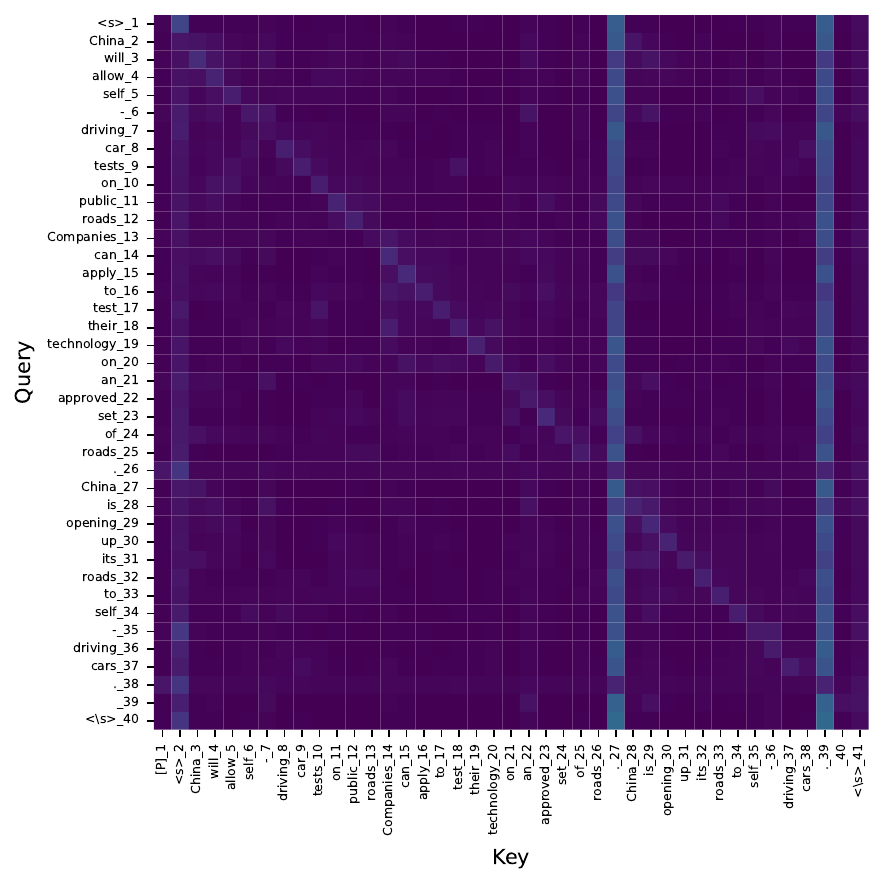}
\end{minipage}
\hfill
\begin{minipage}{.32\textwidth}
    \includegraphics[width=\textwidth]{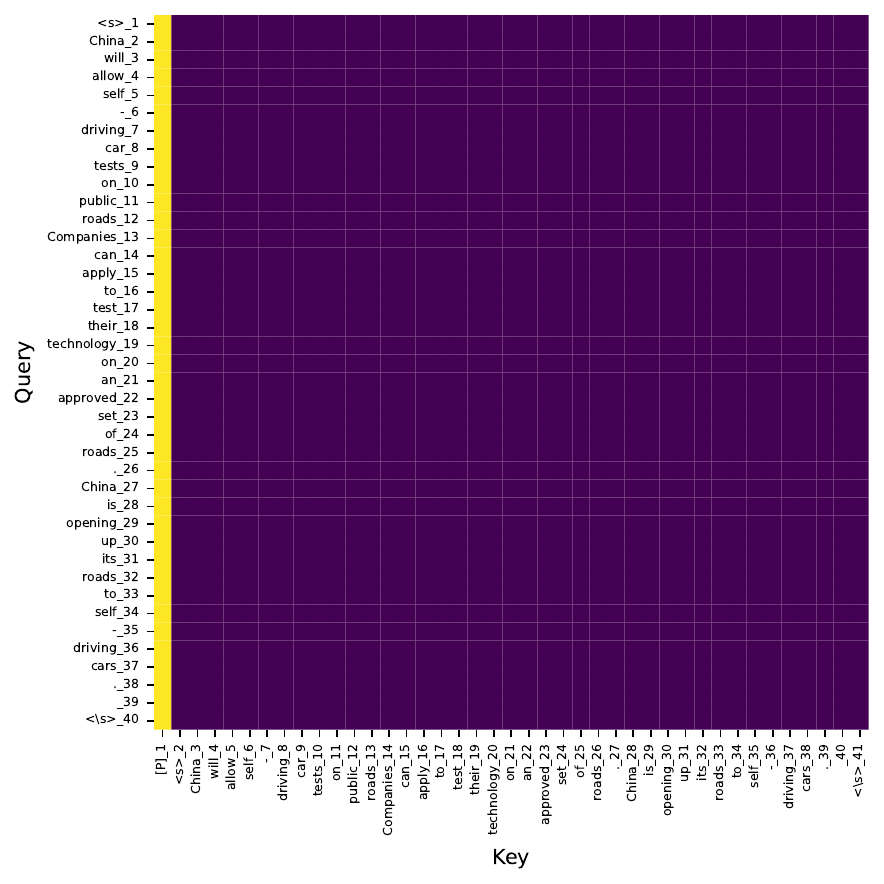}
    % \caption{}
\end{minipage}
% %%%%%%%%%%%%%%%%%%%%%%%%%%%%%
    \begin{minipage}{.32\textwidth}
\includegraphics[width=\textwidth]{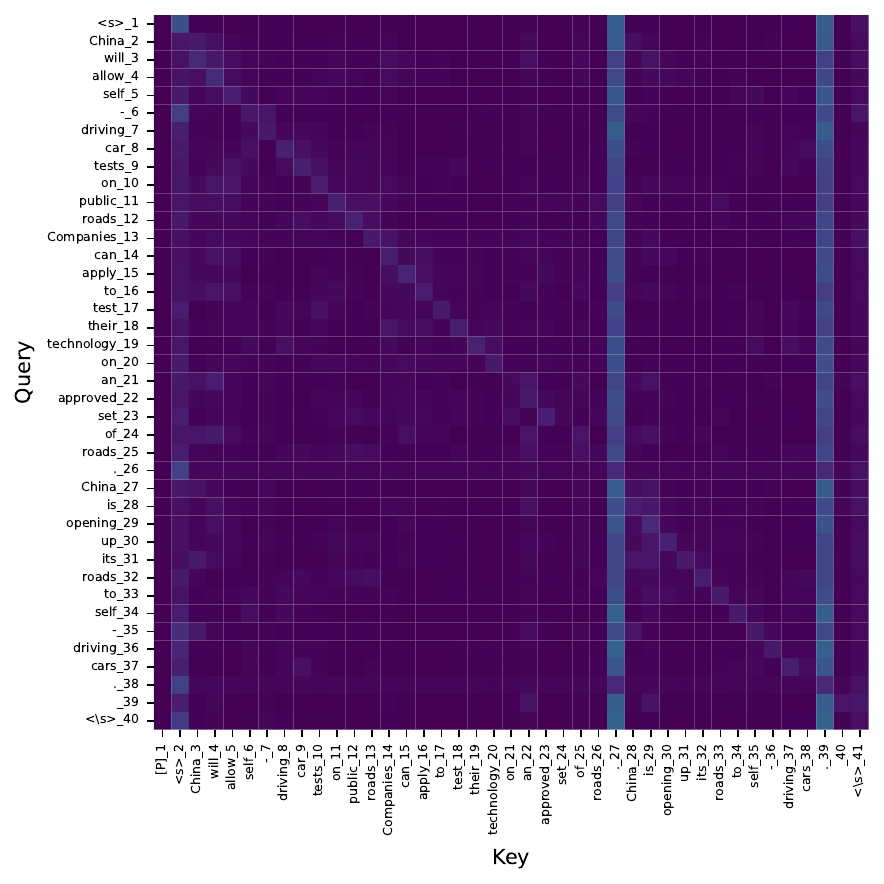}
\end{minipage}
\hfill
\begin{minipage}{.32\textwidth}
    \includegraphics[width=\textwidth]{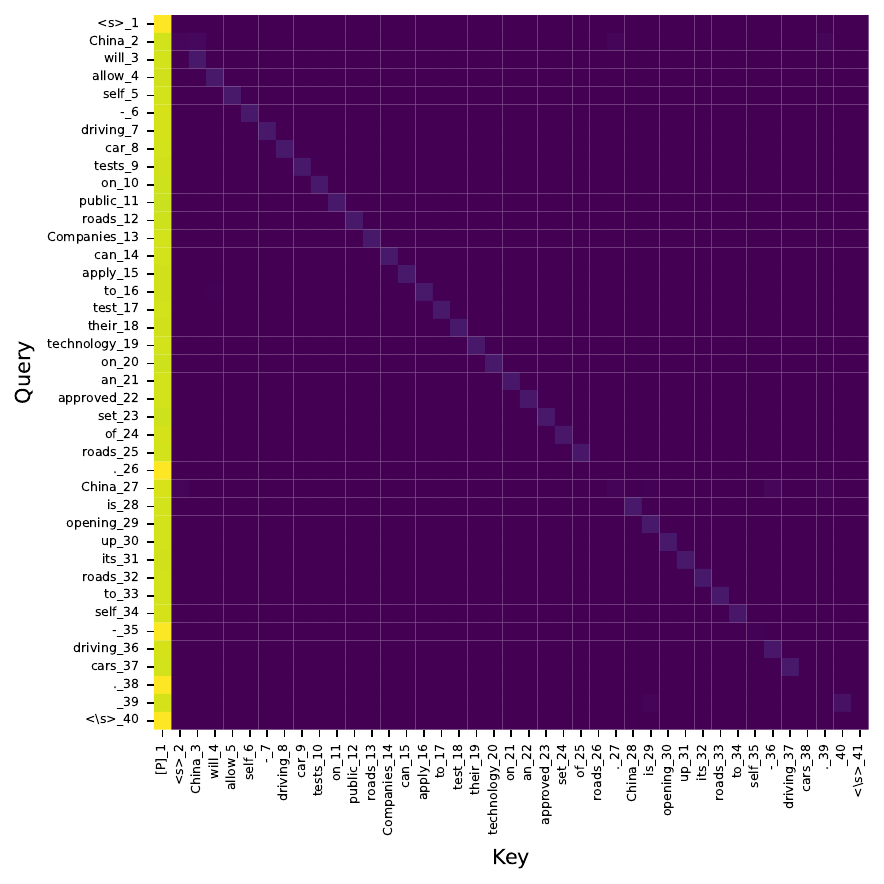}
\end{minipage}
\hfill
\begin{minipage}{.32\textwidth}
    \includegraphics[width=\textwidth]{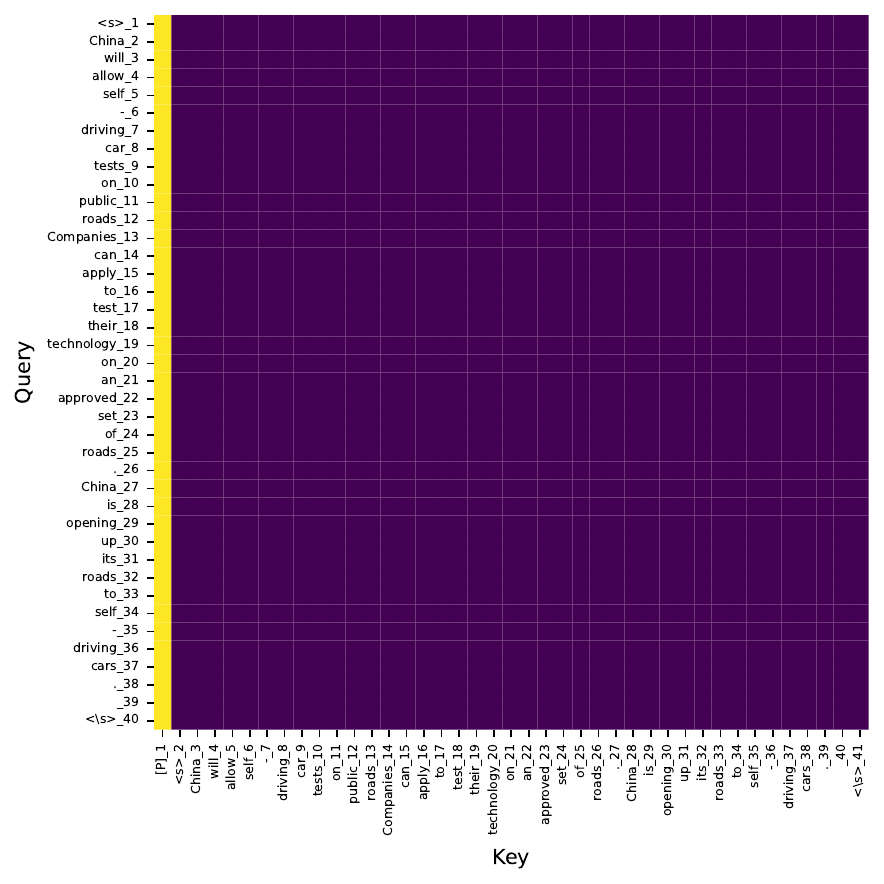}
    % \caption{}
\end{minipage}
\caption{
Encoder at layer 10 averaged over all heads self-attention maps. \textbf{Top:} CNN/Dailymail. \textbf{Bottom:} Xsum. 
\textbf{Left:} Equivalence initialisation. \textbf{Middle:} Best post-training NVIB regularisation from validation set \textbf{Right:} Prior collapse.
}
\label{fig:attention_encoders}
\end{figure}

\begin{figure}
    \begin{minipage}{.32\textwidth}
\includegraphics[width=\textwidth]{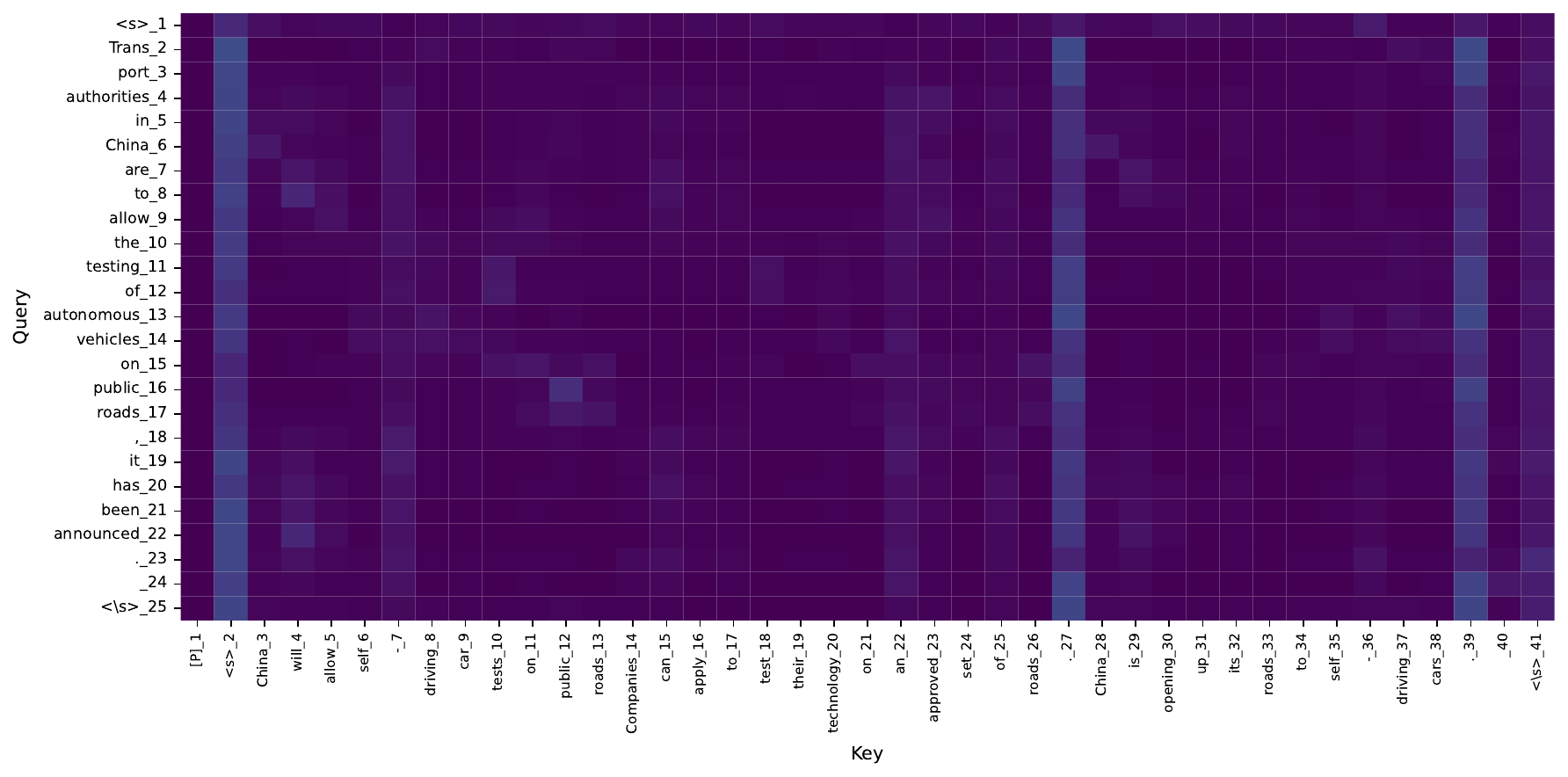}
\end{minipage}
\hfill
\begin{minipage}{.32\textwidth}
    \includegraphics[width=\textwidth]{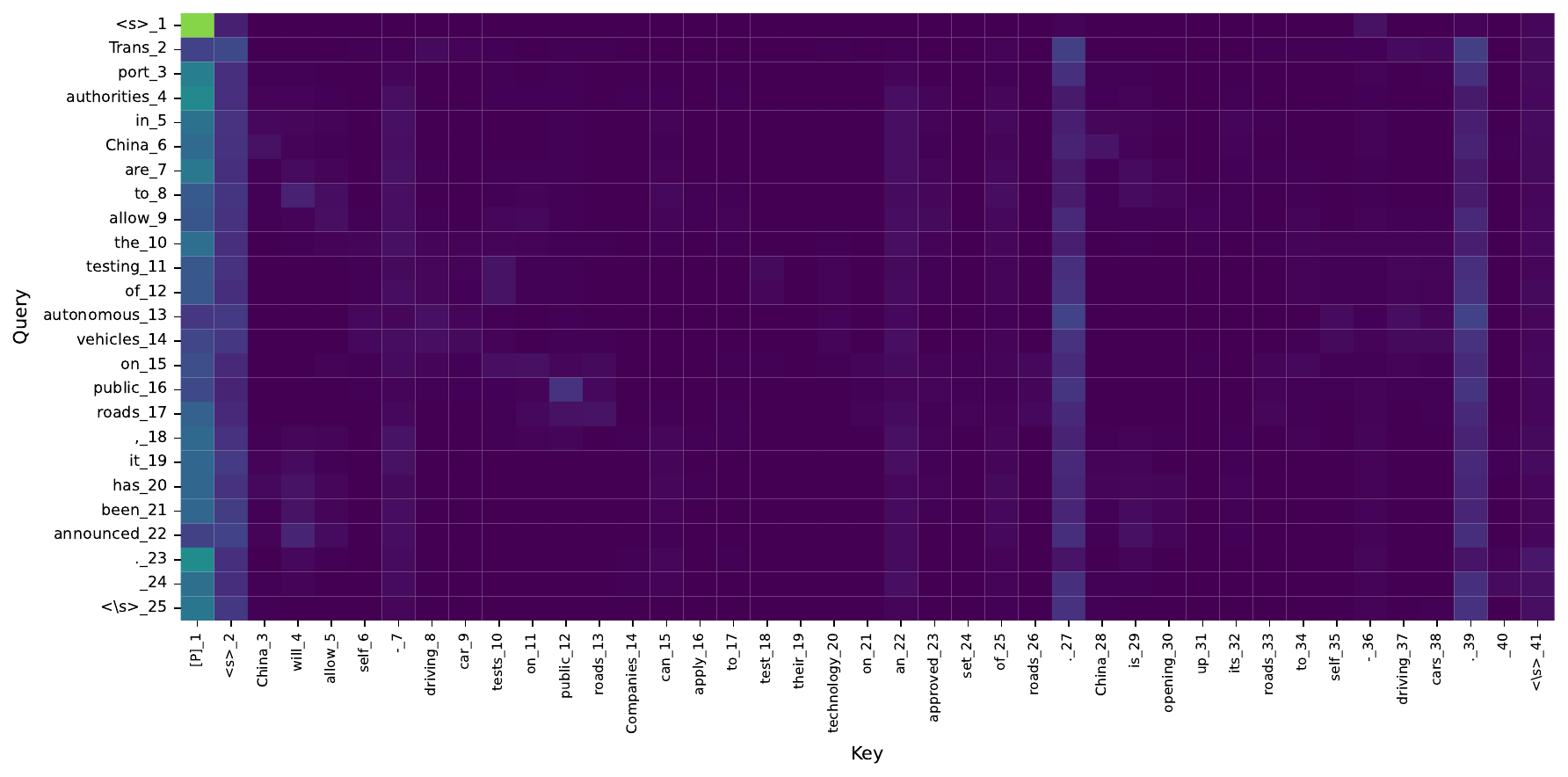}
\end{minipage}
\hfill
\begin{minipage}{.32\textwidth}
    \includegraphics[width=\textwidth]{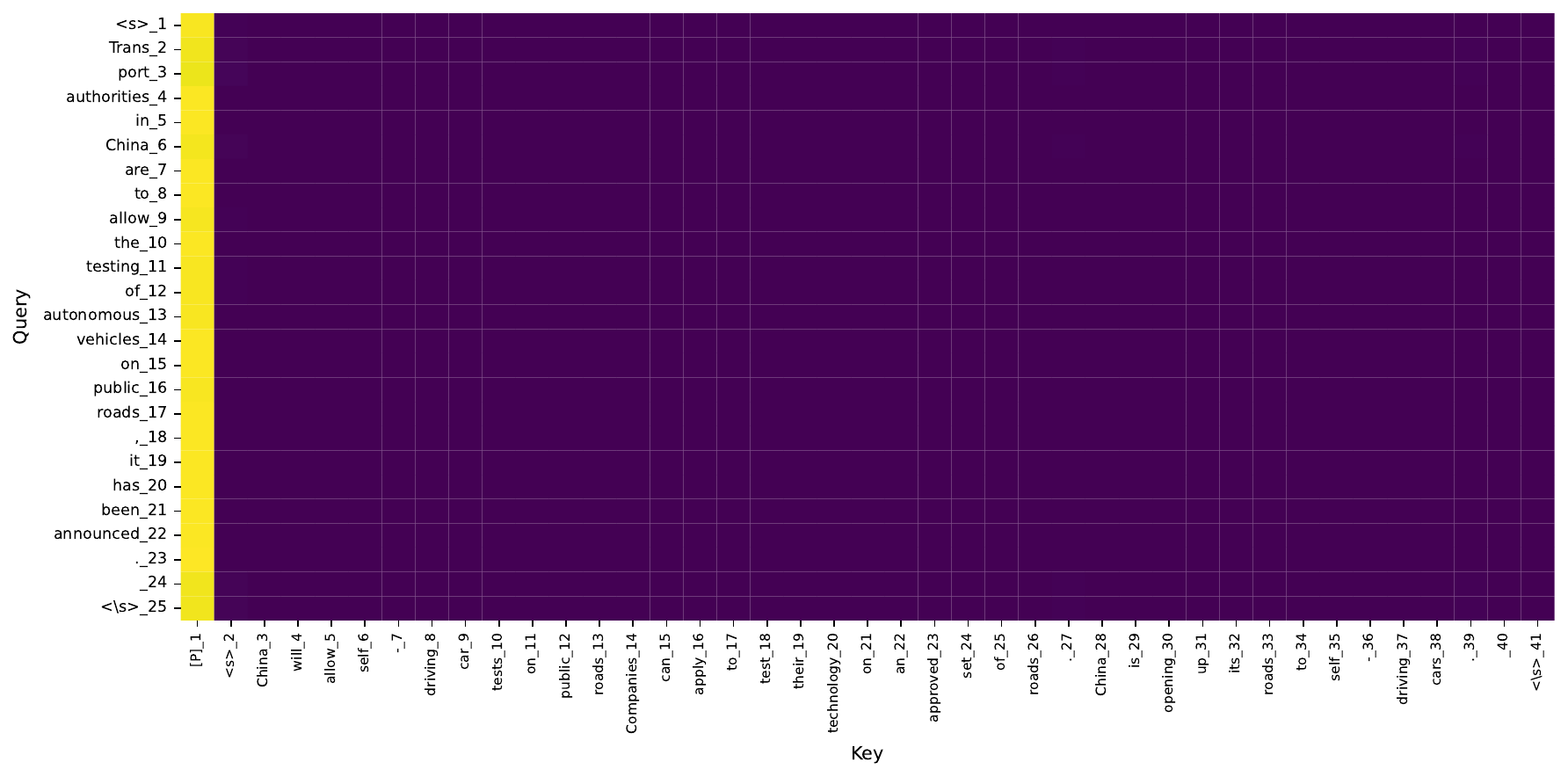}
\end{minipage}
% %%%%%%%%%%%
  \begin{minipage}{.32\textwidth}
\includegraphics[width=\textwidth]{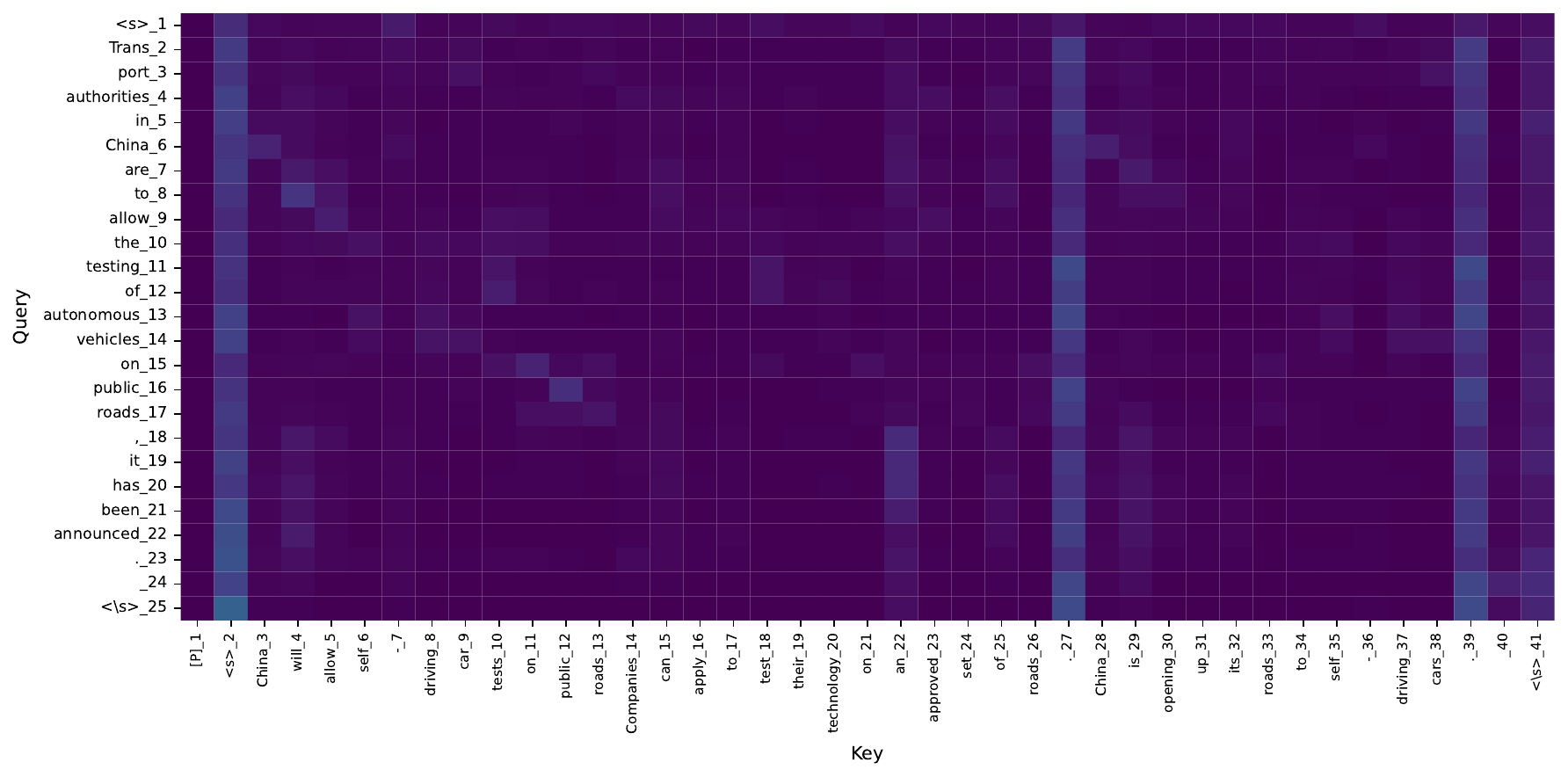}
\end{minipage}
\hfill
\begin{minipage}{.32\textwidth}
    \includegraphics[width=\textwidth]{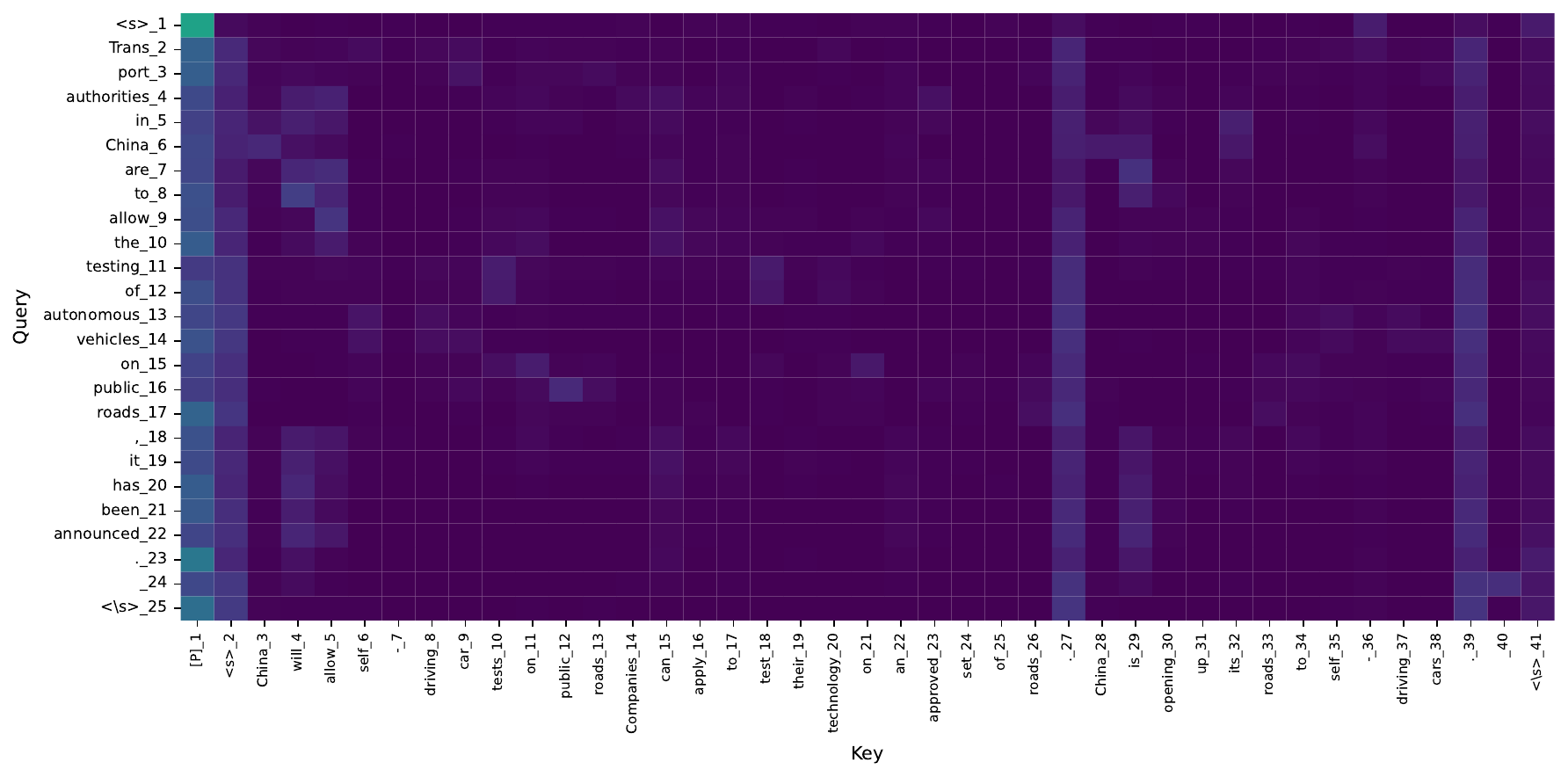}
\end{minipage}
\hfill
\begin{minipage}{.32\textwidth}
    \includegraphics[width=\textwidth]{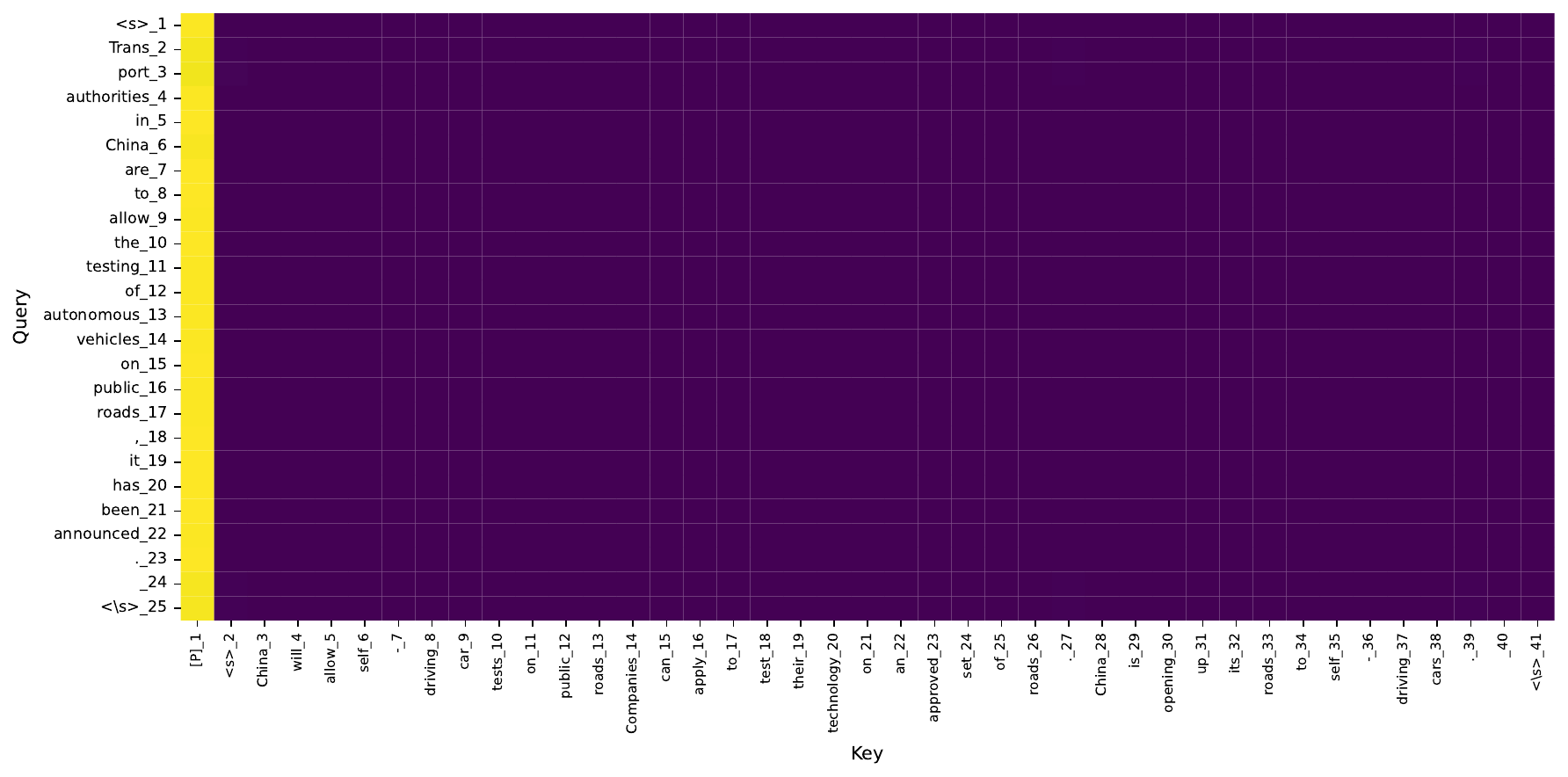}
\end{minipage}
\caption{
Decoder cross-attention maps at layer 3 averaged over all heads. \textbf{Top:} CNN/Dailymail. \textbf{Bottom:} Xsum. 
\textbf{Left:} Equivalence initialisation. \textbf{Middle:} Best post-training NVIB regularisation from validation set \textbf{Right:} Prior collapse.
}
\label{fig:attention_cross}
\end{figure}

\begin{figure}
    \begin{minipage}{.32\textwidth}
\includegraphics[width=\textwidth]{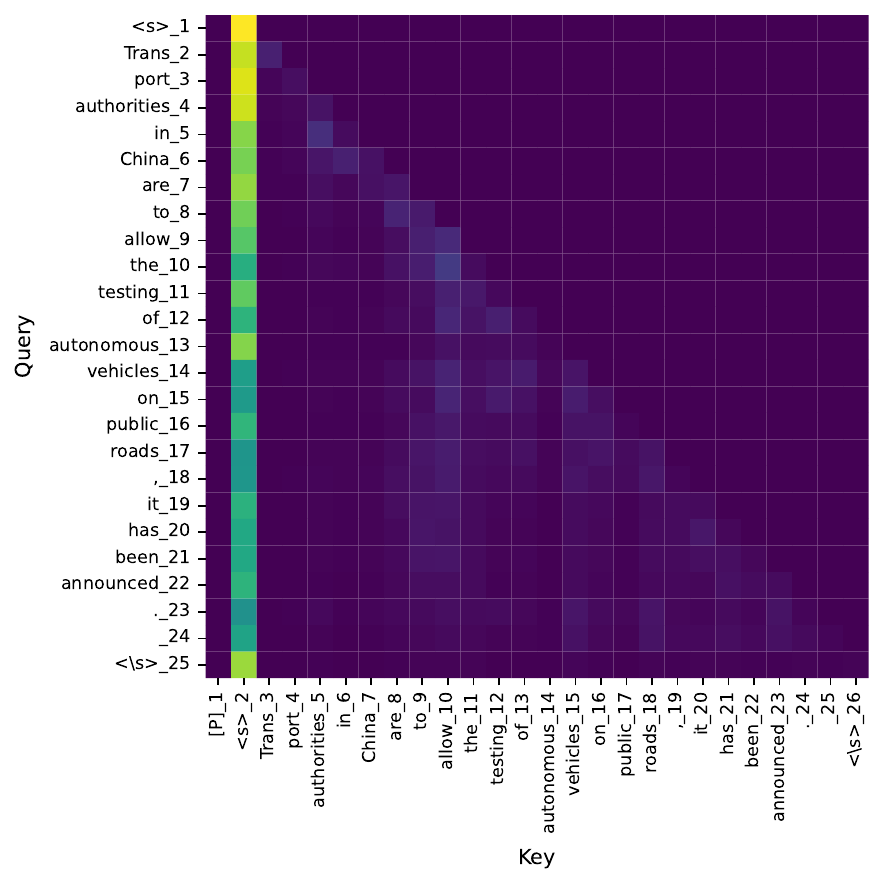}
\end{minipage}
\hfill
\begin{minipage}{.32\textwidth}
    \includegraphics[width=\textwidth]{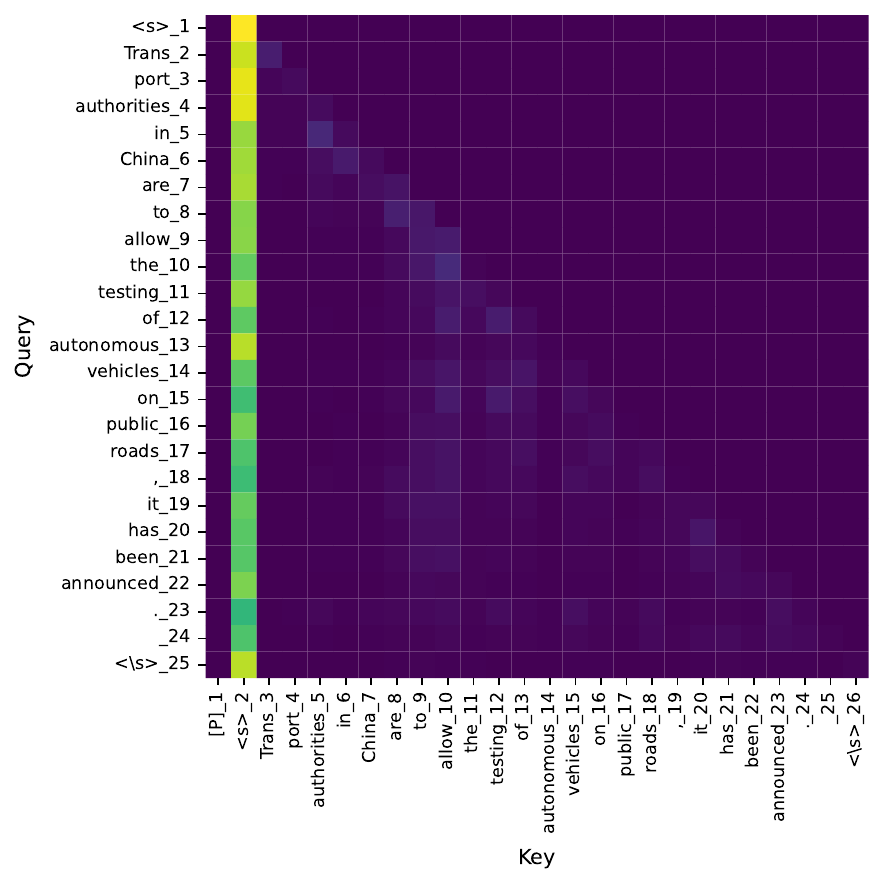}
\end{minipage}
\hfill
\begin{minipage}{.32\textwidth}
    \includegraphics[width=\textwidth]{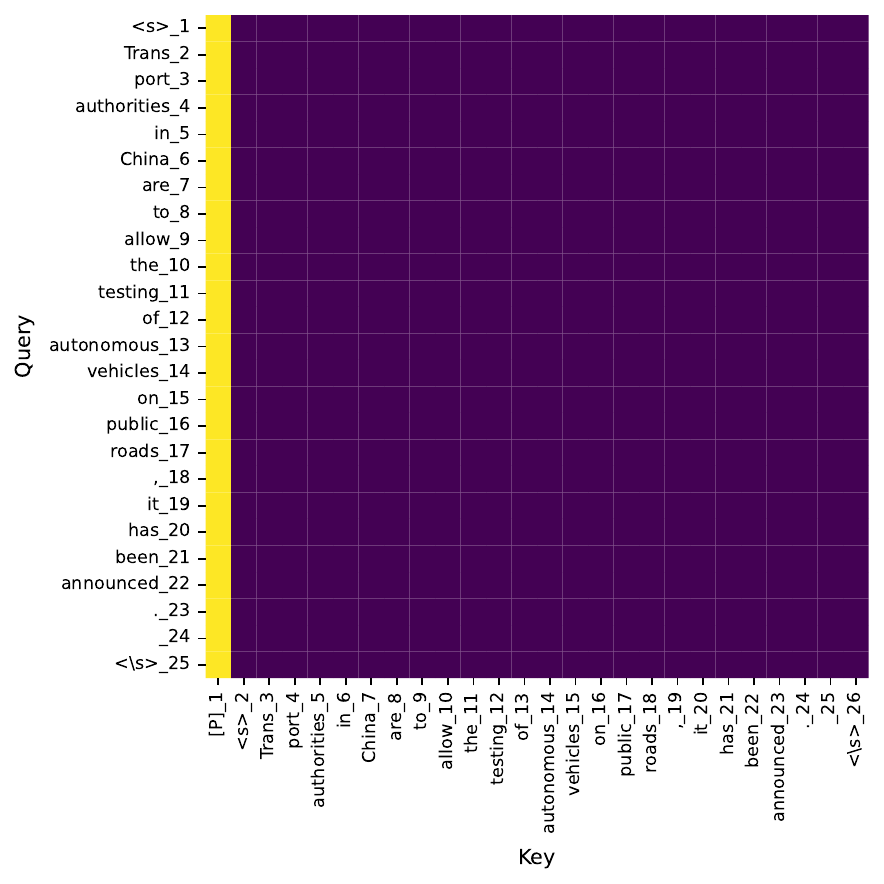}
    % \caption{}
\end{minipage}
%%%%%%%%%%%%%%%%%%%%%%%%%%%%%%%%
  \begin{minipage}{.32\textwidth}
\includegraphics[width=\textwidth]{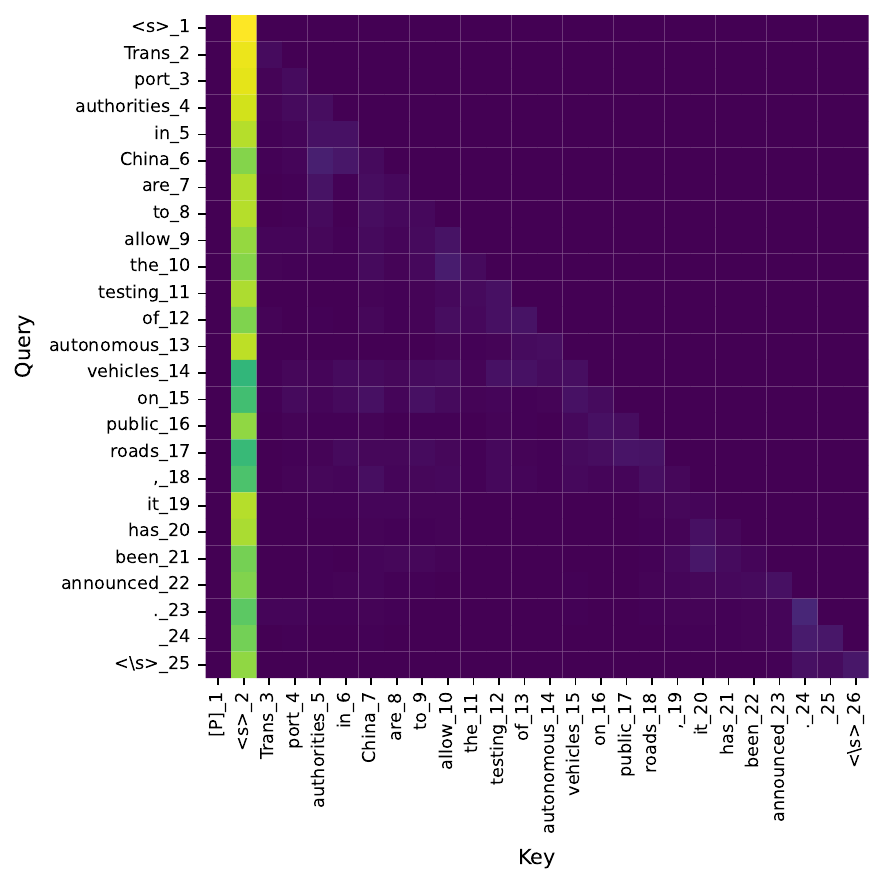}
\end{minipage}
\hfill
\begin{minipage}{.32\textwidth}
    \includegraphics[width=\textwidth]{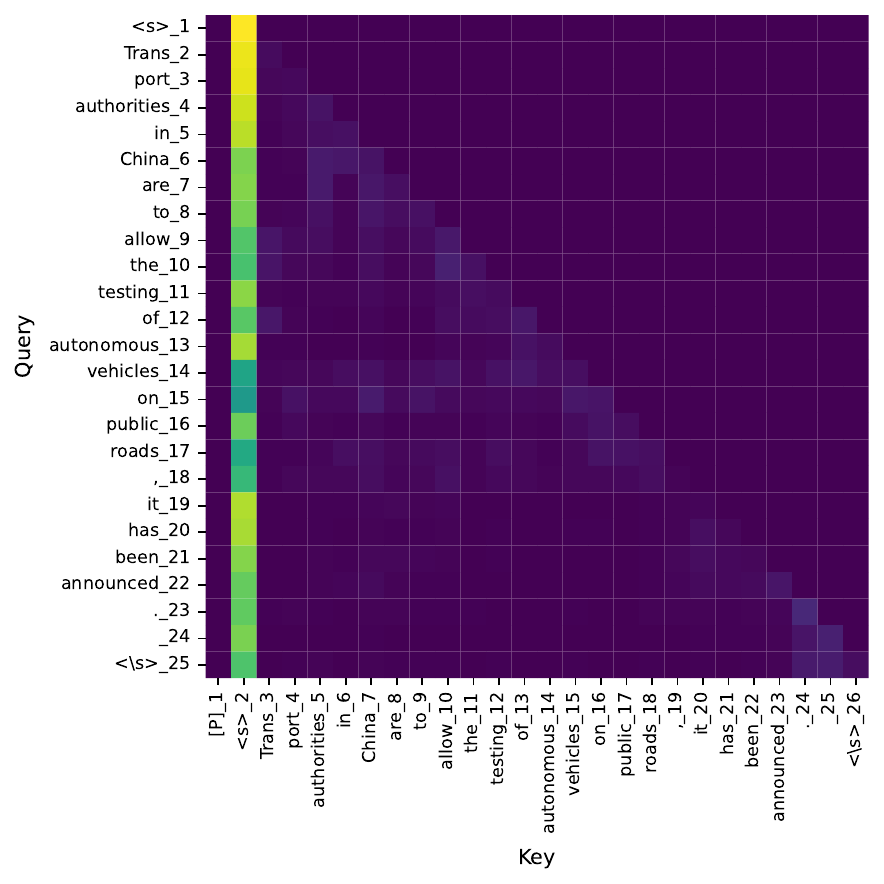}
\end{minipage}
\hfill
\begin{minipage}{.32\textwidth}
    \includegraphics[width=\textwidth]{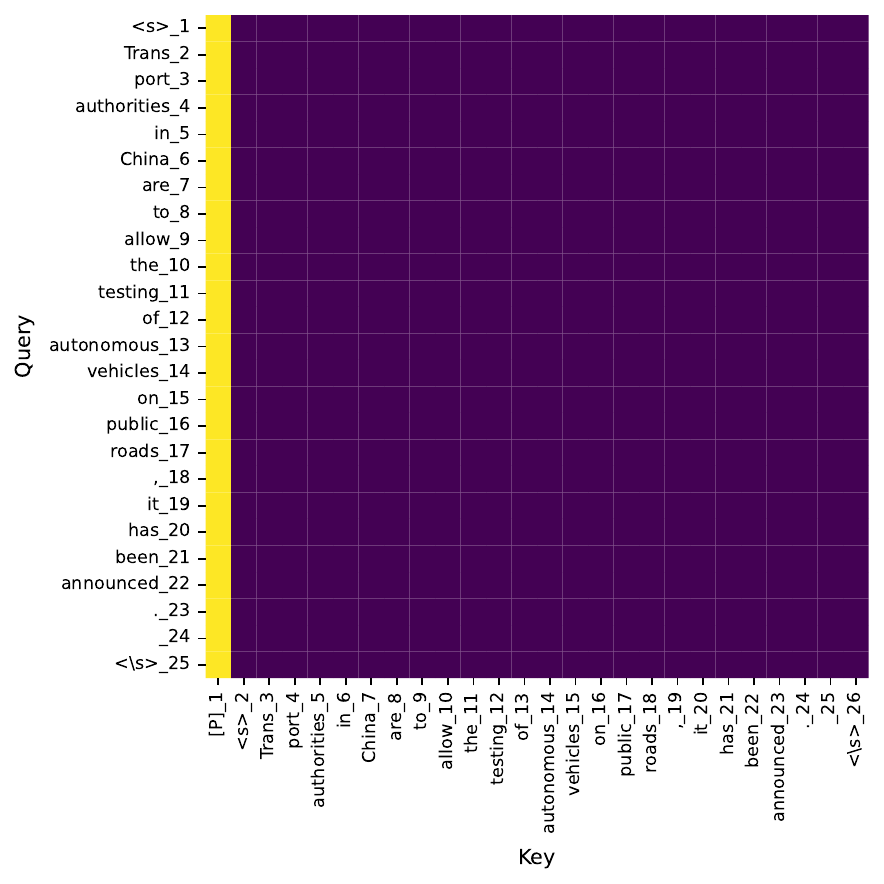}
    % \caption{}
\end{minipage}
\caption{
Decoder causal self-attention maps averaged over all heads. \textbf{Top:} CNN/Dailymail layer 6. \textbf{Bottom:} Xsum layer 6. 
\textbf{Left:} Equivalence initialisation. \textbf{Middle:} Best post-training NVIB regularisation from validation set \textbf{Right:} Prior collapse.
}
\label{fig:attention_decoders}
\end{figure}

In Figures \ref{fig:attention_encoders}, \ref{fig:attention_cross} and \ref{fig:attention_decoders} we see that when the model uses the identity initialisation the attention maps completely ignore the prior component representation $[P]$ and has equivalence in attention scores across all attention functions. In contrast, when $\tau_\alpha^i =-30$ the model has over emphasized the importance of the prior component, such that the attention patterns collapse to only considering the prior. When the model is regularised according to the best validation performance, we notice the attention patterns generally shift attention away from the vertical bars at special characters like punctuation and towards the prior. This shows that the models that are getting improvements have attention distributions which are regularised towards the prior.

% Used curation corpus data example. Then no regularisation is alphas 0 and sigmas approx zero then over regularised is sigma 0 and alphas +30 this means 30 standard deviations above the mean scaled L2 norm (prior) in log space. The the regularised model is the best selected model on the validation. This shows that the models that are getting improvements on the validation are indeed being regularised in attention toward the prior. We selected the layers that would emphasize this change. For both models CNN and XSUM encoder layer10 cross 3 and decoder 6 and 9 respectively. The models decoder is more sensitive to the regularisation thats why its different.

% In this section we analyse the attention plots given our NVIB regularisation. We consider the BART model trained on CNN/DailyMail and Xsum and calculate the empirical prior from their respective training corpora. We average over all 16 heads and show the model's 

% We use an shorted version of the curation dataset example. This is because both models saw an improvement in this dataset. We then selected the regularised model from the hyperparmaters and show selected layers. The affect on layers is not always obvious so we select specific layers where the model is looking at the prior.

% CNN model - Encoder layer 10, cross layer 3 , decoder layer 6 
% XSUM model -  Encoder layer 10, cross layer 3, decoder layer 9

% From the above Figures \ref{fig:attention_encoders}, \ref{fig:attention_cross} and \ref{fig:attention_decoders} 

\newpage

\subsection{Generated outputs} \label{sec:generated_outputs}
In this section we provide example output summaries from our baseline and reinterpreted models for the first example of each validation dataset. Each model only has a context of 1024 tokens in the BART model. We see that the models with post-training regularisation produce similar but different and sometimes improved predictions.

\begin{table}[h]
\begin{tabular}{|p{13.2cm}|}
\toprule
\textbf{SAMsum} \\
\midrule
\textbf{Article} \small{A: Hi Tom, are you busy tomorrow’s afternoon? B: I’m pretty sure I am. What’s up? A: Can you go with me to the animal shelter?. B: What do you want to do? A: I want to get a puppy for my son. B: That will make him so happy. A: Yeah, we’ve discussed it many times. I think he’s ready now. B: That’s good. Raising a dog is a tough issue. Like having a baby ;-) A: I'll get him one of those little dogs. B: One that won't grow up too big;-) A: And eat too much;-)) B: Do you know which one he would like? A: Oh, yes, I took him there last Monday. He showed me one that he really liked. B: I bet you had to drag him away. A: He wanted to take it home right away ;-). B: I wonder what he'll name it. A: He said he’d name it after his dead hamster – Lemmy - he's a great Motorhead fan :-))} \\ 
\textbf{Summary} \small{A will go to the animal shelter tomorrow to get a puppy for her son. They already visited the shelter last Monday and the son chose the puppy.} \\
\\
\textcolor{nvibgreen}{\textbf{BART} \scriptsize{(CNN/DM)}} \small{A mother wants to get a puppy for her son. She asks her son to accompany her to the animal shelter to buy him a dog. The boy takes the dog home and names it Lemmy after his dead hamster – Lemmy - a Motorhead fan. The dog is taken back to the shelter.} \\
\textcolor{nvibpurple}{\textbf{NV-BART} \scriptsize{(CNN/DM)}} \small{A mother wants to get a puppy for her son. She asks her son to take her to the animal shelter. He wants a little dog. She says he wanted to take it home right away. He said he’d name it after his dead hamster – Lemmy.}\\
\textcolor{nvibgreen}{\textbf{BART} \scriptsize{(Xsum)}} \small{A conversation between a mother and her son has been caught on camera, in which the mother says she wants to get a puppy for her son.} \\
\textcolor{nvibpurple}{\textbf{NV-BART} \scriptsize{(Xsum)}} \small{I’ve been chatting to my friend Tom and he’s got a puppy for his son and we’re planning to get him a dog.} \\
\bottomrule
\end{tabular}
\end{table}

\begin{table}
\begin{tabular}{|p{13.2cm}|}
\toprule
\textbf{XSum} \\
\midrule
\textbf{Article} \small{The ex-Reading defender denied fraudulent trading charges relating to the Sodje Sports Foundation - a charity to raise money for Nigerian sport. Mr Sodje, 37, is jointly charged with elder brothers Efe, 44, Bright, 50 and Stephen, 42. Appearing at the Old Bailey earlier, all four denied the offence. The charge relates to offences which allegedly took place between 2008 and 2014. Sam, from Kent, Efe and Bright, of Greater Manchester, and Stephen, from Bexley, are due to stand trial in July. They were all released on bail.} \\ 
\textbf{Summary} \small{Former Premier League footballer Sam Sodje has appeared in court alongside three brothers accused of charity fraud.} \\
\\
\textcolor{nvibgreen}{\textbf{BART} \scriptsize{(CNN/DM)}} \small{Sam Sodje, 37, is jointly charged with elder brothers Efe, 44, Bright, 50 and Stephen, 42. The charge relates to offences which allegedly took place between 2008 and 2014. Sam, from Kent, Efe and Bright, of Greater Manchester, and Stephen,. from Bexley, are due to stand trial in July.} \\
\textcolor{nvibpurple}{\textbf{NV-BART} \scriptsize{(CNN/DM)}} \small{ the Sodje Sports Foundation - a charity to raise money for Nigerian sport. Mr Sodje, 37, is jointly charged with elder brothers Efe, 44, Bright, 50, and Stephen, 42. The four deny fraudulent trading charges relating to the alleged offences between 2008 and 2014.
}\\
\textcolor{nvibgreen}{\textbf{BART} \scriptsize{(Xsum)}} \small{Former Premier League footballer Sam Sodje has appeared in court charged with fraud.} \\
\textcolor{nvibpurple}{\textbf{NV-BART} \scriptsize{(Xsum)}} \small{Former Premier League footballer Sam Sodje has appeared in court charged with fraud.
} \\
\bottomrule
\end{tabular}
\end{table}

\begin{table}
\begin{tabular}{|p{13.2cm}|}
\toprule
\textbf{Curation Corpus} \\
\midrule
\textbf{Article} \small{
China will allow self-driving car tests on public roads Companies can apply to test their technology on an approved set of roads.China is opening up its roads to self-driving cars. The Beijing Municipal Transport Commission released a statement today saying that on certain roads and under certain conditions, companies registered in China will be able to test their autonomous vehicles. Last year, Chinese authorities banned self-driving vehicles from the country's highways until new regulations could be created and approved. But those have been slow to arrive, which is why Chinese company Baidu and its CEO Robin Li came under fire this summer for conducting an apparently illegal demonstration of its driverless technology. The new regulations will allow companies to apply for temporary permission to test their autonomous vehicles on a yet-to-be-determined group of approved roads. The cars will have to have traffic accident liability insurance and a human behind the wheel to take over if anything goes wrong. With this development, China now joins a number of other countries allowing self-driving technology developers to test their products in real-life scenarios.
} \\ 
\textbf{Summary} \small{
iThe Beijing Municipal Transport Commission will license companies based in China to conduct tests of self-driving cars on some roads under certain conditions, set out in new regulations. These include a requirement for traffic accident insurance and for a human back-up driver to be behind the wheel to take over if necessary.
} \\
\\
\textcolor{nvibgreen}{\textbf{BART} \scriptsize{(CNN/DM)}} \small{China will allow self-driving car tests on public roads. Companies can apply to test their technology on an approved set of roads. The cars will have to have traffic accident liability insurance and a human behind the wheel to take over if anything goes wrong. China now joins a number of other countries allowingSelf-driving technology developers totest their products in real-life scenarios.} \\
\textcolor{nvibpurple}{\textbf{NV-BART} \scriptsize{(CNN/DM)}} \small{China will allow self-driving car tests on public roads. Companies can apply to test their technology on an approved set of roads. The cars will have to have traffic accident liability insurance and a human behind the wheel to take over if anything goes wrong. China joins a number of other countries allowing technology developers to test products in real-life scenarios.
}\\
\textcolor{nvibgreen}{\textbf{BART} \scriptsize{(Xsum)}} \small{China is opening up its roads to self-driving cars.} \\
\textcolor{nvibpurple}{\textbf{NV-BART} \scriptsize{(Xsum)}} \small{China is opening up its roads to self-driving car tests on public roads and under certain conditions.} \\
\bottomrule
\end{tabular}
\end{table}

\begin{table}
\begin{tabular}{|p{13.2cm}|}
\toprule
\textbf{CNN/DailyMail} \\
\midrule
\textbf{Article} \small{(CNN)Share, and your gift will be multiplied. That may sound like an esoteric adage, but when Zully Broussard selflessly decided to give one of her kidneys to a stranger, her generosity paired up with big data. It resulted in six patients receiving transplants. That surprised and wowed her. "I thought I was going to help this one person who I don't know, but the fact that so many people can have a life extension, that's pretty big," Broussard told CNN affiliate KGO. She may feel guided in her generosity by a higher power. "Thanks for all the support and prayers," a comment on a Facebook page in her name read. "I know this entire journey is much bigger than all of us. I also know I'm just the messenger." CNN cannot verify the authenticity of the page. ... (Beyond 1024 tokens input)
} \\ 
\textbf{Summary} \small{Zully Broussard decided to give a kidney to a stranger.} \\
\\
\textcolor{nvibgreen}{\textbf{BART} \scriptsize{(CNN/DM)}} \small{Zully Broussard gave one of her kidneys to a stranger. Her generosity paired up with big data. It resulted in six patients receiving transplants. "I thought I was going to help this one person who I don't know," she says. The chain of surgeries is to be wrapped up Friday.} \\
\textcolor{nvibpurple}{\textbf{NV-BART} \scriptsize{(CNN/DM)}} \small{Zully Broussard gave one of her kidneys to a stranger. Her generosity paired up with big data. It resulted in six patients receiving transplants. "I thought I was going to help this one person who I don't know, but the fact that so many people can have a life extension, that's pretty big"}\\
\textcolor{nvibgreen}{\textbf{BART} \scriptsize{(Xsum)}} \small{A San Francisco woman has become the first person in the world to donate a kidney to six people at once.} \\
\textcolor{nvibpurple}{\textbf{NV-BART} \scriptsize{(Xsum)}} \small{A woman in California has become the first person in the world to donate a kidney to a stranger in a "super-donor chain" by giving one of her own organs to another person in exchange for another person's kidney, in what is being hailed as an act of generosity.} \\
\bottomrule
\end{tabular}
\end{table}

\begin{table}
\begin{tabular}{|p{13.2cm}|}
\toprule
\textbf{WikiHow} \\
\midrule
\textbf{Article} \small{Adult humans should drink 2-3 litres(or roughly eight 8 oz glasses) of water per day whilst children should drink 1-2 litres ( or roughly five 8 oz glasses.That is in addition to things like tea and coffee. Water keeps bodies at the correct temperature and removes toxins that are the inevitable result of metabolism and industrial life. You'll be healthier automatically.
} \\ 
\textbf{Summary} \small{Drink more water.} \\
\\
\textcolor{nvibgreen}{\textbf{BART} \scriptsize{(CNN/DM)}} \small{Adult humans should drink 2-3 litres(or roughly eight 8 oz glasses) of water per day. Water keeps bodies at the correct temperature and removes toxins. Water also clears your skin, helps your kidneys, helps to control your appetite, and keeps you energized. It also keeps you from drinking unhealthy beverages like soda and juice.
} \\
\textcolor{nvibpurple}{\textbf{NV-BART} \scriptsize{(CNN/DM)}} \small{Adult humans should drink 2-3 litres of water per day. Children should drink 1-2 litres ( roughly five 8 oz glasses) Water keeps bodies at the correct temperature and removes toxins. Water also clears your skin, helps your kidneys, and keeps you energized. It also keeps you from drinking unhealthy beverages like soda and juice.
 }\\
\textcolor{nvibgreen}{\textbf{BART} \scriptsize{(Xsum)}} \small{If you want to be healthier, you need to drink more water.
} \\
\textcolor{nvibpurple}{\textbf{NV-BART} \scriptsize{(Xsum)}} \small{If you want to be healthier, and have better skin, you should drink more water.
} \\
\bottomrule
\end{tabular}
\end{table}

\end{document}